\documentclass[twoside]{article}

\usepackage[preprint]{aistats2027}

\usepackage[round]{natbib}

\usepackage[utf8]{inputenc} 
\usepackage[T1]{fontenc}    
\usepackage{url}            
\usepackage{booktabs}       
\usepackage{amsfonts}       
\usepackage{nicefrac}       
\usepackage{microtype}      
\usepackage{xcolor}         
\usepackage{color-edits}
\usepackage{times}
\usepackage{enumitem}

\usepackage{microtype}
\usepackage{graphicx}
\usepackage{booktabs} 

\usepackage{graphicx}
\usepackage{float} 
\usepackage{wrapfig}

\usepackage{amsthm}
\usepackage{amsmath}
\usepackage{amsfonts}
\usepackage{nicefrac}
\usepackage{microtype}
\usepackage{comment}
\usepackage{color}
\usepackage{cases}

\usepackage{tikz}
\usetikzlibrary{matrix}
\usetikzlibrary{calc}
\usetikzlibrary{decorations.pathreplacing}

\usepackage{amscd,latexsym,amsfonts,amssymb,amsmath,amsxtra}

\usepackage{tikz}
\usepackage{pgf, tikz}
\usetikzlibrary{arrows, automata}
\RequirePackage{amssymb}
\definecolor{mygray}{gray}{0.85}

\definecolor{mydarkblue}{rgb}{0,0.08,0.45}

\definecolor{mydarkblue}{rgb}{0,0.08,0.45}
\definecolor{antiquegold}{RGB}{205,127,50} 
\definecolor{deepskyblue}{RGB}{0,191,255} 
\definecolor{deepgreen}{RGB}{34, 139, 34}
\definecolor{deepred}{RGB}{178, 34, 34}
\definecolor{customBlue}{RGB}{24, 116, 205}

\usepackage{hyperref}
\hypersetup{
  colorlinks=true,
  linkcolor=mydarkblue ,    
  filecolor=mydarkblue ,  
  urlcolor=red ,     
  citecolor=mydarkblue       
}

\newcommand{\argmax}{\mathop{\mathrm{argmax}}}

     \newcommand{\BR}{{\mathbb {R}}}

     \newcommand{\CB}{{\mathcal {B}}}
     
     \newcommand{\CF}{{\mathcal {F}}}

     \newcommand{\CL}{{\mathcal {L}}}
     
    \newcommand{\CO}{{\mathcal {O}}} 
     
    \newcommand{\CS}{{\mathcal {S}}} \newcommand{\CT}{{\mathcal {T}}}

    \newcommand{\Sign}{{\mathrm{Sign}}}

    \newtheorem{thm}{Theorem}[section]
    
    \newtheorem{lem}[thm]{Lemma}
    \newtheorem{prop}[thm]{Proposition}

    \newtheorem{defn}[thm]{Definition}
    \newtheorem{assu}[thm]{Assumption}

\usepackage[font=small]{caption}
\begin{document}

\runningtitle{Adversarial Training with Large Step Sizes}
\runningauthor{Yi Feng, Andrea Paudice}

\twocolumn[

\aistatstitle{Adversarial Training with Large Step Sizes:\\
Implicit Bias, Optimization, and Sharpness}

\aistatsauthor{Yi Feng \And Andrea Paudice}

\aistatsaddress{Aarhus University, Denmark \\ \texttt{yi.feng@cs.au.dk}
\And Aarhus University, Denmark \\ \texttt{apaudice@cs.au.dk}}
]

\begin{abstract}%
\emph{Adversarial Training} (AT) is an important method for improving model robustness. We study the dynamics of deterministic \emph{Gradient Descent} (GD) in AT. Most existing works in this direction focus on the small or vanishing step sizes regime, where GD resembles a continuous-time flow. However, such step sizes can make training inefficient and are rarely used in practice. In this paper, we study the AT dynamics of GD with large constant step sizes. We first study GD in adversarial linear logistic regression, focusing on two aspects: where the model parameters converge (\emph{implicit bias}) and how quickly the losses decrease (\emph{optimization}). We then turn to neural networks training and study how the \emph{sharpness} of losses evolve during AT. Our results advance the understanding of the behaviors of GD with large step sizes in AT and highlight three key differences from its behavior in Standard Training (ST): (i) the implicit bias gap between AT and ST, previously established only for small step sizes, persists for arbitrarily large constant step sizes, (ii) different from ST, there is no uniform rate of convergence for the optimization error in AT, and (iii) sharpness can exhibit two-stage dynamics that differ from the \emph{Edge-of-Stability} behavior observed in ST.

\end{abstract}

\section{Introduction}
\emph{Adversarial training} (AT) \citep{szegedy2014,Goodfellow2015,Madry2018} is widely regarded as the \emph{de facto} approach for training models that are robust to perturbations of the input data \citep{Athalye2018}. Conceptually, adversarial training recasts empirical risk minimization as minimizing the worst-case loss over an $\ell_p$ ball around each training instance, typically choosing $p=2$ or $p=\infty$ to bound the perturbation's Euclidean norm or per-coordinate magnitude \citep{Pang2021}.

\begin{figure*}[t]
    \centering
    \includegraphics[width=\textwidth]{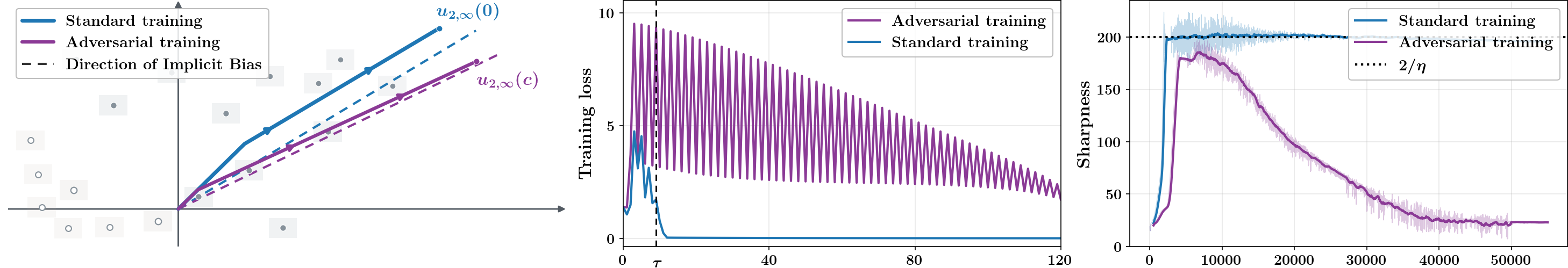}
    \caption{Illustrations of the main findings of this work. \textit{Left}: AT and ST  have different implicit bias under large constant step sizes. \textit{Middle}: The ST loss decreases monotonically after a phase transition time $\tau$, whereas the AT loss increases infinitely often. It can also be observed that the AT loss decreases very slowly. \textit{Right}: ST and AT can exhibit different sharpness evolutions. In particular, ST exhibits the edge-of-stability phenomenon, in which the sharpness oscillates around $2/\eta$, while the sharpness of AT decreases to a value much lower than $2/\eta$.}
    \label{fig:three_images}
\end{figure*}

In practice, the AT is typically solved using stochastic first-order methods, where the gradient calculation for the outer parameters relies on an approximate solution to the inner maximization problem \citep{Madry2018,Wong2020}. While highly effective empirically, analyzing the trajectory and convergence of these algorithms presents substantial challenges. This complexity stems from two primary sources. First, the inherent min-max nature of AT makes its optimization landscape considerably more intricate than that of \emph{Standard Training} (ST) \citep{liu2020loss}. Second, the stochasticity of the optimizer introduces an additional layer of complexity, compounded by the noise and approximations involved in solving the inner problem. To isolate these phenomena and gain a fundamental understanding of the unique features of AT compared to ST, recent literature \citep{Li2020Implicit,Zhang2020,Zhang2024,tsilivis2024price,lyu2025analyzing} has shifted its focus toward deterministic \emph{Gradient Descent} (GD), which dates back to the classical work of \citet{Cauchy1847} and forms the foundation of all first-order methods \citep{nesterov2013introductory}.

In this work, we focus on three complementary aspects of the training dynamics of AT via GD. The \emph{implicit bias} describes which solutions are selected by the optimization algorithm among those compatible with minimizing the training objective. The \emph{optimization error} quantifies how quickly the loss approaches its infimum. Finally, \emph{sharpness} measures the local curvature of the loss landscape along the training trajectory and plays an important role in understanding the optimization dynamics of deep neural networks \citep{foret2020sharpness}. Together, these perspectives address which predictors training selects, how efficiently it reaches low loss, and how the local geometry evolves. While these foundational questions have been extensively probed in ST, existing theoretical work on AT remains severely constrained in its algorithmic scope. Specifically, the vast majority of current literature on the dynamics of AT relies on infinitesimal learning rates, which makes the dynamics become a continuous-time gradient flow \citep{gu2021adversarial,tsilivis2024price}, or restricts the analysis to small step sizes that are exponentially small in the dimension \citep{Li2020Implicit,lv2022implicit,lyu2025analyzing}. Such frameworks fail to capture the behavior of practical optimization regimes, which frequently employ large constant step sizes \citep{Pang2021} that induce non-monotonic loss behaviors and complex sharpness behaviors like the \emph{Edge-of-Stability} (EoS) phenomena \citep{cohen2021gradient}. In this work, we begin by studying the implicit bias and optimization of GD with \textbf{arbitrary constant step sizes} in adversarial linear logistic regression with separable data, whose standard training counterpart is known to exhibit new phenomena absent from the small step size analyses \citep{wu2023implicit,wu2024large}. We then turn to an empirical study of adversarial neural network training, studying how the sharpness evolves during adversarial training and comparing it with the EoS phenomena in standard training.

\subsection{Our Contributions.}

In this work, we focus mainly on $\ell_{\infty}$-adversarial attacks since it is the mostly used in practice, and also less theoretically understood compared with other norms \citep{wang2025faster}. Most of our results can also be  extended to $\ell_p$-attacks for $p\in[1,\infty]$ using the $\ell_p/\ell_q$ duality, except for the results in Section~\ref{sec:op}, which relies on properties specific to the $\ell_{\infty}$ norm. We summarize our main contributions below.

\paragraph{Implicit Bias of GD for Adversarial Logistic Regression.} For adversarial logistic regression, we prove that, under standard separability assumption on the dataset, GD with \emph{arbitrary} constant step size converges to a \emph{robust max-margin direction} $\mathbf{u}_{2,\infty}(c)$ depending on the adversarial radius  $c>0$.  
(Theorem~\ref{thm:main1}) This behaves differently from ST, where the implicit bias is driven toward the standard max-margin direction on the original data, see the left of Figure~\ref{fig:three_images} for a comparison between these two directions. While this discrepancy was first shown by \citet{Li2020Implicit}, our work extends their findings in two aspects. First, our result holds for arbitrary constant step sizes, whereas \citet{Li2020Implicit} require step sizes exponentially small with dimension. Second, we establish a rate of order $\ln \ln T/\ln T$ for the implicit bias, improving over \cite{Li2020Implicit} by a factor of order $1/\sqrt{\ln T}$. In Proposition~\ref{prop:tight_rate} we also show that this rate is tight, thereby providing a comprehensive characterization of the implicit bias for this problem.

\paragraph{Optimization of GD for Adversarial Logistic Regression.} Recent work by \citet{wu2024large} characterizes GD for standard logistic regression with large step sizes through two phases separated by a transition time $\tau$: the loss behaves non-monotonically before $\tau$ and monotonically afterward. 
They further show that this non-monotonic phase is necessary to achieve an accelerated loss convergence rate of $\tilde{\mathcal{O}}(1/T^2)$. Surprisingly, we find that both the existence of the transition time and the acceleration effect of large step size can fail for AT. In fact, the situation of AT is even worse: Theorem~\ref{prop:counter_exp_opt} shows that, for any adversarial radius $c>0$ and time budget $T$, there exists a dataset on which GD's optimization error is bounded below by a positive constant depending only on $c$.  Moreover, there exist infinitely many steps $t$ where the loss increases from $t$ to $t+1$.  In the middle of Figure~\ref{fig:three_images}, we illustrate this phenomenon. These results show that the optimization aspect of GD with large constant step sizes can behave differently in AT than in ST.

\paragraph{Sharpness Evolution in AT.} 
We conduct experiments on AT with neural networks to study how sharpness evolves along the training  trajectory. We observe that the sharpness initially increases monotonically and then begins to oscillate  with an overall decreasing trend. This behavior remains consistent across different architectures, datasets, and adversarial attack generation approaches. It also differs from the EoS phenomenon in ST of neural network with deterministic GD \citep{cohen2021gradient}, where sharpness oscillates around $2/\eta$ for step size $\eta$ after an initial monotone increasing stage. The right of Figure~\ref{fig:three_images} present their difference. To develop theoretical understanding of this phenomenon, we analyze a min-max variant of the theoretical model proposed in \citet{ahn2023learning}. Our analysis shows that the sharpness of this model converges points with a sharpness level significantly below the standard sharpness level $2/\eta$, consistent with our empirical findings. These results show that sharpness evolves very differently in standard and adversarial neural network training.

\subsection{Related Works}
\paragraph{Implicit Bias of  Linear Logistic Regression.} In standard training, \citet{soudry2018implicit} and \citet{ji2018risk} showed that GD converges to the max-margin direction for linear models on separable data. Subsequent work studied other training algorithms than GD, like steepest descent \cite{gunasekar2018characterizing}, momentum methods \citep{wang2022does,zhang2024implicit}, and spectral descent method like Muon \citep{fan2026implicit}.
Most of these analyses require sufficiently small step sizes. An exception is \citet{wu2023implicit}, who established the implicit bias of GD with any constant step size under additional assumptions. For AT, \citet{Li2020Implicit} showed that GD for separable adversarial  logistic regression converges to \textit{robust max-margin direction} with exponentially small step size. Subsequent work extended the results on adversarial training to other linear networks and training dynamics \citep{lv2022implicit,tsilivis2024price,lyu2025analyzing,lyu2025effects}. \citet{wang2025faster} derived faster robust
margin-maximization rates for other optimization methods than GD, and their analysis does not extend to $\ell_{\infty}$ adversarial attacks.

\paragraph{Optimization with Large Step Sizes.} 
In practice, large step sizes can lead to non-monotonic loss trajectories while preserving favorable long-term behavior. For standard logistic regression, \citet{wu2024large,wu2026large} showed that large step size gradient descent can accelerate the convergence rate despite oscillations. \citet{cai2024large} extends the results to non-Homogeneous two-layer neural networks. \citet{pmlr-v336-crawshaw26a} established tighter bounds for two-dimensional separable logistic regression. Beyond the convergence rate analysis, \cite{andriushchenko2023sgd} showed that large step sizes help neural networks to learn sparse features to improve generalization. \cite{meng2024gradient} studied the bifurcation behaviors of gradient descent with large step sizes in logistic regression with \textit{non-separable} dataset. \cite{liang2026gradient} characterized critical step sizes for scalar-vector factorization and showed that large step sizes can induce  chaotic optimization dynamics.

\paragraph{Sharpness \& Edge-of-Stability.} Sharpness
plays an important role in neural network training. \citet{cohen2021gradient} observed the edge-of-stability (EoS) phenomenon, in which sharpness initially increases and then hovers near $2/\eta$. Several works have developed theoretical explanations for these dynamics. For example, \citet{damian2022self} identified the self-stabilization mechanism.
\citet{cohen2025understanding} introduced central flows to model the time-averaged trajectories of oscillatory optimizers. \citet{ahn2023learning} connected EoS to generalization performance of neural networks. Another lines of work theoretically studied the EoS through a minimalist example approach \citep{kreisler2023gradient,zhu2023understanding,wang2025good,liu2026minimalist}. Recent works of \cite{islamov2026non} and \cite{song2026zeroth} generalized the EoS to Riemannian and zero-th order optimization settings.


\section{Preliminaries}\label{sec:pre}

\subsection{Adversarial Linear Logistic Regression}\label{sec:alr}
Given a dataset $\{(\mathbf{x}_i,y_i)\}_{i=1}^N$ where $(\mathbf{x}_i,y_i)
\in \mathbb{R}^d\times\{\pm1\}$, we consider
$\ell_\infty$-adversarial logistic regression with adversarial radius $c\ge0$.
The objective is to minimize $\CL_{\mathrm{Adv}}(\mathbf{w})$
over $\mathbf{w}\in\mathbb{R}^d$, where
\begin{equation}
\begin{aligned}
\CL_{\mathrm{Adv}}(\mathbf{w})
&:= \sum_{i=1}^N
\max_{\|\boldsymbol{\delta}_i\|_\infty\le c}
\ln\bigl(
1+e^{-y_i(\mathbf{x}_i+\boldsymbol{\delta}_i)^\top\mathbf{w}}
\bigr)\\
&= \sum_{i=1}^N
\ln\bigl(
1+e^{-y_i\mathbf{x}_i^\top\mathbf{w}
+c\|\mathbf{w}\|_1}
\bigr).
\end{aligned}
\label{equ:advloss}
\tag{$\ell_{\infty}$-adversarial logistic regression}
\end{equation}
Here the second equality follows from the duality between $\ell_{\infty}$ and $\ell_1$ norms. When $c=0$, this reduces to the standard logistic loss $\CL(\mathbf w)$. We consider the (sub)gradient descent (GD) dynamics:
\begin{align}\label{equ:gd}
\mathbf{w}_{t+1} &=    \mathbf{w}_{t} - \eta\, \partial_{ \mathbf{w}}  \CL_{\mathrm{Adv}}(\mathbf{w}_t). \tag{GD}
\end{align}
Here $\eta >0 $ is the step size. In this work, following \cite{wu2024large}, we use the term \textit{large step size} to refer to any constant step size for the iteration in \eqref{equ:gd}. We also always select the minimum Euclidean-norm subgradient of
\(\|\mathbf w\|_1\), whose \(j\)-th component is zero whenever \(w_j=0\).
This choice is standard in the literature
\citep{JMLR:v12:shalev-shwartz11a,scetbon2023robust} and is consistent with PyTorch's autograd
convention.

\begin{defn}\label{def:111}
For data set $\{(\mathbf{x}_i,y_i)\}_{i=1}^N$ and adversarial radius $c$, we define the following notations:
\begin{itemize}[leftmargin=*]
\item \textit{Robust maximum margin}:
\[
\gamma_{2,\infty}(c) = \max_{\lVert \mathbf{w} \rVert_2=1} \min_{i \in [N]}\min_{\lVert \boldsymbol\delta_i\rVert_{\infty}\le c} y_i\langle \mathbf{x}_i+\boldsymbol\delta_i, \mathbf{w}\rangle
\]
\item \textit{Robust max-margin direction}:
\[
\mathbf{u}_{2,\infty}(c) := \argmax_{\lVert \mathbf{w} \rVert_2=1} \min_{i \in [N]}\min_{\lVert \boldsymbol\delta_i\rVert_{\infty}\le c} y_i\langle \mathbf{x}_i+\boldsymbol\delta_i, \mathbf{w}\rangle
\]
\end{itemize}    
\end{defn}
Here, $\gamma_{2,\infty}(c)$ represents the $\ell_2$ margin of the extended dataset: \[\{(\mathbf{x}',y')\mid \exists i,\ \|\mathbf{x}'-\mathbf{x}_i\|_{\infty}\le c,\ y'=y_i\},\] which includes all admissible perturbations of the original samples with their labels unchanged, and $\mathbf u_{2,\infty}(c)$ is the corresponding max-margin direction. We assume that the dataset is robustly linearly separable, meaning that the extended dataset remains linearly separable. This assumption is standard in analyses of \eqref{equ:gd} for adversarial training \citep{charles2019convergence,Li2020Implicit,lyu2025analyzing}.
\begin{assu}\label{ass1}
The dataset $\{(\mathbf{x}_i, y_i)\}_{i=1}^N$ is robustly linearly separable, i.e., $\gamma_{2,\infty}>0$.
\end{assu}

\subsection{General Adversarial Training}\label{sec:gat}
Adversarial training (AT) is a widely used approach to improving
the robustness of machine learning models.
For $\ell_\infty$-adversarial attack with radius $c$, it minimizes
$\CL_{\mathrm{Adv}}(\mathbf w)$, defined as
\begin{equation}\label{equ:AT}
\begin{aligned}
\CL_{\mathrm{Adv}}(\mathbf w)
&= \frac{1}{N}\sum_{i=1}^N g_c\left(\mathbf{x}_i,\mathbf w\right),
\quad \mathrm{where}\\
g_c(\mathbf{x}_i,\mathbf w)
&= \max_{\|\mathbf{x}'_i-\mathbf{x}_i\|_\infty\le c}
g(\mathbf{x}'_i,\mathbf w).
\end{aligned}
\tag{AT}
\end{equation}
Here, $g$ is a loss function, such as mean squared error
(MSE), with dependence on the label $y_i$
suppressed for simplicity. In practice, a common approximation of the inner maximization  problem is the fast gradient sign method
(FGSM) \citep{Goodfellow2015}:
\begin{align}\label{equ:fgsm}
\mathbf{x}'_i
= \mathbf{x}_i
+ c\,\Sign\!\left(\nabla_{\mathbf x}g(\mathbf{x}_i,\mathbf w)\right).
\tag{FGSM}
\end{align}
Alternatively, projected gradient descent (PGD)
\citep{Madry2018} approximately solves the inner
maximization problem using $\CT$ projected gradient ascent
steps. Starting from $\mathbf{x}_{i,0}\in\CB_i$, it updates
\begin{align}
\mathbf{x}_{i,t}
&= \mathrm{Proj}_{\CB_i}\!\left(
\mathbf{x}_{i,t-1}
+ \alpha\,\Sign\!\left(
\nabla_{\mathbf x}g(\mathbf{x}_{i,t-1},\mathbf w)
\right)\right), \nonumber
\label{equ:pgd}
\tag{PGD}
\end{align}
for $t=1,\ldots,\CT$. Here $\alpha>0$ is the attack step size and
$\CB_i:=\{\mathbf{x}' :
\|\mathbf{x}'-\mathbf{x}_i\|_\infty\le c\}$.
Then the adversarial example used in training is defined as  $\mathbf{x}'_i:=\mathbf{x}_{i,\CT}$. We define the \textit{sharpness} at $\mathbf w$ as the largest
eigenvalue of the parameter Hessian evaluated on fixed
adversarial examples $\{\mathbf x'_i\}_{i=1}^N$, namely,
\begin{align}\label{equ:adv_sharp}
\lambda_{\max}\!\bigl(
\tfrac{1}{N}\sum\nolimits_{i=1}^{N}
\nabla_{\mathbf w}^{2}g(\mathbf x_i',\mathbf w)
\bigr)\tag{sharpness}
\end{align}
Following the standard practice in sharpness analyses of AT, we compute the parameter Hessian while holding the adversarial examples fixed \citep{yao2018hessian,liu2020loss,sitawarin2021sat,wang2023improving,NEURIPS2025_36f4029b,walter2026flatness}. Computing the Hessian while accounting for the dependence of $\mathbf{x}_i'$ on $\mathbf w$ would require differentiating through the adversarial example generation, making it computationally prohibitive.  When $c=0$, the notation is same as the sharpness in ST.

\section{Implicit Bias of Adversarial Logistic Regression}\label{sec:ib}
In this section, we study the implicit bias of gradient descent for $\ell_{\infty}$-adversarial logistic regression, while proofs are deferred to Appendix~\ref{app:ptmain1}. Following earlier analyses of implicit bias of standard training \citep{pmlr-v89-nacson19b,ji2021characterizing,wu2023implicit}, we decompose each gradient descent iterate of adversarial training into its projection onto the robust max-margin direction and its orthogonal component:
\begin{align}\label{equ:odmain}
\mathbf w_t = v_t \mathbf u_{2,\infty}(c) + \mathbf r_t,    \tag{orthogonal decomposition}
\end{align}
where $v_t = \langle \mathbf{w}_t,\mathbf{u}_{2,\infty}(c)\rangle$
is the projection of $\mathbf{w}_t$ onto the robust
max-margin direction, and $\mathbf{r}_t$ is the orthogonal
component, i.e.,
$\langle \mathbf{r}_t,\mathbf{u}_{2,\infty}(c)\rangle = 0$. The ratio $\lVert \mathbf r_t \rVert_2/v_t$ determines the convergence rate of implicit bias. We present the main theorem of this section below.

\begin{thm}\label{thm:main1}
Under Assumption~\ref{ass1}, fix any constant step size $\eta>0$ and let
$\{\mathbf{w}_t\}_{t\ge0}$ be given by the \eqref{equ:gd} dynamic with $\mathbf{w}_0 = \mathbf 0$\footnote{The zero initialization is assumed only for simplicity, and the results can be extended to any fixed $\mathbf w_0\in\mathbb R^d$.}. Then, for every $t \geq 1$, we have 
\begin{enumerate}[label=\textnormal{(\roman*)}]
\item\label{itm:main1_vt} 
$v_t\ge\frac1{\gamma_{2,\infty}(c)}
\ln\left(1+\frac{\eta\gamma_{2,\infty}^2(c)}2t\right)$.
\item\label{itm:main1_rt} $\exists\,C_{\mathbf r} > 0$, independent of $t$, s.t. $\lVert \mathbf r_t \rVert_2 \le C_{\mathbf r} \ln \ln t.$
\item\label{itm:main1_rate} $\left\lVert\, \frac{\mathbf{w}_t}{\lVert\mathbf{w}_t\rVert_2}-\mathbf{u}_{2,\infty}(c)\,\right\rVert_2 \le 2\lVert\mathbf{r}_t\rVert_2/v_t
= \mathcal{O} \left(\frac{\ln\ln t}{\ln t}\right)$.
\end{enumerate}
\end{thm}
We make the following comments. Previous work of \citet{Li2020Implicit} established the implicit bias of
\eqref{equ:gd} toward the robust max-margin direction $\mathbf{u}_{2,\infty}(c)$ under Assumption~\ref{ass1} and with step size at most of order $e^{-\Theta(\sqrt d)}$. Theorem~\ref{thm:main1} strengthens their result in two respects. First, it allows  any positive constant step size. This is practically relevant because constant step sizes are often chosen empirically, and exponentially small step sizes are rarely used in practice \citep{Pang2021}. Second, it improves the implicit bias convergence rate established by \citet{Li2020Implicit} from $\mathcal{O}(1/\sqrt{\ln t})$ to $\mathcal{O}(\ln \ln t / \ln t)$. The next result shows that this improved rate is tight. 
\begin{prop}\label{prop:tight_rate} 
The dataset $\mathbf x_1 =  (1/2,1/4),\, \mathbf x_2 =  (1/2,1/2),\,y_1=y_2=1,\, c=1/4$ is robustly linearly separable. The \ref{equ:gd} trajectory $\{\mathbf w_t\}^{\infty}_{t=1}$ on it satisfies
\begin{align*}
\lim_{t \to \infty}\frac{\ln t}{\ln\ln t}
\left\lVert\frac{\mathbf{w}_t}{\lVert\mathbf{w}_t\rVert_2}-\mathbf u_{2,\infty}(c)\right\rVert_2=1.
\end{align*}
\end{prop}
%
%
Finally, Theorem~\ref{thm:main1} directly implies a convergence rate of
\begin{align}\label{eq:ub_rate}
\CL_{\mathrm{Adv}}(\mathbf w_t) = \widetilde{\mathcal{O}}\left(1/t\right),
\end{align}
where $\widetilde{\mathcal{O}}$ hide the $\log$ factors and the dependence on $\eta$ and the other parameters of the problem. 

\paragraph{Proof Overview of Theorem~\ref{thm:main1}.} 
Our analysis builds on the proof strategy of \citet{wu2023implicit} for ST with large step sizes. We handle the irregularities in adversarial training, such as nonsmoothness, through refined
comparison points and tighter cumulative loss difference bounds as explained below. In the following, we illustrate our proof by recovering the $\CO(1/\sqrt{\ln t})$ rate of \citet{Li2020Implicit}
under arbitrary constant step sizes. We then explain how to refine this argument to obtain the full result. The complete proof is provided in Appendix~\ref{app:ptmain1}.

Part \ref{itm:main1_vt} of Theorem~\ref{thm:main1} follows from the fact that, at each iteration, there exists at least one data $\mathbf{x}_j$ whose robust margin $ \min_{\lVert \boldsymbol\delta_i\rVert\le c}y_j\langle \mathbf{x}_j +\boldsymbol\delta_i ,\mathbf{w}_t\rangle$ is at most $\gamma_{2,\infty}v_t$, ensuring sufficient growth of $v_t$ (see Proposition~\ref{lem:trainingproj}, similar argument also used in Lemma B.6 of \cite{wu2023implicit}). In the following we focus on bounding the term $\lVert \mathbf r_t\rVert_2$. Let $\{\tilde{\mathbf w}_t\}_{t=1}^{\infty}$ be a series of comparison points with known bounds on its norms and loss values. The proof starts from the following comparison inequality, which holds for any $\eta>0$ and is derived using the convexity of \ref{equ:advloss}:
\begin{equation}\label{equ:advproofmain}
\begin{aligned}
\lVert \mathbf w_{t} -  \tilde{\mathbf w}_{t}\rVert^2_2 \le{}&
2\,\eta \sum_{s < t} \left( \CL_{\mathrm{Adv}}(\tilde{\mathbf w}_s) - \CL_{\mathrm{Adv}}(\mathbf w_s) \right)\\
&+ \sum_{s < t} \lVert \mathbf{w}_{s+1} -\mathbf{w}_s \rVert^2_2
\end{aligned}
\end{equation}
We choose the comparison points $\{\tilde{\mathbf w}_t\}$ according to 
\[
\tilde{\mathbf w}_t = \frac1{\gamma_{2,\infty}(c)}
\ln\left(1+\frac{\eta\gamma_{2,\infty}^2(c)}2t\right) \mathbf{u}_{2,\infty}(c),
\]
whose norm and cumulative adversarial loss are both $\mathcal O(\ln t)$. Intuitively, the growth of $\mathbf w_t$ along $\mathbf u_{2,\infty}(c)$, measured by $v_t=\langle\mathbf w_t,\mathbf u_{2,\infty}(c)\rangle$, controls the updates of $\mathbf{w}_t$, allowing us to establish
$\sum_{s<t}\lVert\mathbf w_{s+1}-\mathbf w_s\rVert_2^2\le Kv_t$
for some constant $K$ (see Lemma~\ref{lem:ib_preliminary}). Dropping the negative loss term in \eqref{equ:advproofmain}, and using
\[v_t=\langle\mathbf w_t-\tilde{\mathbf w}_t,\mathbf u_{2,\infty}(c)\rangle+\mathcal O(\ln t)
\le\lVert\mathbf w_t-\tilde{\mathbf w}_t\rVert_2+\mathcal O(\ln t),\] we get
\[
\lVert \mathbf r_t \rVert^2_2 \le \lVert \mathbf w_{t} -  \tilde{\mathbf w}_{t}\rVert^2_2 \le \mathcal O(\ln t) + K \lVert \mathbf w_{t} -  \tilde{\mathbf w}_{t}\rVert_2,
\]
which implies $\lVert \mathbf r_t \rVert_2 \le \lVert \mathbf w_{t} -  \tilde{\mathbf w}_{t}\rVert_2  = \mathcal O(\sqrt{\ln t})$.  Combing this with the lower bound of $v_t$ in Part $\ref{itm:main1_vt}$ of Theorem~\ref{thm:main1} gives the $\CO(1/\sqrt {\ln t})$ rate of implicit bias. Note that even in this simple setting, our analysis already differs from that of \citet{wu2023implicit} for standard training, where additional rank assumptions on the dataset (Assumption 2 \& 3 in their work) are used to establish a uniform bound on $\lVert \mathbf r_t \rVert_2$. We avoid such assumption by constructing suitable comparison points, which adaptive to the adversarial training process and gives a refined growth rate of $\lVert \mathbf r_t \rVert_2$.

To obtain the $\mathcal O(\ln\ln t/\ln t)$ rate, we sharpen the bound on $\|\mathbf r_t\|_2$ from $\mathcal O(\sqrt{\ln t})$ in above argument to $\CO(\ln\ln t)$ by choosing more suitable comparison points $\{\tilde{\mathbf w}_t\}_{t=1}^{\infty}$. The new construction includes vector directions that are orthogonal to $\mathbf{u}_{2,\infty}(c)$, thus is more close to the gradient descent trajectories for adversarial logistic regression (see Proposition~\ref{lem:ib_approx}). We also refine the analysis of the cumulative adversarial loss difference in \eqref{equ:advproofmain} by retaining the negative loss term (see Proposition~\ref{lem:ib_loss_gap}), and use the fact that the cumulative squared updates in \eqref{equ:advproofmain} are in fact uniformly bounded (see Lemma~\ref{lem:ib_preliminary}).

\section{Optimization of Adversarial Logistic Regression}\label{sec:op}
\begin{figure*}[t]
    \centering
    \includegraphics[width=\textwidth]{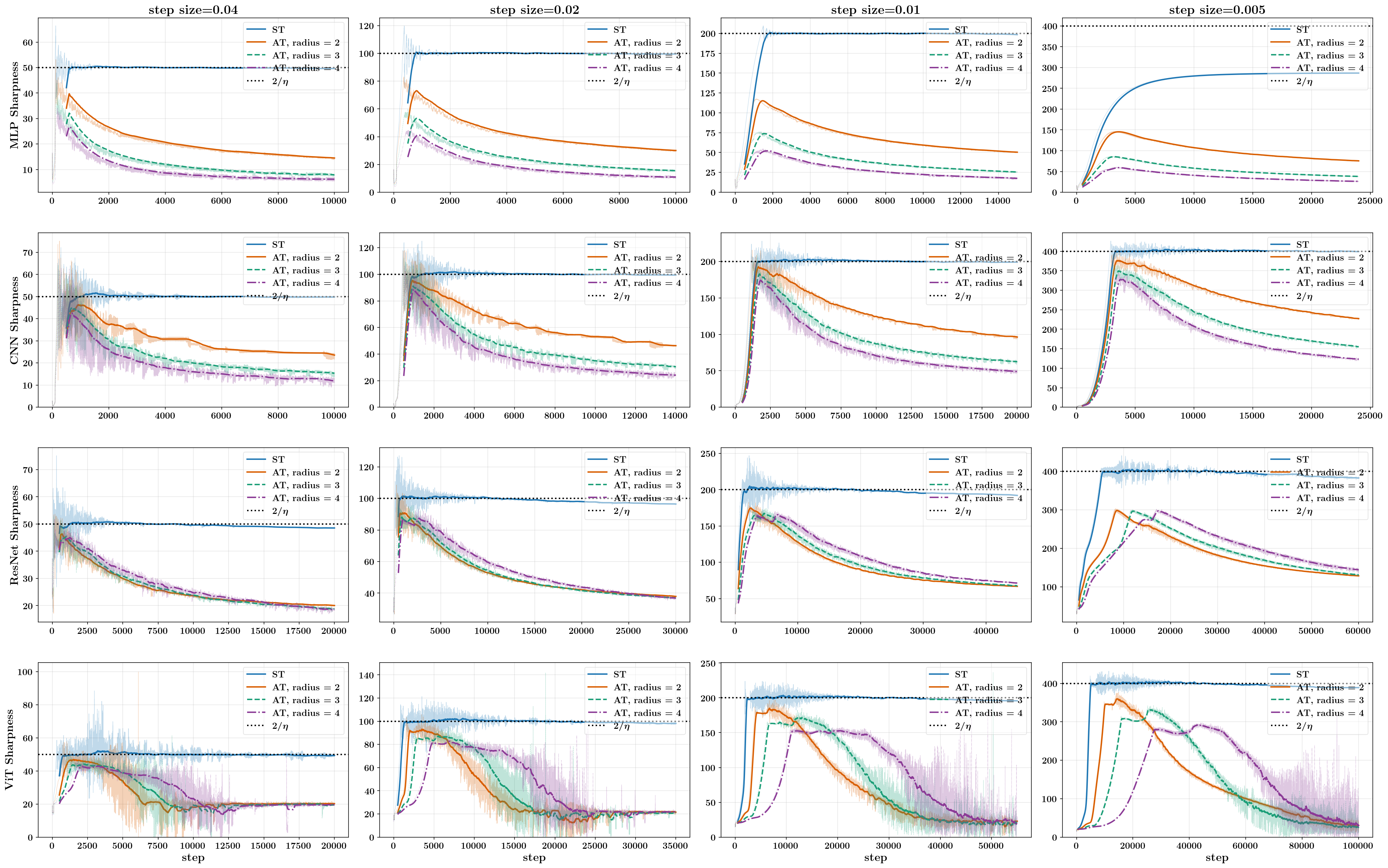}
    \caption{Sharpness evolution across neural network architectures
(MLP, CNN, ResNet, and Vision Transformer) and step sizes under
standard training (ST) and $\boldsymbol{\ell_{\infty}}$-adversarial training on CIFAR-10.
Adversarial examples are generated using \eqref{equ:fgsm}.
Except for the MLP at step size $0.005$, sharpness under ST initially increases and then oscillates around $2/\eta$, consistent with the edge-of-stability phenomenon \citep{cohen2021gradient}.
Under AT, sharpness also initially increases, but subsequently
oscillates with an overall decreasing trend. Opaque curves show smoothed trends for clarity, while semi-transparent curves show the raw values.}
    \label{fig:sharpness_1}
\end{figure*}

In this section, we study how the \ref{equ:advloss} evolves under \ref{equ:gd} with arbitrary constant step sizes. It is well-known that gradient descent dynamics with $\eta > 2/L$ may exhibit \textit{non-monotonic} loss behavior, where $L$ is the Lipschitz constant of the gradient. Recent work of \citet{wu2024large} characterize these non-monotonic loss behavior in standard linear logistic regression ($c=0$).  We informally restate their main results below using our notation:

\begin{thm}[\cite{wu2024large}]\label{thm:wure}
Under Assumption~\ref{ass1} with $c=0$, and normalize the dataset such that $\lVert \mathbf x_i \rVert \le 1$ for $i \in [N]$. For standard logistic regression, gradient descent with a constant step size $\eta>0$ satisfies:
\begin{itemize}[leftmargin=*]
   \item (Phase transition time)
   There exists some phase transition time $\tau$, depending on the step size $\eta$ and the dataset, before which the loss can be non-monotonic and after which $\CL(\mathbf w_{t+1})\le\CL(\mathbf w_t)$ for all $t\ge\tau$.
    \item (Accelerated Convergence Rate)
    For a sufficiently large time budget $T$, an appropriate constant step size $\eta=\Theta(T)$ yields a upper bound  $\CL(\mathbf w_T) = \tilde{\mathcal O}(1/T^2)$.
\end{itemize}
\end{thm}
The work of \citet{ji2018risk} also established a  $\tilde O(1/T)$ loss bound for every fixed step size $\eta\in(0,1]$ for ST. These results nature questions for AT: (i) does there also exist a phase transition time? (ii) can any constant step size, possibly tuned to the horizon $T$, that guarantees a fast vanishing loss bound? The following theorem shows that the answer is negative.
\begin{thm}\label{prop:counter_exp_opt}
For any  $0<c<2$ and time budget $T>64/c$, there exists a dataset $\mathcal{D}_T$  whose dimension depends on $T$ and satisfies Assumption~\ref{ass1}, such that for any step size $\eta>0$, gradient descent dynamics on $\mathcal{D}_T$ satisfies 
\begin{itemize}[leftmargin=*]
\item (Uniform Constant Lower Bound) The loss function has a uniform constant lower bound $\mathcal{L}_{\mathrm{Adv}}(\mathbf{w}_T) \ge \mathcal{C}$, where $\mathcal{C}$ only depends on $c$. 
\item (No phase transition time) When gradient descent is run indefinitely on $\mathcal{D}_T$,
there exist infinitely many $t$ such that
$\mathcal{L}_{\mathrm{Adv}}(\mathbf{w}_{t+1})
> \mathcal{L}_{\mathrm{Adv}}(\mathbf{w}_t)$.
\end{itemize}
\end{thm}
We have two remarks. First, for ST, Theorem~\ref{thm:wure} shows that an appropriately chosen large constant step size can yield a polynomial acceleration of the loss convergence rate, from the $1/T$ scale to $\widetilde{\mathcal O}(1/T^2)$. In contrast, Theorem~\ref{prop:counter_exp_opt} shows that the convergence is intrinsically non-uniform over the problem class: for every $T$, there exists a problem for which the optimization error at time $T$ is bounded below by a positive real for every constant step size. Note that there is no contradiction with eq. ~\eqref{eq:ub_rate}, which instead ensures an instance dependent convergence. Second, Theorem~\ref{thm:wure} shows that, in standard training, there exists a transition time before which the loss can be non-monotonic and after which it decreases monotonically. Moreover, \citet{wu2024large} show that this non-monotonic phase is essential for the acceleration obtained with large step sizes. In contrast, Theorem~\ref{prop:counter_exp_opt} shows that in AT such a transition time need not exist: the loss may increase infinitely often. Notably, these increases occur even between differentiable iterates, so the persistent non-monotonicity is not an artifact of evaluating subgradients at nonsmooth points.



\paragraph{Proof Overview of Theorem~\ref{prop:counter_exp_opt}.}We construct a single sample dataset $\mathcal D_T=\{(\mathbf x,1)\}$, where
\[
\mathbf x=(2,q^{-1},\ldots,q^{-1})\in\mathbb R^{1+q^2},
\qquad q=4T^2.
\]
We have $\lVert\mathbf x \rVert_2 = \sqrt{5}$. This data has one dominant coordinate and $T^4$
small coordinates whose magnitudes are below the adversarial radius. It can be seen that $\mathcal{D}_T$ is linearly robustly separable for $c<2$ with robust max-margin direction $\mathbf{u}_{2,\infty}(c)=(1,0,\ldots,0)$ and $\gamma_{2,\infty}(c) = 2-c$.
The nonsmooth $\ell_1$ term in the adversarial loss 
causes the last $q^2$ small coordinates of $\mathbf w_t$ to oscillate around zero, which makes the gradient descent dynamics unstable since it depends on the \textit{sign} of these small coordinates. Large step sizes can amplify these oscillations, whose combined
contribution to the loss outweighs the improvement along the
dominant coordinate. More precisely, consider the robust margin 
$\min_{\|\boldsymbol\delta\|_\infty\le c}
\langle\mathbf x+\boldsymbol\delta,\mathbf w_t\rangle$.
For sufficiently large step sizes,
the first coordinate of $\mathbf x$ contributes at most $4\eta T$ to the robust margin,
whereas the remaining coordinates contribute at most
$-\eta cT^2/2$ in total. Thus the negative
contribution dominates, keeping the adversarial loss large
throughout the time budget. The detailed proof of Theorem~\ref{prop:counter_exp_opt} is presented
in Appendix~\ref{app_s4}. 

\begin{figure*}[t]
    \centering
    \includegraphics[width=\textwidth]{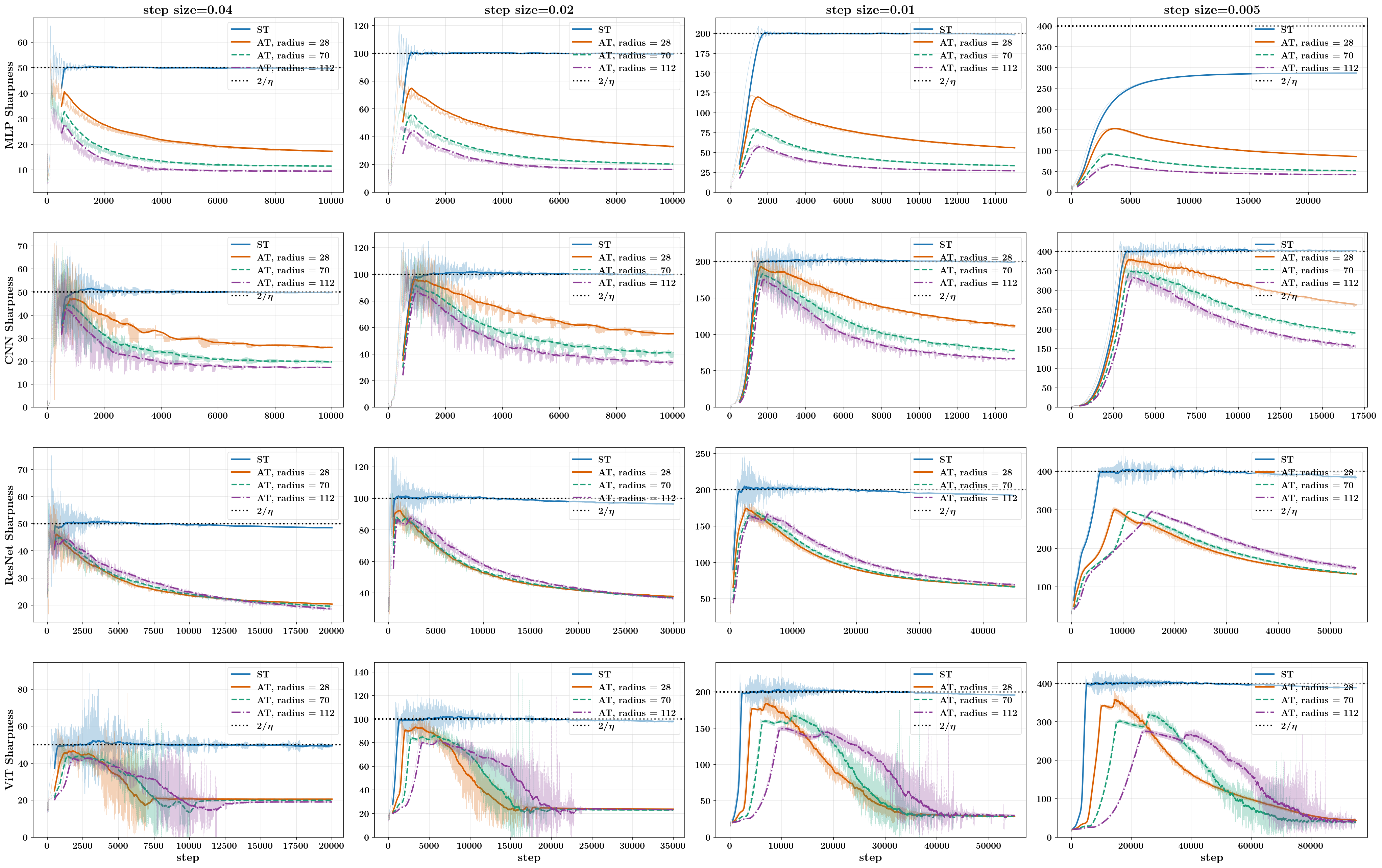}
    \caption{The experimental settings are identical to those in Figure~\ref{fig:sharpness_1} except that adversarial examples are generated by $\boldsymbol{\ell_2}$ attacks, and adversarial radius is adjusted per  common practice. }
    \label{fig:sharpness_2}
\end{figure*}

\section{Evolution of Sharpness in Adversarial Training}\label{sec:eos}

In this section, we study how the \ref{equ:adv_sharp} evolve along the adversarial training trajectories. Sharpness, i.e., the maximal eigenvalue of the loss Hessian, plays an important role in neural network training, from guiding the design of optimization algorithms \citep{foret2020sharpness} to improve understanding of generalization  \citep{andriushchenko2023modern}. The classical descent lemma for GD in optimization problem \(\min_{\theta} f(\theta)\) states the loss decreases monotonically under GD if the sharpness remains below \(2/\eta\) along the optimization process, where \(\eta\) is the step size \citep{nesterov2013introductory}. However, in neural networks training, people usually employ large step sizes in the regime where the loss exhibits non-monotonic behavior.
\citet{cohen2021gradient} observe that sharpness can evolve in two stages during \textit{standard training} of neural network with deterministic gradient descent:
\begin{itemize}[leftmargin=*]
\item {\textit{Progressive Sharpening}}: If the sharpness is less than $2/\eta$, it tends to increase monotonically.
\item {\textit{Edge-of-stability}}: The sharpness oscillate around $2/\eta$.
\end{itemize}
It is natural to ask whether AT of neural networks exhibits a similar sharpness evolution to standard training. Our experiments reveal a different pattern: after an initial increase similar to progressive sharpening, adversarial sharpness oscillates with an overall decreasing trend. We report these experimental results in Section~\ref{sec:er}. In Section~\ref{sec:tiome}, we provide theoretical support by showing that adversarial training can reduce sharpness in an adversarial variant of a minimalist example introduced by \citet{ahn2023learning}.

\subsection{Experimental Results}\label{sec:er}
In Figure \ref{fig:sharpness_1} and Figure \ref{fig:sharpness_2}, we present our experiments on the sharpness evolution of AT under MLP, CNN, ResNet and Vision Transformer (ViT) on the CIFAR-10 dataset. The training algorithm is the deterministic gradient descent on CIFAR-10. We adopt the architectures and implementation of \citet{cohen2021gradient}, keeping our setup consistent with prior work on edge-of-stability. We also use the \textit{Mean Squared Error} (MSE) loss as it can most align with the edge-of-stability phenomena in ST. The experimental details are provided in Appendix~\ref{app:aes1}. Our main observation is that the sharpness evolution of AT follows a very different two-stage dynamics as with ST:
\begin{itemize}[leftmargin=*]
\item {\textit{Stage $\mathrm{I}$}}: In this stage, sharpness tends to increase monotonically, similar to the progressive sharpening stage in standard training reported by \citet{cohen2021gradient}. 
\item {\textit{Stage $\mathrm{II}$}}: After \textit{Stage $\mathrm{I}$}, sharpness begins to oscillate. Although the oscillation patterns may vary across different settings, they share a common feature: \ref{equ:adv_sharp} exhibits an overall \textbf{decreasing trend over longer timescales}. 
\end{itemize}
\begin{figure}[t]
    \centering
    \includegraphics[width=\columnwidth]{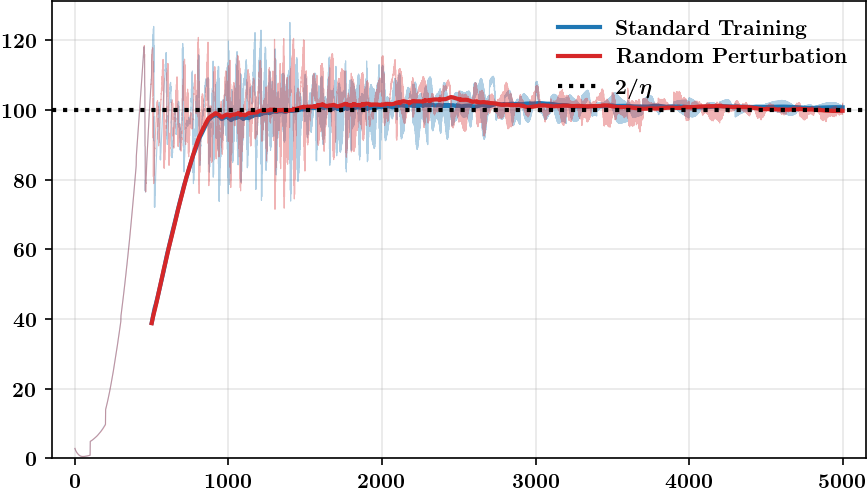}
    \caption{Sharpness evolution under randomly perturbed losses resembles that of standard training.}
    \label{fig:randomp1}
\end{figure}
More experiments are reported in Appendix~\ref{app:aes2}, and the experimental results are consistent with the two-stage dynamics as described above. One major property of the sharpness in AT is that it tracks the curvature of a time-varying loss function, as the adversarial examples $\mathbf{x}_i'$ change at each iteration (see Section~\ref{sec:gat}). However, our experiments suggest that time variation alone does not explain the observed dynamics. In Figure~\ref{fig:randomp1}, we track sharpness along standard CNN training trajectories using loss functions with small random perturbations instead of adversarial attack on the training data at each iteration. The resulting sharpness evolution remains similar to that of standard training and differs markedly from that of adversarial training. Together, these results suggest that the observed two-stage behavior stem inherently from the adversarial training dynamics and persist across different experimental settings. 

\subsection{Theoretical Investigation on a Minimalist Example}\label{sec:tiome}
Due to the complexity of neural network training dynamics, theoretical studies of EoS often adopt a minimalist example approach. In the following, we study an adversarial variant of a  model introduced by \citet{ahn2023learning} to investigate the EoS phenomenon in standard training. Let $\ell(s)$ be the Huber loss function:
\begin{align}\label{equ:huber_loss}
\ell(s) := \frac{s^2}{2}\mathbf{1}_{\{|s|\le 1\}} + \left(|s|-\frac{1}{2}\right)\mathbf{1}_{\{|s|>1\}},    \tag{Huber loss}
\end{align}
which is widely used for regression \citep{meyer2021alternative}. The theoretical model proposed by \citet{ahn2023learning} is the two-dimensional function
$\ell(xy)$ for $(x,y)\in\mathbb{R}^2$. One main reason for the difference between the sharpness behavior of ST and AT is is the special min-max structure of AT. Motivated by this observation, we study the following adversarial variant of $\ell(xy)$ for an adversarial radius $c \ge 0$, which recovers the the standard training objective of \citet{ahn2023learning} when $c=0$:
\begin{align}\label{equ:advhuber}
\min_{(x,y) \in \BR^2}\CL_{c} (x,y) =\min_{(x,y) \in \BR^2} \max_{|\delta| \le c} \ell\left((x+\delta)y\right)\tag{Adversarial Huber loss}
\end{align}
This objective can be viewed as adversarial regression on a single example with input \(z=0\) and target \(0\), using the predictor \(f_{x,y}(z)=y(x+z)\). \citet{ahn2023learning} studied the following gradient descent dynamic when $c=0$:
\begin{align}
&(x_{t+1},y_{t+1}) =(x_{t},y_{t})  - \eta\, \partial\CL_{c}(x_t,y_t),\quad \mathrm{with}\nonumber\\
&(x_0,y_0) = \sqrt{(2+\delta)/\eta}\,(\tilde{x},\tilde{y}).\tag{EoS initialization}\label{equ:eos_reg}
\end{align}
Here $\delta>0, \, \tilde{y} > \tilde{x}>0$ and $ \tilde{y}^2 - \tilde{x}^2 = 1$. This initialization condition distinguishes gradient descent from continuous-time gradient flow. The sharpness of \eqref{equ:advhuber} at the $t$-th iteration of gradient descent is conventionally defined as the maximum eigenvalue of the Hessian, given by
$\lambda^{c}_t:= \lambda_{\max}\left( \nabla^2 \CL_{c}(x_t,y_t) \right)$. 

\begin{prop}\label{prop:stand_adv}
Consider the \ref{equ:gd} with \ref{equ:eos_reg}. There exist constants $\eta_0>0$,
depending only on $\delta,\tilde x,\tilde y$ and $c$
such that under mild boundary-avoidance assumption as described in
Assumption~\ref{ass:bmbm}, for every $0<\eta\le\eta_0$, the limiting sharpness
$\lambda_\infty^c:=\limsup_{t\to\infty}\lambda_t^c$ satisfies:
\begin{itemize}[leftmargin=*]
    \item \textit{Standard training}
    \citep[Theorem~2]{ahn2023learning}:
    if $c=0$, then
    $
    \frac{2}{\eta}-\mathcal{O}(1)
    \le \lambda_\infty^0 \le \frac{2}{\eta}.
    $
    \item \textit{Adversarial training}:
    if $c >0 $, then
    $
    \lambda_\infty^c = \mathcal{O}(\eta + c^2).
    $
    \end{itemize}
\end{prop}



\noindent
Proposition~\ref{prop:stand_adv} reveals a transition in sharpness evolution between $\ell(xy)$ and its min-max variant. The adversarial dynamics converge to points with sharpness $\mathcal{O}(\eta+c^2)$, much smaller than the edge-of-stability level $2/\eta$ for $\eta\,, c \ll 1$, which is typical in practice. This is consistent with the experimental results reported in Section~\ref{sec:er}. The difference arises from changes in both the structure of the minimizer set and the training dynamics induced by adversarial perturbations. When $c=0$, all points $(x,y)$ satisfying $xy=0$ are global minimizers, and the trajectory converges to $(0,y_\infty)$ with $y_\infty\neq0$ and sharpness close to $2/\eta$. When $c>0$, the minimizer set remains large: every point $(x,0)$ is a global minimizer. The sharpness at such a point is $(|x|+c)^2$, which can be arbitrarily large. However, the training dynamics ensure that the limiting point satisfies $x_\infty=\mathcal{O}(\sqrt{\eta})$. The detailed proof is given in Appendix~\ref{app_3}.


\section{Conclusion and Future Direction}
In this paper, we studied the dynamics of deterministic gradient descent with large step sizes in adversarial training, focusing on three aspects: implicit bias and optimization in adversarial logistic regression, and sharpness evolution in adversarial neural network training. Our results reveal that the effects of large step sizes in adversarial training can differ substantially from those in standard training. An interesting direction for future work is to find a general mechanism, possibly analogous to the self-stabilization mechanism in standard training \citep{damian2022self}, that explains the decreasing trend in sharpness observed in Section~\ref{sec:eos}.

\bibliographystyle{apalike}
\bibliography{Tex/Bibli}

\newcommand{\checklisttodo}[1]{\textcolor{red}{[TODO: #1]}}

\clearpage
\appendix
\thispagestyle{empty}
\onecolumn
\aistatstitle{Adversarial Training with Large Step Sizes:\\
Implicit Bias, Optimization, and Sharpness\\
Supplementary Materials}

\section{More backgrounds}

In the following, without loss of generality, we assume $y_i = 1$ for all $i \in [N]$ in what follows since if $y_i = -1$, we can equivalently take the negative of $ \mathbf{x}_i$ as input. We introduce the following decomposition of $\BR^d$, which was motivated by \cite{wu2023implicit}:
\begin{align}
\CF:  \BR^d &\to \BR^d \times \BR^{d}\nonumber \\
 \mathbf{x} &\to \left(\langle \mathbf{x},\mathbf{u}_{2,\infty}(c) \rangle \mathbf{u}_{2,\infty}(c), \mathrm{Proj}_{\mathbf{u}^{\perp}_{2,\infty}(c)}(\mathbf{x})\right).
\end{align}
We write 
\begin{align}\label{equ:decwur}
\mathbf{w}_t = v_t \mathbf{u}_{2,\infty}(c) + \mathbf{r}_t,    
\end{align}
where $\langle\mathbf{r}_t, \mathbf{u}_{2,\infty}(c) \rangle = 0$. We have
\begin{align*}
-\partial_{ \mathbf{w}}  \CL_{\mathrm{Adv}}(\mathbf{w}_t) & =  - \sum^N_{i=1} \partial_{\mathbf{w}}\ln\left(1 +\exp(-\mathbf{x}^{\top}_i\mathbf{w}_t + c \lVert \mathbf{w}_t \rVert_1)  \right) \\
& = \sum^N_{i=1} \lambda_{i,t} \left(  \mathbf{x}_i - c \,\mathbf{s}_t  \right),
\end{align*}
where 
\begin{align}\label{equ:deflambda}
\mathbf{s}_t  = (s_{t,j})^d_{i=1} \in \partial \lVert \mathbf{w}_t \rVert_1, \quad
\lambda_{i,t} = \frac{\exp(-\mathbf{x}^{\top}_i\mathbf{w}_t + c \lVert \mathbf{w}_t \rVert_1)}{1+ \exp(-\mathbf{x}^{\top}_i\mathbf{w}_t + c \lVert \mathbf{w}_t \rVert_1)} = \frac{1}{1+ \exp(\mathbf{x}^{\top}_i\mathbf{w}_t - c \lVert \mathbf{w}_t \rVert_1)} >0.
\end{align} 

Thus the \eqref{equ:gd} dynamics can be written as
\begin{align}\label{equ:deflambda2}
\mathbf{w}_{t+1} = \mathbf{w}_{t} +\eta \sum^N_{i=1} \lambda_{i,t} \left(  \mathbf{x}_i - c\,\mathbf{s}_t  \right)
\end{align}
\subsection{Some useful Lemmas}\label{app_lemma}

\begin{lem}\label{lem:expl}
Assume the dataset $\{(\mathbf{x}_i, y_i)\}_{i=1}^N$ is linearly separable, i.e., $\exists\ \mathbf{w} \in \mathbb{R}^d$ such that $y_i \mathbf{w}^\top \mathbf{x}_i > 0$ for all $i \in [N]$. If $c < \gamma_{\infty}$, we have $\gamma_{2,\infty}(c) > 0$.
\end{lem}

\begin{proof}
Let $\mathbf{u}_{\infty} = \argmax_{\lVert \mathbf{w} \rVert_1=1} \min_{i \in [n]} \mathbf{x}^{\top}_i\mathbf{w}$. By definition, we have $\gamma_{\infty} = \min_{i \in [N]}\mathbf{x}^{\top}_i\mathbf{u}_{\infty}$.

Moreover, we have
\begin{align*}
\gamma_{2,\infty}(c) & \ge \min_{i \in [N]}\min_{\lVert \delta_i\rVert_{\infty}\le c} \left(\mathbf{x}_i+\delta_i\right)^{\top}\mathbf{u}_{\infty}   \\
& = \min_{i \in [N]} \mathbf{x}^{\top}_i \mathbf{u}_{\infty} - c\lVert \mathbf{u}_{\infty} \rVert_{1}\\
& = \gamma_{\infty} - c \qquad ( \mathrm{Since} \min_{i \in [N]} \mathbf{x}^{\top}_i \mathbf{u}_{\infty} = \gamma_{\infty}\ \mathrm{and} \ \lVert \mathbf{u}_{\infty} \rVert_{1}=1)\\
& > 0.
\end{align*}
This completes the proof.
\end{proof}

\begin{lem}
\label{lem:one-sample-robust-margin}
Let \(D=\{(\mathbf{x},1)\}\) be a one-sample dataset and let
\(0\leq c<\|\mathbf{x}\|_{\infty}\). Then for this data set, we have
\[
\begin{aligned}
\gamma_{2,\infty}(c)
=\left\|
   \bigl(|\mathbf{x}|-c\mathbf{1}\bigr)_+
   \right\|_2,
\end{aligned}
\]
where the absolute value and positive part are applied
coordinate wise. 
\end{lem}

\begin{proof}
For every \(\mathbf{u}\in\mathbb{R}^d\), we have
\[
\min_{\|\boldsymbol{\delta}\|_{\infty}\leq c}
\left\langle
\mathbf{x}+\boldsymbol{\delta},\mathbf{u}
\right\rangle
=
\mathbf{x}^{\top}\mathbf{u}-c\|\mathbf{u}\|_1,
\]
where the minimum is attained by
\(\delta_j=-c\operatorname{sign}(u_j)\). Therefore,
\[
\begin{aligned}
\mathbf{x}^{\top}\mathbf{u}-c\|\mathbf{u}\|_1
\leq
\sum_{j=1}^d (|x_j|-c)|u_j| 
\leq
\sum_{j=1}^d (|x_j|-c)_+|u_j|
&\leq
\left\|(|\mathbf{x}|-c\mathbf{1})_+\right\|_2
\|\mathbf{u}\|_2,
\end{aligned}
\]
where the final inequality uses the Cauchy-Schwarz inequality. For \(\|\mathbf{u}\|_2=1\), the last expression equals
\(\|(|\mathbf{x}|-c\mathbf{1})_+\|_2\).
Since \(c<\|\mathbf{x}\|_{\infty}\), this quantity is positive,
and equality is attained by \(\mathbf{u}=\mathbf{u}^*\), where
\[
u_j^*
=
\frac{
\operatorname{sign}(x_j)(|x_j|-c)_+
}{
\left\|(|\mathbf{x}|-c\mathbf{1})_+\right\|_2
},
\qquad j\in[d].
\]
\end{proof}

\subsection{\texorpdfstring{$\ell_\infty$}{l-infinity}-Robust
Support Vector Machine}\label{sec:rsvm}
The $\ell_{\infty}$-robust support vector machine (SVM) problem can be formulated as
\begin{align}\label{equ:rsvm}
\min_{\mathbf{w} \in \BR^d} \frac{1}{2} \lVert \mathbf{w}\rVert^2_2,\qquad \mathrm{such\ that}\ \min_{\lVert \boldsymbol\delta_i \rVert_{\infty} \le c} \langle \mathbf{x}_i+\boldsymbol{\delta}_i,\mathbf{w}\rangle \ge 1  \tag{$\ell_{\infty}$-Robust SVM}
\end{align}
Let $\hat{\mathbf{w}}(c)$ be the solution of \eqref{equ:rsvm}. We have
$
\mathbf{u}_{2,\infty} = \hat{\mathbf{w}}(c)/\lVert \hat{\mathbf{w}}(c)\rVert_2,\ \mathrm{and}\ \gamma_{2,\infty}(c) = 1/\lVert\hat{\mathbf{w}}(c)\rVert_2.
$ We use $\mathbf{z}_i(c)$ to denote the worst-case adversarial perturbation that an adversary can construct on $\mathbf{x}_i$ against the classifier parameterized by $\hat{\mathbf{w}}(c)$: 
\begin{align}\label{equ:zic}
\mathbf{z}_i(c) := \mathbf{x}_i - c\, \mathbf{s}^*,\quad  \mathrm{where}\quad \mathbf{s}^* \in \partial \lVert \hat{\mathbf{w}}(c) \rVert_1.
\end{align}
We choose $\mathbf s^*$ jointly with the KKT multipliers so that the representation in Proposition~\ref{prop:dualrsvm} below holds, and fix this choice throughout the subsequent analysis. The adversarial support vectors are defined as:
\begin{align}\label{equ:asv}
\CS(c) := \{ i \in [N] \mid \langle \mathbf{z}_i(c),\mathbf{u}_{2,\infty}(c) \rangle = \gamma_{2,\infty}(c)  \}  \tag{Adversarial Support Vectors}
\end{align}

\begin{prop}\label{prop:dualrsvm}[\citet{boyd2004convex}]
There exists a subgradient $\mathbf{s}^* \in \partial \lVert \hat{\mathbf{w}}(c) \rVert_1$, such that we have $\hat{\mathbf{w}}(c) = \sum_{i \in \CS(c)} \alpha_i \mathbf{z}_i(c)$, and $\alpha_i \ge 0$.
\end{prop}
The main property we need for the \eqref{equ:rsvm} is the following:

\begin{lem}\label{lem:ib_kkt}
For every $\mathbf{r}\perp\mathbf{u}_{2,\infty}(c)$, there exists a
$j\in\CS(c)$ with $\alpha_j>0$ such that
$\langle\mathbf{z}_j(c),\mathbf{r}\rangle\le0$.    
\end{lem}

\begin{proof}
For $\mathbf{r}\perp\mathbf{u}_{2,\infty}(c)$, we have
\begin{align*}
0=\frac{\langle\mathbf{u}_{2,\infty}(c),\mathbf{r}\rangle}
{\gamma_{2,\infty}(c)}
=\langle\hat{\mathbf{w}}(c),\mathbf{r}\rangle
=\sum_{i\in\CS(c)}\alpha_i
\langle\mathbf{z}_i(c),\mathbf{r}\rangle.
\end{align*}
If every inner product with $\alpha_i>0$ were positive, then
\begin{align*}
\sum_{i\in\CS(c)}\alpha_i
\langle\mathbf{z}_i(c),\mathbf{r}\rangle
=\sum_{i\in\CS(c):\,\alpha_i>0}\alpha_i
\langle\mathbf{z}_i(c),\mathbf{r}\rangle>0,
\end{align*}
contradicting the preceding equality. This finished the proof.
\end{proof}

\section{Proof of Section~\ref{sec:ib}}\label{app:ptmain1}

\subsection{Preliminary estimate on the convergence rate of implicit bias}

\begin{lem} Let $M=\max_{i\in[N]}\lVert\mathbf{x}_i\rVert_2+c\sqrt d$, then we have
\begin{align}\label{equ:ib_steps_sample_norm}
\lVert\mathbf{x}_i-c\mathbf{s}_t\rVert_2
\le M.
\end{align}
and
\begin{align}
\lVert\mathbf{w}_t\rVert_2
\le\eta NMt.
\label{equ:ib_steps_iterate_norm}
\end{align}
\end{lem}
\begin{proof}
Since the norm of subgradient satisfies $|s_{t,j}|\le 1$, we have 
\begin{align}
\lVert\mathbf{x}_i-c\mathbf{s}_t\rVert_2
&\le\lVert\mathbf{x}_i\rVert_2+c\lVert\mathbf{s}_t\rVert_2
\le\lVert\mathbf{x}_i\rVert_2+c\sqrt d\le M.\nonumber
\end{align}
Moreover, 
\begin{align}
\lVert\mathbf{w}_t\rVert_2
&=\left\lVert\sum_{s=0}^{t-1}(\mathbf{w}_{s+1}-\mathbf{w}_s)\right\rVert_2
\le\sum_{s=0}^{t-1}\lVert\mathbf{w}_{s+1}-\mathbf{w}_s\rVert_2
\le\eta NMt.\nonumber
\end{align}
\end{proof}

\begin{prop}[Growth of $v_t$]\label{lem:trainingproj}
For every $t\ge0$, we have
\begin{align*}
v_{t+1}-v_t
\ge\eta\gamma_{2,\infty}(c)\sum_{i=1}^N\lambda_{i,t}
\ge\frac{\eta\gamma_{2,\infty}(c)}2e^{-\gamma_{2,\infty}(c)v_t}.
\end{align*}
In particular, above recursion gives $v_t>0$, and
\begin{align}
v_t&\ge\frac1{\gamma_{2,\infty}(c)}
\ln\left(1+\frac{\eta\gamma_{2,\infty}^2(c)}2t\right),\qquad
e^{-\gamma_{2,\infty}(c)v_t}
\le\frac1{1+\frac{\eta\gamma_{2,\infty}^2(c)}2t}
\le\frac{C_1}{t+1}.\label{equ:ib_v_lower}
\end{align}
where $C_1 = \max\left\{1,\frac{2}{\eta\gamma_{2,\infty}^2(c)}\right\}$, which is independent of $t$.
\end{prop}
\begin{proof} Recall in \eqref{equ:deflambda}, we define $\mathbf{s}_t  = (s_{t,j})^d_{i=1} \in \partial \lVert \mathbf{w}_t \rVert_1$. Since $|s_{t,j}|\le1$, we have
\begin{align}
\langle\mathbf{s}_t,\mathbf{u}_{2,\infty}(c)\rangle
=\sum_{j=1}^d s_{t,j}u_{2,\infty,j}(c)
\le\sum_{j=1}^d|s_{t,j}|\,|u_{2,\infty,j}(c)|
\le\sum_{j=1}^d|u_{2,\infty,j}(c)|
=\lVert\mathbf{u}_{2,\infty}(c)\rVert_1.\label{equ:ib_steps_subgradient_projection}
\end{align}
For every $i$, we have
\begin{align}
\langle\mathbf{x}_i-c\mathbf{s}_t,\mathbf{u}_{2,\infty}(c)\rangle
&=\mathbf{x}_i^\top\mathbf{u}_{2,\infty}(c)
-c\langle\mathbf{s}_t,\mathbf{u}_{2,\infty}(c)\rangle\nonumber\\
\nonumber\\
&\overset{\eqref{equ:ib_steps_subgradient_projection}}{\ge}\mathbf{x}_i^\top\mathbf{u}_{2,\infty}(c)
-c\lVert\mathbf{u}_{2,\infty}(c)\rVert_1\nonumber\\
\nonumber\\
&\ge\min_{k\in[N]}
\bigl[\mathbf{x}_k^\top\mathbf{u}_{2,\infty}(c)
-c\lVert\mathbf{u}_{2,\infty}(c)\rVert_1\bigr]
=\gamma_{2,\infty}(c).\label{equ:ib_projection_bound}
\end{align}
Projecting the update into the direction of $\mathbf{u}_{2,\infty}$ yields
\begin{align}
v_{t+1}-v_t
&=\langle\mathbf{w}_{t+1}-\mathbf{w}_t,\mathbf{u}_{2,\infty}(c)\rangle\\
&=\eta\sum_{i=1}^N\lambda_{i,t}
\langle\mathbf{x}_i-c\mathbf{s}_t,\mathbf{u}_{2,\infty}(c)\rangle\nonumber\\
&\overset{\eqref{equ:ib_projection_bound}}{\ge}\eta\sum_{i=1}^N\lambda_{i,t}\gamma_{2,\infty}(c)
=\eta\gamma_{2,\infty}(c)\sum_{i=1}^N\lambda_{i,t}>0.\label{equ:vt}
\end{align}
Since $v_0=0$, \eqref{equ:vt} gives $v_t>0$ for $t\ge1$.

Apply Lemma~\ref{lem:ib_kkt} to $\mathbf{r}_t$ and choose its index $j$ according to Lemma~\ref{lem:ib_kkt}. Recall we define $\mathbf{s}^* \in \partial \lVert \hat{\mathbf{w}}(c) \rVert_1$ in \eqref{equ:zic}. The same coordinatewise argument gives
\begin{align}
\langle\mathbf{s}^{*},\mathbf{w}_t\rangle
\le\sum_{k=1}^d|s_k^{*}|\,|w_{t,k}|
\le\lVert\mathbf{w}_t\rVert_1.\label{equ:ib_steps_fixed_subgradient_norm}
\end{align}
Consequently,
\begin{align}
\mathbf{x}_j^\top\mathbf{w}_t-c\lVert\mathbf{w}_t\rVert_1
&\overset{\eqref{equ:ib_steps_fixed_subgradient_norm}}{\le}\mathbf{x}_j^\top\mathbf{w}_t
-c\langle\mathbf{s}^{*},\mathbf{w}_t\rangle
=\langle\mathbf{z}_j(c),\mathbf{w}_t\rangle\nonumber\\
&=v_t\langle\mathbf{z}_j(c),\mathbf{u}_{2,\infty}(c)\rangle
+\langle\mathbf{z}_j(c),\mathbf{r}_t\rangle\nonumber\\
&=\gamma_{2,\infty}(c)v_t
+\langle\mathbf{z}_j(c),\mathbf{r}_t\rangle\nonumber \\
& \le\gamma_{2,\infty}(c)v_t.\label{equ:ib_one_small_score} \quad (\mathrm{Since\,Lemma\,} \eqref{lem:ib_kkt})
\end{align}
Hence
\begin{align}
&1+\exp(\mathbf{x}_j^\top\mathbf{w}_t-c\lVert\mathbf{w}_t\rVert_1)
\le 1+\exp(\gamma_{2,\infty}(c)v_t),\nonumber\\
&\lambda_{j,t}
= \frac1{1+\exp(\mathbf{x}_j^\top\mathbf{w}_t-c\lVert\mathbf{w}_t\rVert_1)}
\ge \frac1{1+\exp(\gamma_{2,\infty}(c)v_t)}.\label{equ:ib_steps_lambda_reciprocal}
\end{align}
Since $\gamma_{2,\infty}(c)v_t\ge0$, we have
\begin{align}
& \frac1{1+\exp(\gamma_{2,\infty}(c)v_t)}
=\frac{\exp(-\gamma_{2,\infty}(c)v_t)}
{1+\exp(-\gamma_{2,\infty}(c)v_t)}
\ge\frac12\exp(-\gamma_{2,\infty}(c)v_t),\label{equ:ib_steps_half_exponential}\\
\nonumber\\
& v_{t+1}-v_t
\overset{\eqref{equ:vt}}{\ge}\eta\gamma_{2,\infty}(c)\sum_{i=1}^N\lambda_{i,t}
\ge\eta\gamma_{2,\infty}(c)\lambda_{j,t}
\overset{\eqref{equ:ib_steps_lambda_reciprocal},\eqref{equ:ib_steps_half_exponential}}{\ge}\frac{\eta\gamma_{2,\infty}(c)}2
\exp(-\gamma_{2,\infty}(c)v_t).\label{equ:ib_steps_v_increment_substitution}
\end{align}
For $x\ge0$, we have the following general inequality:
\begin{align}
e^x-1\ge x.\label{equ:ib_scalar_exp_lower}
\end{align}
Applying this
with $x=\gamma_{2,\infty}(c)(v_{t+1}-v_t)\ge0$ gives
\begin{align}
e^{\gamma_{2,\infty}(c)v_{t+1}}-e^{\gamma_{2,\infty}(c)v_t}
&=e^{\gamma_{2,\infty}(c)v_t}
\left[e^{\gamma_{2,\infty}(c)(v_{t+1}-v_t)}-1\right]\nonumber\\
&\overset{\eqref{equ:ib_scalar_exp_lower}}{\ge} e^{\gamma_{2,\infty}(c)v_t}\gamma_{2,\infty}(c)(v_{t+1}-v_t)\nonumber\\
&\overset{\eqref{equ:ib_steps_v_increment_substitution}}{\ge} e^{\gamma_{2,\infty}(c)v_t}\gamma_{2,\infty}(c)
\frac{\eta\gamma_{2,\infty}(c)}2e^{-\gamma_{2,\infty}(c)v_t}
=\frac{\eta\gamma_{2,\infty}^2(c)}2.\label{equ:ib_longitudinal_increment}
\end{align}
Summing the increments from $s=0$ to $s=t-1$ gives
\begin{align}
& e^{\gamma_{2,\infty}(c)v_t}
=e^{\gamma_{2,\infty}(c)v_0}
+\sum_{s=0}^{t-1}\left[e^{\gamma_{2,\infty}(c)v_{s+1}}
-e^{\gamma_{2,\infty}(c)v_s}\right]
\overset{\eqref{equ:ib_longitudinal_increment}}{\ge}1+\frac{\eta\gamma_{2,\infty}^2(c)}2t,\label{equ:ib_steps_v_telescope}\\
\end{align}
which implies
\begin{align}
&v_t \ge \frac1{\gamma_{2,\infty}(c)}
\ln\left(1+\frac{\eta\gamma_{2,\infty}^2(c)}2t\right), \qquad e^{-\gamma_{2,\infty}(c)v_t}
\le \frac1{1+\frac{\eta\gamma_{2,\infty}^2(c)}2t}.\nonumber
\end{align}
Finally, we have
\begin{align*}
1+\frac{\eta\gamma_{2,\infty}^2(c)}2t
&\ge\min\left\{1,\frac{\eta\gamma_{2,\infty}^2(c)}2\right\}(t+1)
= \frac{t+1}{C_1}.
\end{align*}
This proves all the stated bounds.
\end{proof}

\begin{lem}[Preliminary estimates]\label{lem:ib_preliminary}
Let
\begin{align}
\tilde{\mathbf{w}}_t=\hat{\mathbf{w}}(c)
\ln\left(1+\frac{\eta\gamma_{2,\infty}^2(c)}2t\right).\label{equ:ib_det_tilde}
\end{align}
There are fixed $C_2,C_3,C_4>0$ such that, for all $t\ge0$,
\begin{align*}
\lVert\mathbf{r}_t\rVert_2
&\le C_2\sqrt{\ln(t+2)},\qquad
\sum_{i=1}^N\lambda_{i,t}\le C_3(t+1)^{-3/4},\qquad
\sum_{s=0}^{\infty}\lVert\mathbf{w}_{s+1}-\mathbf{w}_s\rVert_2^2\le C_4.
\end{align*}
Moreover, $\CL_{\mathrm{Adv}}(\mathbf{w}_t)\to0$ as $t \to \infty$.
\end{lem}
\begin{proof}
Since $\hat{\mathbf{w}}(c)=\mathbf{u}_{2,\infty}(c)/\gamma_{2,\infty}(c)$, we have
\begin{align}
\langle\mathbf{w}_{t+1}-\tilde{\mathbf{w}}_{t+1},
\hat{\mathbf{w}}(c)\rangle & = \frac{v_{t+1}}{\gamma_{2,\infty}(c)}
-\frac1{\gamma_{2,\infty}^2(c)}
\ln\left(1+\frac{\eta\gamma_{2,\infty}^2(c)}2(t+1)\right)\nonumber\\
&\quad=\frac1{\gamma_{2,\infty}(c)}
\left[v_{t+1}-\frac1{\gamma_{2,\infty}(c)}
\ln\left(1+\frac{\eta\gamma_{2,\infty}^2(c)}2(t+1)\right)\right]\overset{\eqref{equ:ib_v_lower}}{\ge}0.\label{equ:ib_steps_comparator_projection}
\end{align}
Moreover, we have
\begin{align*}
\tilde{\mathbf{w}}_{t+1}-\tilde{\mathbf{w}}_t
 = \hat{\mathbf{w}}(c)
\ln \left(\frac{1+\frac{\eta\gamma_{2,\infty}^2(c)}2(t+1)}
{1+\frac{\eta\gamma_{2,\infty}^2(c)}2t}\right),
\end{align*}
which implies
\begin{align}
\langle\mathbf{w}_{t+1}-\tilde{\mathbf{w}}_{t+1},
\tilde{\mathbf{w}}_{t+1}-\tilde{\mathbf{w}}_t\rangle
&=\ln\left(\frac{1+\frac{\eta\gamma_{2,\infty}^2(c)}2(t+1)}
{1+\frac{\eta\gamma_{2,\infty}^2(c)}2t}\right)
\langle\mathbf{w}_{t+1}-\tilde{\mathbf{w}}_{t+1},
\hat{\mathbf{w}}(c)\rangle\overset{\eqref{equ:ib_steps_comparator_projection}}{\ge}0.\label{equ:ib_steps_comparator_cross}
\end{align}
 Expanding the square, we have
\begin{align}
\lVert\mathbf{w}_{t+1}-\tilde{\mathbf{w}}_t\rVert_2^2
&=\left\lVert
(\mathbf{w}_{t+1}-\tilde{\mathbf{w}}_{t+1})
+(\tilde{\mathbf{w}}_{t+1}-\tilde{\mathbf{w}}_t)
\right\rVert_2^2\nonumber\\
&=\lVert\mathbf{w}_{t+1}-\tilde{\mathbf{w}}_{t+1}\rVert_2^2
+2\langle\mathbf{w}_{t+1}-\tilde{\mathbf{w}}_{t+1},
\tilde{\mathbf{w}}_{t+1}-\tilde{\mathbf{w}}_t\rangle +\lVert\tilde{\mathbf{w}}_{t+1}-\tilde{\mathbf{w}}_t\rVert_2^2\nonumber\\
& \overset{\eqref{equ:ib_steps_comparator_cross}}{\ge}\lVert\mathbf{w}_{t+1}-\tilde{\mathbf{w}}_{t+1}\rVert_2^2.\label{equ:ib_comparator_move}
\end{align}

Since \eqref{equ:advloss} is convex and $(\mathbf{w}_t-\mathbf{w}_{t+1})/\eta$ is a subgradient at
$\mathbf{w}_t$, we have
\begin{align*}
\CL_{\mathrm{Adv}}(\tilde{\mathbf{w}}_t)
&\ge \CL_{\mathrm{Adv}}(\mathbf{w}_t)
+\frac1\eta\langle\mathbf{w}_t-\tilde{\mathbf{w}}_t,
\mathbf{w}_{t+1}-\mathbf{w}_t\rangle.
\end{align*}
Rearranging and multiplying by $\eta>0$ yields
\begin{align}
\langle\mathbf{w}_t-\tilde{\mathbf{w}}_t,
\mathbf{w}_{t+1}-\mathbf{w}_t\rangle
\le\eta\bigl[\CL_{\mathrm{Adv}}(\tilde{\mathbf{w}}_t)
-\CL_{\mathrm{Adv}}(\mathbf{w}_t)\bigr].\label{equ:ib_steps_subgradient_cross}
\end{align}
Combining this with the preceding square expansion,
\begin{align}
\lVert\mathbf{w}_{t+1}-\tilde{\mathbf{w}}_{t+1}\rVert_2^2
&\overset{\eqref{equ:ib_comparator_move}}{\le}\lVert\mathbf{w}_{t+1}-\tilde{\mathbf{w}}_t\rVert_2^2
=\left\lVert(\mathbf{w}_t-\tilde{\mathbf{w}}_t)
+(\mathbf{w}_{t+1}-\mathbf{w}_t)\right\rVert_2^2\nonumber\\
&=\lVert\mathbf{w}_t-\tilde{\mathbf{w}}_t\rVert_2^2
+2\langle\mathbf{w}_t-\tilde{\mathbf{w}}_t,
\mathbf{w}_{t+1}-\mathbf{w}_t\rangle
+\lVert\mathbf{w}_{t+1}-\mathbf{w}_t\rVert_2^2\nonumber\\
&\overset{\eqref{equ:ib_steps_subgradient_cross}}{\le}\lVert\mathbf{w}_t-\tilde{\mathbf{w}}_t\rVert_2^2
+2\eta\bigl[\CL_{\mathrm{Adv}}(\tilde{\mathbf{w}}_t)
-\CL_{\mathrm{Adv}}(\mathbf{w}_t)\bigr]
+\lVert\mathbf{w}_{t+1}-\mathbf{w}_t\rVert_2^2.\label{equ:ib_convex_distance}
\end{align}

Since the scalar multiplying $\hat{\mathbf{w}}(c)$
in $\tilde{\mathbf{w}}_t$ is nonnegative, we have
\begin{align}
&\mathbf{x}_i^\top\tilde{\mathbf{w}}_t-c\lVert\tilde{\mathbf{w}}_t\rVert_1\overset{\eqref{equ:ib_det_tilde}}{=}\ln\left(1+\frac{\eta\gamma_{2,\infty}^2(c)}2t\right)
\bigl[\mathbf{x}_i^\top\hat{\mathbf{w}}(c)
-c\lVert\hat{\mathbf{w}}(c)\rVert_1\bigr]
\ge\ln\left(1+\frac{\eta\gamma_{2,\infty}^2(c)}2t\right),\label{equ:ib_steps_comparator_score}
\end{align}
where the last inequality comes from $\hat{\mathbf{w}}(c)$ is a solution of \eqref{equ:rsvm}. Moreover, we have
\begin{align}
\CL_{\mathrm{Adv}}(\tilde{\mathbf{w}}_t)
=\sum_{i=1}^N\ln\left(1+
e^{-\mathbf{x}_i^\top\tilde{\mathbf{w}}_t+
c\lVert\tilde{\mathbf{w}}_t\rVert_1}\right)\le\sum_{i=1}^N
&e^{-\mathbf{x}_i^\top\tilde{\mathbf{w}}_t+
c\lVert\tilde{\mathbf{w}}_t\rVert_1}\overset{\eqref{equ:ib_steps_comparator_score}}{\le}\sum_{i=1}^N
e^{-\ln(1+\frac{\eta\gamma_{2,\infty}^2(c)}2t)}\nonumber\\
& =\frac{N}{1+\frac{\eta\gamma_{2,\infty}^2(c)}2t},\label{equ:ib_comparator_loss}
\end{align}
where the second inequality used $\ln(1+x)\le x, x \ge 0$. 

Also, since $0<\lambda_{i,t}<1$ and $\mathbf{w}_0=0$, we have
\begin{align}\label{equ:ib_update_bounds}
\lVert\mathbf{w}_{t+1}-\mathbf{w}_t\rVert_2
&=\eta\left\lVert\sum_{i=1}^N
\lambda_{i,t}(\mathbf{x}_i-c\mathbf{s}_t)\right\rVert_2  \le\eta\sum_{i=1}^N\lambda_{i,t}
\lVert\mathbf{x}_i-c\mathbf{s}_t\rVert_2 
\le\eta M\sum_{i=1}^N\lambda_{i,t}.
\end{align}
Then $\sum_i\lambda_{i,t}\le N$ implies
\begin{align}
\lVert\mathbf{w}_{t+1}-\mathbf{w}_t\rVert_2^2
&\le\eta^2M^2\left(\sum_{i=1}^N\lambda_{i,t}\right)^2
\le\eta^2NM^2\sum_{i=1}^N\lambda_{i,t}.\label{equ:ib_steps_update_square}
\end{align}
Summing the recursion inequality of $v_t$ in \eqref{equ:vt} and using $v_0=0$ gives
\begin{align}
\eta\gamma_{2,\infty}(c)\sum_{s=0}^{t-1}\sum_{i=1}^N\lambda_{i,s}
&\overset{\eqref{equ:vt}}{\le}\sum_{s=0}^{t-1}(v_{s+1}-v_s)=v_t,\nonumber \quad \mathrm{and}\\
&\sum_{s=0}^{t-1}\lVert\mathbf{w}_{s+1}-\mathbf{w}_s\rVert_2^2
\overset{\eqref{equ:ib_steps_update_square}}{\le}\eta^2NM^2\sum_{s=0}^{t-1}\sum_{i=1}^N\lambda_{i,s}
\le\frac{\eta NM^2}{\gamma_{2,\infty}(c)}v_t.\label{equ:ib_update_sum_initial}
\end{align}

Now sum the squared-distance recursion. Since
$\mathbf{w}_0=\tilde{\mathbf{w}}_0=0$ and the loss is nonnegative,
\begin{align}
\lVert\mathbf{w}_t-\tilde{\mathbf{w}}_t\rVert_2^2 & =\sum_{s=0}^{t-1}
\left(\lVert\mathbf{w}_{s+1}-\tilde{\mathbf{w}}_{s+1}\rVert_2^2
-\lVert\mathbf{w}_s-\tilde{\mathbf{w}}_s\rVert_2^2\right)\nonumber\\
&\quad\overset{\eqref{equ:ib_convex_distance}}{\le}2\eta\sum_{s=0}^{t-1}\CL_{\mathrm{Adv}}(\tilde{\mathbf{w}}_s)
-2\eta\sum_{s=0}^{t-1}\CL_{\mathrm{Adv}}(\mathbf{w}_s)
+\sum_{s=0}^{t-1}\lVert\mathbf{w}_{s+1}-\mathbf{w}_s\rVert_2^2\nonumber\\
&\quad\overset{\eqref{equ:ib_comparator_loss},\eqref{equ:ib_update_sum_initial}}{\le}2\eta N\sum_{s=0}^{t-1}
\frac1{1+\frac{\eta\gamma_{2,\infty}^2(c)}2s}
+\frac{\eta NM^2}{\gamma_{2,\infty}(c)}v_t,\label{equ:ib_steps_preliminary_telescope}
\end{align}
here we use a rather loss bound by ignoring the effect of the $2\eta\sum_{s=0}^{t-1}\CL_{\mathrm{Adv}}(\mathbf{w}_s)$ term.  

For $t\ge1$, we have
\begin{align}
\sum_{s=0}^{t-1}\frac1{1+\frac{\eta\gamma_{2,\infty}^2(c)}2s}
=1+\sum_{s=1}^{t-1}\frac1{1+\frac{\eta\gamma_{2,\infty}^2(c)}2s}\nonumber
&\le1+\int_0^t\frac{\mathrm dx}{1+\frac{\eta\gamma_{2,\infty}^2(c)}2x}\nonumber\\
&=1+\frac2{\eta\gamma_{2,\infty}^2(c)}
\ln\left(1+\frac{\eta\gamma_{2,\infty}^2(c)}2t\right).\label{equ:ib_steps_preliminary_integral}
\end{align}
It follows that
\begin{align}
\lVert\mathbf{w}_t-\tilde{\mathbf{w}}_t\rVert_2^2
&\overset{\eqref{equ:ib_steps_preliminary_telescope},\eqref{equ:ib_steps_preliminary_integral}}{\le}2\eta N+\frac{4N}{\gamma_{2,\infty}^2(c)}
\ln\left(1+\frac{\eta\gamma_{2,\infty}^2(c)}2t\right)
+\frac{\eta NM^2}{\gamma_{2,\infty}(c)}v_t.\label{equ:ib_distance_with_v}
\end{align}
The right side also bounds the zero initial distance at $t=0$. To eliminate $v_t$, apply Cauchy--Schwarz and
$\lVert\mathbf{u}_{2,\infty}(c)\rVert_2=1$, we get
\begin{align}
v_t
&=\langle\mathbf{w}_t-\tilde{\mathbf{w}}_t,
\mathbf{u}_{2,\infty}(c)\rangle
+\langle\tilde{\mathbf{w}}_t,\mathbf{u}_{2,\infty}(c)\rangle\nonumber\\
&\overset{\eqref{equ:ib_det_tilde}}{\le}\lVert\mathbf{w}_t-\tilde{\mathbf{w}}_t\rVert_2
+\frac1{\gamma_{2,\infty}(c)}
\ln\left(1+\frac{\eta\gamma_{2,\infty}^2(c)}2t\right).\label{equ:ib_steps_v_from_distance}
\end{align}
Substitution gives
\begin{align}
\lVert\mathbf{w}_t-\tilde{\mathbf{w}}_t\rVert_2^2
&\overset{\eqref{equ:ib_distance_with_v},\eqref{equ:ib_steps_v_from_distance}}{\le}2\eta N+\frac{4N+\eta NM^2}{\gamma_{2,\infty}^2(c)}
\ln\left(1+\frac{\eta\gamma_{2,\infty}^2(c)}2t\right)\nonumber\\
&\quad+\frac{\eta NM^2}{\gamma_{2,\infty}(c)}
\lVert\mathbf{w}_t-\tilde{\mathbf{w}}_t\rVert_2.\label{equ:ib_steps_distance_substitution}
\end{align}
The inequality $2ab\le a^2+b^2$ gives
\begin{align}
\frac{\eta NM^2}{\gamma_{2,\infty}(c)}
\lVert\mathbf{w}_t-\tilde{\mathbf{w}}_t\rVert_2
\le\frac12\lVert\mathbf{w}_t-\tilde{\mathbf{w}}_t\rVert_2^2
+\frac12\left(\frac{\eta NM^2}{\gamma_{2,\infty}(c)}\right)^2.\label{equ:ib_steps_young_distance}
\end{align}
Subtracting the $\frac{1}{2}\lVert\mathbf{w}_t-\tilde{\mathbf{w}}_t\rVert_2^2$ term from both sides and
multiplying by two yields
\begin{align}
\lVert\mathbf{w}_t-\tilde{\mathbf{w}}_t\rVert_2^2
&\overset{\eqref{equ:ib_steps_distance_substitution},\eqref{equ:ib_steps_young_distance}}{\le}4\eta N+\left(\frac{\eta NM^2}{\gamma_{2,\infty}(c)}\right)^2
+\frac{2(4N+\eta NM^2)}{\gamma_{2,\infty}^2(c)}
\ln\left(1+\frac{\eta\gamma_{2,\infty}^2(c)}2t\right).\label{equ:ib_distance_closed}
\end{align}

In the following, we want to rewrite \eqref{equ:ib_distance_closed} in the form
$\lVert\mathbf{w}_t-\tilde{\mathbf{w}}_t\rVert_2^2
\le C_2^2\ln(t+2)$, with $C_2$ be some constant independent of $t$.
For brevity, set
\[
a:=\frac{\eta\gamma_{2,\infty}^2(c)}{2},
\qquad
A:=4\eta N+
\left(\frac{\eta NM^2}{\gamma_{2,\infty}(c)}\right)^2,
\qquad
B:=\frac{2(4N+\eta NM^2)}{\gamma_{2,\infty}^2(c)}.
\]
Then \eqref{equ:ib_distance_closed} reads
\[
\lVert\mathbf{w}_t-\tilde{\mathbf{w}}_t\rVert_2^2
\le A+B\ln(1+at).
\]
For every $t\ge0$, we have
$1+at\le\max\{1,a\}(t+2)$ and $\ln(t+2)\ge\ln2$.
Consequently,
\begin{align*}
A+B\ln(1+at)
&\le B\ln(t+2)+A+B\ln\max\{1,a\}\le
\left(
B+\frac{A+B\ln\left(\max\{1,a\}\right)}{\ln2}
\right)\ln(t+2).
\end{align*}
Thus, defining the positive constant $C_2$ by
\begin{equation}\label{equ:ib_const_2}
C_2^2
:=B+\frac{A+B\ln\left(\max\{1,a\}\right)}{\ln2},
\end{equation}
we obtain
\begin{equation}\label{equ:ib_steps_preliminary_distance_final}
\lVert\mathbf{w}_t-\tilde{\mathbf{w}}_t\rVert_2^2
\le C_2^2\ln(t+2),
\qquad t\ge0.
\end{equation}

Since $\tilde{\mathbf{w}}_t$ is parallel to $\mathbf{u}_{2,\infty}(c)$, we have
\begin{align}
\mathbf{w}_t-\tilde{\mathbf{w}}_t
&\overset{\substack{\eqref{equ:decwur},\eqref{equ:ib_det_tilde}}}{=}\left[v_t-\frac{\ln(1+\frac{\eta\gamma_{2,\infty}^2(c)}2t)}
{\gamma_{2,\infty}(c)}\right]\mathbf{u}_{2,\infty}(c)+\mathbf{r}_t.\label{equ:ib_steps_distance_decomposition}
\end{align}
Taking the square of both side on \eqref{equ:ib_steps_distance_decomposition}, the cross term in its squared norm vanishes because
$\langle\mathbf{u}_{2,\infty}(c),\mathbf{r}_t\rangle=0$. Therefore we get
\begin{align*}
&\lVert\mathbf{w}_t-\tilde{\mathbf{w}}_t\rVert_2^2
\overset{\eqref{equ:ib_steps_distance_decomposition}}{=}\left[v_t-\frac{\ln(1+\frac{\eta\gamma_{2,\infty}^2(c)}2t)}
{\gamma_{2,\infty}(c)}\right]^2
\lVert\mathbf{u}_{2,\infty}(c)\rVert_2^2+\lVert\mathbf{r}_t\rVert_2^2 \ge\lVert\mathbf{r}_t\rVert_2^2,\nonumber\\
\end{align*}
which implies
\begin{align}
\lVert\mathbf{r}_t\rVert_2
\le\lVert\mathbf{w}_t-\tilde{\mathbf{w}}_t\rVert_2
\overset{\eqref{equ:ib_steps_preliminary_distance_final}}{\le} C_2\sqrt{\ln(t+2)}.\label{equ:ib_preliminary}
\end{align}

The triangle inequality and Cauchy--Schwarz give
\begin{align}
\left(\mathbf{x}_i^\top\mathbf{w}_t-c\lVert\mathbf{w}_t\rVert_1\right)
-\left(\mathbf{x}_i^\top\tilde{\mathbf{w}}_t
-c\lVert\tilde{\mathbf{w}}_t\rVert_1\right) \ge  -\lVert\mathbf{x}_i\rVert_2
\lVert\mathbf{w}_t-\tilde{\mathbf{w}}_t\rVert_2
-c\sqrt d\,\lVert\mathbf{w}_t-&\tilde{\mathbf{w}}_t\rVert_2\nonumber\\
\ge-M\lVert\mathbf{w}_t-\tilde{\mathbf{w}}_t\rVert_2.\label{equ:ib_steps_score_lipschitz}
\end{align}
Combining above bounds, we get
\begin{align}
&\mathbf{x}_i^\top\mathbf{w}_t-c\lVert\mathbf{w}_t\rVert_1
\overset{\substack{\eqref{equ:ib_steps_score_lipschitz},\eqref{equ:ib_steps_comparator_score}\\\eqref{equ:ib_preliminary}}}{\ge}\ln\left(1+\frac{\eta\gamma_{2,\infty}^2(c)}2t\right)
-MC_2\sqrt{\ln(t+2)},\label{equ:ib_score_and_loss}
\end{align}
which gives
\begin{align}
&\CL_{\mathrm{Adv}}(\mathbf{w}_t)
\le\sum_{i=1}^Ne^{-\mathbf{x}_i^\top\mathbf{w}_t+c\lVert\mathbf{w}_t\rVert_1}
\overset{\eqref{equ:ib_score_and_loss}}{\le}\frac{N\exp(MC_2\sqrt{\ln(t+2)})}
{1+\frac{\eta\gamma_{2,\infty}^2(c)}2t},\nonumber\\
&\sum_{i=1}^N\lambda_{i,t}
\le\sum_{i=1}^Ne^{-\mathbf{x}_i^\top\mathbf{w}_t+c\lVert\mathbf{w}_t\rVert_1}
\overset{\eqref{equ:ib_score_and_loss}}{\le}\frac{N\exp(MC_2\sqrt{\ln(t+2)})}
{1+\frac{\eta\gamma_{2,\infty}^2(c)}2t}.\label{equ:ib_steps_weights_loss_bound}
\end{align}

Now we want to bound the right hand side term of \eqref{equ:ib_steps_weights_loss_bound} more explicit in terms of $t$. Since $C_1=\max\{1,2/(\eta\gamma_{2,\infty}^2(c))\}$,
we have
\[(1+\eta\gamma_{2,\infty}^2(c)t/2)^{-1}\le C_1/(t+1).\]
Moreover, the inequality
$(\sqrt{x}/2-MC_2)^2\ge0$ gives
$MC_2\sqrt{x}\le x/4+M^2C_2^2$ for every $x\ge0$.
Applying this with $x=\ln(t+2)$ yields
\begin{align}
\frac{N\exp(MC_2\sqrt{\ln(t+2)})}
{1+\frac{\eta\gamma_{2,\infty}^2(c)}2t}
&\overset{\eqref{equ:ib_v_lower}}{\le}\frac{NC_1}{t+1}
\exp\left(\frac14\ln(t+2)+M^2C_2^2\right)\nonumber\\
&=\frac{NC_1e^{M^2C_2^2}(t+2)^{1/4}}{t+1}
\le\frac{2^{1/4}NC_1e^{M^2C_2^2}}{(t+1)^{3/4}}
=\frac{C_3}{(t+1)^{3/4}},\label{equ:ib_steps_weight_decay}
\end{align}
where $C_3=2^{1/4}NC_1\exp(M^2C_2^2)$ is independent of $t$. In particular, we get
\begin{align}
0\le\CL_{\mathrm{Adv}}(\mathbf{w}_t)
\le\frac{C_3}{(t+1)^{3/4}},\qquad
\sum_{i=1}^N\lambda_{i,t}\le\frac{C_3}{(t+1)^{3/4}}.\label{equ:ib_steps_loss_zero}
\end{align}
Finally, since
\begin{align*}
\sum_{t=0}^{\infty}(t+1)^{-3/2}
&=1+\sum_{k=2}^{\infty}k^{-3/2}
\le 1+\int_1^\infty x^{-3/2}\,\mathrm dx=3,
\end{align*}

we get
\begin{align}
\sum_{t=0}^\infty\lVert\mathbf{w}_{t+1}-\mathbf{w}_t\rVert_2^2
&\overset{\eqref{equ:ib_update_bounds}}{\le}\eta^2M^2\sum_{t=0}^\infty
\left(\sum_{i=1}^N\lambda_{i,t}\right)^2
\overset{\eqref{equ:ib_steps_loss_zero}}{\le}\eta^2M^2C_3^2\sum_{t=0}^\infty(t+1)^{-3/2}
\le C_4,\label{equ:ib_summable_updates}
\end{align}
where $c_4 = 3\eta^2M^2C_3^2$. This finished the proof of Lemma~\ref{lem:ib_preliminary}.
\end{proof}

Lemma~\ref{lem:ib_preliminary} already implies a preliminary convergence rate for implicit bias. Indeed, recalling
$\mathbf{w}_t=v_t\mathbf{u}_{2,\infty}(c)+\mathbf{r}_t$
with $\mathbf{r}_t\perp\mathbf{u}_{2,\infty}(c)$,
we have $\|\mathbf{w}_t\|_2\ge v_t>0$ for $t\ge1$.
The triangle inequality also gives
$\|\mathbf{w}_t\|_2-v_t\le\|\mathbf{r}_t\|_2$.
Hence,
\begin{align*}
\left\|
\frac{\mathbf{w}_t}{\|\mathbf{w}_t\|_2}
-\mathbf{u}_{2,\infty}(c)
\right\|_2
&\le
\frac{\|\mathbf{r}_t\|_2}{\|\mathbf{w}_t\|_2}
+\frac{\|\mathbf{w}_t\|_2-v_t}{\|\mathbf{w}_t\|_2}\le \frac{2\|\mathbf{r}_t\|_2}{v_t}\\
\\
&\le
\frac{2\gamma_{2,\infty}(c)C_2\sqrt{\ln(t+2)}}
{\ln\!\left(1+\frac{\eta\gamma_{2,\infty}^2(c)}{2}t\right)}
=\mathcal{O}\!\left(\frac{1}{\sqrt{\ln t}}\right),
\qquad t\to\infty.
\end{align*}
This convergence rate proved above is the same as that of Theorem~3.4 in \cite{Li2020Implicit}. In the following, we will improve this bound by choosing  more suitable comparing points than $\tilde{\textbf{w}}_t$.

\subsection{Contraction of comparison point}\label{sec:ji1}

For $\mathbf r\perp\mathbf u_{2,\infty}(c)$, define
\begin{align}
p_i(\mathbf r)
&=\langle\mathbf x_i,\mathbf r\rangle
-c\sum_{j:\,u_{2,\infty,j}(c)\ne0}
\Sign(u_{2,\infty,j}(c))r_j
-c\sum_{j:\,u_{2,\infty,j}(c)=0}|r_j|.
\label{equ:ib_p_function}
\end{align}
The quantity $p_i(\mathbf r)$ represents the change in the robust
margin of $\mathbf x_i$ induced by adding the orthogonal component
$\mathbf r$ to $v\mathbf u_{2,\infty}(c)$, provided that this addition
preserves the signs on the nonzero coordinates of
$\mathbf u_{2,\infty}(c)$.
Specifically, if $v>0$ and
$
\Sign\bigl(vu_{2,\infty,j}(c)+r_j\bigr)
=\Sign\bigl(u_{2,\infty,j}(c)\bigr)\,
\text{for every }j\text{ with }u_{2,\infty,j}(c)\ne0,
$
then
\begin{align*}
p_i(\mathbf r)
&=\Bigl[
\mathbf x_i^\top\bigl(v\mathbf u_{2,\infty}(c)+\mathbf r\bigr)
-c\bigl\|v\mathbf u_{2,\infty}(c)+\mathbf r\bigr\|_1
\Bigr]-\Bigl[
\mathbf x_i^\top\bigl(v\mathbf u_{2,\infty}(c)\bigr)
-c\bigl\|v\mathbf u_{2,\infty}(c)\bigr\|_1
\Bigr].
\end{align*}

We define
\[
F(\mathbf r)=\sum_{i\in\CS(c)}e^{-p_i(\mathbf r)}.
\]
For each fixed $\mathbf r\perp\mathbf u_{2,\infty}(c)$,
the sign preservation condition above holds for all sufficiently
large $v$, which will be the case of \eqref{equ:gd} dynamics as shown by Proposition~\ref{lem:trainingproj}. The contribution of the robust support vectors to
the adversarial logistic loss then satisfies
\[
\sum_{i\in\CS(c)}
\ln\left(1+e^{-\gamma_{2,\infty}(c)v-p_i(\mathbf r)}\right)
\approx e^{-\gamma_{2,\infty}(c)v}F(\mathbf r),
\qquad \mathrm{for\, sufficient\, large\, } v.
\]

Thus, for fixed $\mathbf r$ and sufficiently large $v$,
$F(\mathbf r)$ captures the contribution of the 
component $\mathbf r$ of parameter $\mathbf w = v\mathbf u_{2,\infty}(c) + \mathbf r$ to the adversarial logistic loss. Moreover, It can also be seen that 
\begin{align}
|p_i(\mathbf{r})|
&\overset{\eqref{equ:ib_p_function}}{\le}|\langle\mathbf{x}_i,\mathbf{r}\rangle|
+c\sum_{j:\,u_{2,\infty,j}(c)\ne0}|r_j|
+c\sum_{j:\,u_{2,\infty,j}(c)=0}|r_j|
&\le\lVert\mathbf{x}_i\rVert_2\lVert\mathbf{r}\rVert_2
+c\lVert\mathbf{r}\rVert_1
\le M\lVert\mathbf{r}\rVert_2.\label{equ:ib_steps_p_norm}
\end{align}


The following Propoaition is an extension of Lemma~3.3 in \cite{ji2018risk} to the adversarial loss. 

\begin{prop}[Adversarial version of {\citet[Lemma~3.3]{ji2018risk}}]\label{lem:ib_approx}
There are fixed $\bar{\mathbf{r}},\mathbf{v}\perp\mathbf{u}_{2,\infty}(c)$
and $C_5,C_6\ge1$, independent of $t$, such that
$\mathbf{q}_{\varepsilon}=\bar{\mathbf{r}}+
\mathbf{v}\ln(C_5/\varepsilon)$ satisfies, for $0<\varepsilon\le1$,
\begin{align}
F(\mathbf{q}_{\varepsilon})
&\le\inf_{\mathbf{r}\perp\mathbf{u}_{2,\infty}(c)}F(\mathbf{r})
+\varepsilon, \quad
\lVert\mathbf{q}_{\varepsilon}\rVert_2
\le C_6\bigl(1+\ln(1/\varepsilon)\bigr).\label{equ:ib_approx_statement}
\end{align}
\end{prop}

Let 
\[
H:=\{\mathbf r:\langle\mathbf r,\mathbf u_{2,\infty}(c)\rangle=0\},
\qquad
F_*:=\inf_{\mathbf r\in H}F(\mathbf r).
\]
To specify what $\bar{\mathbf r}$ and $\mathbf v$ must accomplish,
choose a sequence $\mathbf r^{(n)}\in H$ such that
\[
F(\mathbf r^{(n)})\rightarrow F_*, \quad \mathrm{as} \quad n\to \infty,
\quad \mathrm{and}\quad
F(\mathbf r^{(n)})\le N+1.
\]
Such a sequence exists because $F_*\le F(\mathbf0)=|S(c)|\le N$.
Each term $e^{-p_i(\mathbf r^{(n)})}$ belongs to $[0,N+1]$.
After passing to a subsequence, all these terms have limits:
\[
e^{-p_i(\mathbf r^{(n)})}\rightarrow\ell_i,
\quad i\in S(c).
\]
Let
$I:=\{i\in S(c):\ell_i>0\},\,
b_i:=-\ln\ell_i$ for $i\in I$.
Then as $n \to \infty$, we have
\[
p_i(\mathbf r^{(n)})\rightarrow b_i\in\mathbb R \quad \mathrm{for}\quad
i\in I,
\quad
p_j(\mathbf r^{(n)})\rightarrow+\infty \quad \mathrm{for}\quad
j\in S(c)\setminus I,
\]
which also implies $F_*=\sum_{i\in I}e^{-b_i}$. We will construct a convex polyhedral cone $K\subseteq H$
and two vectors $\bar{\mathbf r},\mathbf v\in K$ such that
\vspace{-0.3cm}
\begin{itemize}
\item[(1).] each $p_i(\mathbf r)$ is linear on $K$,
\item[(2).] $p_i(\bar{\mathbf r})=b_i$ for $i\in I$,
\item[(3).] $p_i(\mathbf v)=0$ for $i\in I$ and
$p_j(\mathbf v)\ge1$ for $j\in S(c)\setminus I$.
\end{itemize}
\vspace{-0.2cm}
Condition~(2) ensures that $\bar{\mathbf r}$ realizes the finite
limiting values $b_i$ for $i\in I$. Together with linearity
in~(1), condition~(3) ensures that moving from $\bar{\mathbf r}$
along $\mathbf v$ preserves these values while increasing
$p_j$ for every $j\in S(c)\setminus I$.
If $S(c)\setminus I\ne\varnothing$, then
$F(\bar{\mathbf r})>F_*$ because the corresponding exponential
terms are still positive at $\bar{\mathbf r}$; moving along
$\mathbf v$ makes these terms tend to zero. We first establish the claimed bounds assuming this construction,
and then prove that such $K$, $\bar{\mathbf r}$, and $\mathbf v$ exist.

\begin{proof}[Proof of Proposition~\ref{lem:ib_approx} assuming conditions (1)-(3)]
Since $K$ is a convex cone,
$\bar{\mathbf r}+\tau\mathbf v\in K$ for every $\tau\ge0$. Since $p(\mathrm r)$ is linear on $K$, we have 
\begin{align*}
p_i(\bar{\mathbf r}+\tau\mathbf v)=b_i
\quad \mathrm{for}\quad i\in I, \quad \mathrm{and} \quad 
p_j(\bar{\mathbf r}+\tau\mathbf v)
=p_j(\bar{\mathbf r})+\tau p_j(\mathbf v)
\quad \mathrm{for}\quad j\in S(c)\setminus I.
\end{align*}
Consequently,
\begin{align*}
F(\bar{\mathbf r}+\tau\mathbf v)-F_*
=\sum_{j\in S(c)\setminus I}
e^{-p_j(\bar{\mathbf r})-\tau p_j(\mathbf v)}\le e^{-\tau}
\sum_{j\in S(c)\setminus I}e^{-p_j(\bar{\mathbf r})},
\end{align*}
where the last inequality used condition (3). For $0<\varepsilon\le1$, consider
\[
\mathbf q_\varepsilon
=\bar{\mathbf r}+\mathbf v\ln(C_5/\varepsilon),
\]
where $C_5\ge1$ will be determined below.
Since $\ln(C_5/\varepsilon)\ge0$, the preceding estimate gives
\begin{align}\label{equ:diff_qminF}
F(\mathbf q_\varepsilon)-F_*
\le \frac{\varepsilon}{C_5}
\sum_{j\in S(c)\setminus I}e^{-p_j(\bar{\mathbf r})}.    
\end{align}
Moreover, the triangle inequality yields
\begin{align}\label{equ:esti_qepsilon}
\|\mathbf q_\varepsilon\|_2
&\le \|\bar{\mathbf r}\|_2
+\|\mathbf v\|_2\ln C_5
+\|\mathbf v\|_2\ln(1/\varepsilon)\le
\bigl[\|\bar{\mathbf r}\|_2
+(1+\ln C_5)\|\mathbf v\|_2\bigr]
\bigl(1+\ln(1/\varepsilon)\bigr).
\end{align}
Thus by choosing
\[
C_5:=\max\left\{
1,\sum_{j\in S(c)\setminus I}e^{-p_j(\bar{\mathbf r})}
\right\}, \quad \mathrm{and}\quad
C_6:=\|\bar{\mathbf r}\|_2
+(1+\ln C_5)\|\mathbf v\|_2,
\]
and take them into \eqref{equ:diff_qminF} and \eqref{equ:esti_qepsilon}, we get
\begin{align*}
F(\mathbf q_\varepsilon)-F_*  \le \epsilon, \quad \mathrm{and}\quad \lVert\mathbf{q}_{\varepsilon}\rVert_2 \le C_6 (1+\ln(1/\epsilon)),
\end{align*}
which finished the proof.
\end{proof}

\begin{lem}
The required construction of  $K,\,\bar{\mathbf r}$ and $\mathbf v$ in conditions (1)-(3) exists.
\end{lem}
\begin{proof}
First, we select a region where the absolute values in $p_i(\mathbf r)$
can be replaced by linear expressions.
There are finitely many sign choices for the coordinates
$r_j$ with $u_j=0$.
Hence a further subsequence of $\mathbf r^{(n)}$ lies in one
fixed convex polyhedral cone
\[
K:=
\left\{
\mathbf r\in H:
\sigma_jr_j\ge0\ \text{for every }j\text{ with }u_j=0
\right\},
\qquad
\sigma_j\in\{-1,1\}.
\]
In particular, $\inf_{\mathbf r\in K}F(\mathbf r)=F_*$.
Indeed, since $F(\mathbf r^{(n)})\to F_*$, the function values
along any subsequence converge to the same limit.
For $\mathbf r\in K$, we have
\[
p_i(\mathbf r)
=
\mathbf x_i^\top\mathbf r
-c\sum_{j:u_{2,\infty,j}\ne0}\operatorname{Sign}(u_{2,\infty,j})r_j
-c\sum_{j:u_{2,\infty,j}=0}\sigma_jr_j,
\]
which is a linear function of $\mathbf r$.

In the following, we will use the elementary fact that a linear image of a polyhedron
is a polyhedron and hence is closed, e.g., \citet[Theorem~19.3]{rockafellar1970convex}.

We first show the existence of $\bar{\mathbf r}$. 
If $I=\varnothing$, take $\bar{\mathbf r}=\mathbf0$.
Otherwise, the set
\[
\left\{(p_i(\mathbf r))_{i\in I}:\mathbf r\in K\right\}
\]
is a closed polyhedral image.
Since it contains the sequence
$(p_i(\mathbf r^{(n)}))_{i\in I}\to(b_i)_{i\in I}$,
it also contains the limit.
Therefore some $\bar{\mathbf r}\in K$ satisfies
\[
p_i(\bar{\mathbf r})=b_i \quad \mathrm{for} \quad i\in I.
\]
We then show the existence of $\bar{\mathbf v}$. 
If $I=S(c)$, take $\mathbf v=\mathbf0$.
Otherwise, all $p_j(\mathbf r^{(n)})$ with
$j\in S(c)\setminus I$ tend to $+\infty$.
Passing to a further subsequence and relabeling, arrange that
\[
p_j(\mathbf r^{(n)})\ge n
\quad \mathrm{for} \quad j\in S(c)\setminus I.
\]
Set $\mathbf d^{(n)}:=\mathbf r^{(n)}/n$.
The cone property and linearity on $K$ give
\[
\mathbf d^{(n)}\in K,\qquad
p_i(\mathbf d^{(n)})\rightarrow0 \quad \mathrm{for} \quad i\in I,
\qquad
p_j(\mathbf d^{(n)})\ge1
\quad \mathrm{for} \quad j\in S(c)\setminus I.
\]
Thus $\{\mathbf d^{(n)}\}$ make all $p_i$, $i\in I$, arbitrarily
close to zero, while keeping $p_j\ge1$ for every
$j\in S(c)\setminus I$. To find a vector $\mathbf v\in K$ satisfying
$p_i(\mathbf v)=0$ for every $i\in I$ while retaining
$p_j(\mathbf v)\ge1$ for every $j\in S(c)\setminus I$,
consider the polyhedron
\[
D:=
\left\{
\mathbf r\in K:
p_j(\mathbf r)\ge1
\ \text{for every }j\in S(c)\setminus I
\right\}.
\]
If $I\ne\varnothing$, the image
\[
\left\{(p_i(\mathbf r))_{i\in I}:\mathbf r\in D\right\}
\]
is closed. It contains
$(p_i(\mathbf d^{(n)}))_{i\in I}\to\mathbf0$,
so it contains $\mathbf0$.
Hence there exist some $\mathbf v\in D$ satisfies
$p_i(\mathbf v)=0$ for every $i\in I$.
If $I=\varnothing$, any $\mathbf d^{(n)}$ serves as $\mathbf v$.
In both cases,
\[
p_i(\mathbf v)=0\quad \mathrm{for} \quad i\in I,
\qquad
p_j(\mathbf v)\ge1 \quad \mathrm{for} \quad  j\in S(c)\setminus I.
\]
This proves the required existence.
\end{proof}

\subsection{Cumulative loss comparison}\label{sec:clc}

For $T\ge3$, fix the vectors from Lemma~\ref{lem:ib_approx} and set
\begin{align}
\mathbf{q}_T&=\bar{\mathbf{r}}+\mathbf{v}\ln\bigl(C_5\ln(T+2)\bigr),\label{equ:ib_qT_definition}\\
t_T&=\left\lceil e^{(\ln\ln(T+2))^2}\right\rceil.\label{equ:ib_start_index}
\end{align}

\begin{prop}\label{lem:ib_loss_gap}
There exist an integer $T_0\ge3$ and nonnegative constants
$\tilde{C}$, which is independent of $T$, such that, for every integer
$T\ge T_0$, we have $t_T<T$ and
\begin{align}
\sum_{t=t_T}^{T-1}\bigl[
\CL_{\mathrm{Adv}}(v_t\mathbf{u}_{2,\infty}(c)+\mathbf{q}_T)
-\CL_{\mathrm{Adv}}(\mathbf{w}_t)\bigr]
\le\tilde{C}.\label{equ:ib_cumulative_loss_gap}
\end{align}
\end{prop}

\begin{lem}[Loss decomposition]\label{lem:ib_loss_decomposition}
For every integer $T\ge3$ and $t\ge0$, define
\begin{align}
& E_{1,t,T}
=e^{-\gamma_{2,\infty}(c)v_t}
\bigl(F(\mathbf{q}_T)-F(\mathbf{r}_t)\bigr),
\label{equ:ib_error_optimality}\\
& E_{2,t}
=\frac12e^{-2\gamma_{2,\infty}(c)v_t}
\sum_{i\in\CS(c)}e^{-2p_i(\mathbf{r}_t)},
\label{equ:ib_error_logistic}\\
& E_{3,t,T}=\sum_{i\notin\CS(c)}
\exp\left[-v_t\bigl(\mathbf{x}_i^\top\mathbf{u}_{2,\infty}(c)
-c\lVert\mathbf{u}_{2,\infty}(c)\rVert_1\bigr)
-p_i(\mathbf{q}_T)\right].
\label{equ:ib_error_nonsupport}
\end{align}
There exists an integer $T_0\ge3$, independent of $T$, such that
$t_T<T$ for every integer $T\ge T_0$ and, for $t_T\le t<T$,
\begin{align}
&\CL_{\mathrm{Adv}}(v_t\mathbf{u}_{2,\infty}(c)+\mathbf{q}_T)
-\CL_{\mathrm{Adv}}(\mathbf{w}_t)
\le E_{1,t,T}+E_{2,t}+E_{3,t,T}.
\label{equ:ib_three_loss_terms}
\end{align}
\end{lem}
\begin{proof}
For $T\ge3$, $0<1/\ln(T+2)<1$. Substituting
$\varepsilon=1/\ln(T+2)$ into the construction in Proposition~\ref{lem:ib_approx} gives
\begin{align}
\mathbf{q}_{\varepsilon}
&=\bar{\mathbf{r}}+\mathbf{v}
\ln\left(\frac{C_5}{1/\ln(T+2)}\right)
=\bar{\mathbf{r}}+\mathbf{v}\ln(C_5\ln(T+2))\overset{\eqref{equ:ib_qT_definition}}{=}\mathbf{q}_T,\nonumber\\
F(\mathbf{q}_T)&\overset{\eqref{equ:ib_approx_statement}}{\le}
\inf_{\mathbf{r}\perp\mathbf{u}_{2,\infty}(c)}F(\mathbf{r})
+\frac1{\ln(T+2)},\nonumber\\
\lVert\mathbf{q}_T\rVert_2
&\overset{\eqref{equ:ib_approx_statement}}{\le} C_6\left(1+\ln\frac1{1/\ln(T+2)}\right)
=C_6(1+\ln\ln(T+2)).\label{equ:ib_steps_qT_substitution}
\end{align}
For every $\mathbf{r}\perp\mathbf{u}_{2,\infty}(c)$, we have
\begin{align}
F(\mathbf{q}_T)-F(\mathbf{r})
&\le F(\mathbf{q}_T)
-\inf_{\mathbf{s}\perp\mathbf{u}_{2,\infty}(c)}F(\mathbf{s})
\overset{\eqref{equ:ib_steps_qT_substitution}}{\le}\frac1{\ln(T+2)}
.\label{equ:ib_qT}
\end{align}

By the definition of $t_T$, we have
\begin{align}
e^{(\ln\ln(T+2))^2}\overset{\eqref{equ:ib_start_index}}{\le} t_T
<e^{(\ln\ln(T+2))^2}+1,\qquad
t_T+2<e^{(\ln\ln(T+2))^2}+3\le4e^{(\ln\ln(T+2))^2}.\label{equ:ib_steps_start_ceiling}
\end{align}
The preliminary bound in Proposition~\ref{lem:ib_preliminary} for $\mathbf r_T$ gives
\begin{align}
\lVert\mathbf{r}_{t_T}\rVert_2
&\overset{\eqref{equ:ib_preliminary}}{\le} C_2\sqrt{\ln(t_T+2)}
\overset{\eqref{equ:ib_steps_start_ceiling}}{\le} C_2\sqrt{(\ln\ln(T+2))^2+\ln4}.\label{equ:ib_start_estimates}
\end{align}
For the component along the $\mathbf{u}_{2,\infty}(c)$ direction, we have
\begin{align}
v_{t_T}
&\overset{\eqref{equ:ib_v_lower}}{\ge}\frac1{\gamma_{2,\infty}(c)}
\ln\left(1+\frac{\eta\gamma_{2,\infty}^2(c)}2t_T\right)
\overset{\eqref{equ:ib_steps_start_ceiling}}{\ge}\frac{(\ln\ln(T+2))^2+\ln(\eta\gamma_{2,\infty}^2(c)/2)}
{\gamma_{2,\infty}(c)}.\label{equ:ib_steps_start_v}
\end{align}
Once the last numerator is positive, these inequalities imply
\begin{align}
\frac{\lVert\mathbf{q}_T\rVert_2}{v_{t_T}}
&\overset{\eqref{equ:ib_steps_qT_substitution},\eqref{equ:ib_steps_start_v}}{\le}\frac{\gamma_{2,\infty}(c)C_6(1+\ln\ln(T+2))}
{(\ln\ln(T+2))^2+\ln(\eta\gamma_{2,\infty}^2(c)/2)}
\rightarrow0, \quad \mathrm{as}\quad T \to \infty.\label{equ:ib_steps_sign_ratio}
\end{align}

Since $\lVert\mathbf{u}_{2,\infty}(c)\rVert_2=1$, the vector
$\mathbf{u}_{2,\infty}(c)$ has at least one nonzero coordinate.
Consequently,
\[
\min_{j:\,u_{2,\infty,j}(c)\ne0}
|u_{2,\infty,j}(c)|>0.
\]
Thus we can choose a fixed integer $T_0\ge3$ such that, for every
integer $T\ge T_0$, we have $t_T<T$ and
\begin{align}
(\ln\ln(T+2))^2
+\ln\left(\frac{\eta\gamma_{2,\infty}^2(c)}2\right)
>0,\quad
\frac{\gamma_{2,\infty}(c)C_6(1+\ln\ln(T+2))}
{(\ln\ln(T+2))^2+\ln(\eta\gamma_{2,\infty}^2(c)/2)}
<\min_{j:\,u_{2,\infty,j}(c)\ne0}
|u_{2,\infty,j}(c)|.
\label{equ:ib_steps_sign_threshold}
\end{align}

Moreover, since we have proved $v_t$ is an increasing function in Proposition~\ref{lem:trainingproj}, thus for $t\ge t_T$, we have
$v_t\ge v_{t_T}$. Hence, at each such nonzero coordinate of $\mathbf{u}_{2,\infty}(c)$, we have
\begin{align*}
\Sign(u_{2,\infty,j}(c))(v_tu_{2,\infty,j}(c)+q_{T,j})
&=v_t|u_{2,\infty,j}(c)|
+\Sign(u_{2,\infty,j}(c))q_{T,j}\\
&\ge v_{t_T}|u_{2,\infty,j}(c)|-\lVert\mathbf{q}_T\rVert_2\overset{\eqref{equ:ib_steps_sign_ratio},\eqref{equ:ib_steps_sign_threshold}}{>}0.
\end{align*}
It follows that
\begin{align}
\Sign(v_tu_{2,\infty,j}(c)+q_{T,j})
=\Sign(u_{2,\infty,j}(c))
\quad\text{if }u_{2,\infty,j}(c)\ne0.\label{equ:ib_comparator_signs}
\end{align}

For the actual iterate, $|a|\ge\Sign(u_{2,\infty,j}(c))a$
at each nonzero coordinate of the direction. Thus
\begin{align}
\lVert\mathbf{w}_t\rVert_1
&=\sum_{j:\,u_{2,\infty,j}(c)\ne0}
|v_tu_{2,\infty,j}(c)+r_{t,j}|
+\sum_{j:\,u_{2,\infty,j}(c)=0}|r_{t,j}|\nonumber\\
&\ge v_t\lVert\mathbf{u}_{2,\infty}(c)\rVert_1
+\sum_{j:\,u_{2,\infty,j}(c)\ne0}\Sign(u_{2,\infty,j}(c))r_{t,j}
+\sum_{j:\,u_{2,\infty,j}(c)=0}|r_{t,j}|.\label{equ:ib_steps_actual_norm}
\end{align}
For the comparison point, the sign equality makes the corresponding
bound an equality:
\begin{align}
\lVert v_t\mathbf{u}_{2,\infty}(c)+\mathbf{q}_T\rVert_1
&\overset{\eqref{equ:ib_comparator_signs}}{=}v_t\lVert\mathbf{u}_{2,\infty}(c)\rVert_1
+\sum_{j:\,u_{2,\infty,j}(c)\ne0}\Sign(u_{2,\infty,j}(c))q_{T,j}
+\sum_{j:\,u_{2,\infty,j}(c)=0}|q_{T,j}|.\label{equ:ib_steps_comparison_norm}
\end{align}
Combining above estimate on the $\ell_1$ norm of $\mathbf w_t$ and comparison point
$v_t \mathbf u_{2,\infty}(c) + \mathbf q_T$, we get
\begin{align}
\mathbf{x}_i^\top\mathbf{w}_t-c\lVert\mathbf{w}_t\rVert_1
&\overset{\eqref{equ:ib_steps_actual_norm}}{\le} v_t\mathbf{x}_i^\top\mathbf{u}_{2,\infty}(c)
+\mathbf{x}_i^\top\mathbf{r}_t
-cv_t\lVert\mathbf{u}_{2,\infty}(c)\rVert_1\nonumber\\
&\quad-c\sum_{j:\,u_{2,\infty,j}(c)\ne0}
\Sign(u_{2,\infty,j}(c))r_{t,j}
-c\sum_{j:\,u_{2,\infty,j}(c)=0}|r_{t,j}|\nonumber\\
&\overset{\eqref{equ:ib_p_function}}{=}v_t\bigl(\mathbf{x}_i^\top\mathbf{u}_{2,\infty}(c)
-c\lVert\mathbf{u}_{2,\infty}(c)\rVert_1\bigr)+p_i(\mathbf{r}_t),\label{equ:ib_score_comparison}
\end{align}
and
\begin{align}
&\mathbf{x}_i^\top(v_t\mathbf{u}_{2,\infty}(c)+\mathbf{q}_T)
-c\lVert v_t\mathbf{u}_{2,\infty}(c)+\mathbf{q}_T\rVert_1\overset{\eqref{equ:ib_steps_comparison_norm},\eqref{equ:ib_p_function}}{=}v_t\bigl(\mathbf{x}_i^\top\mathbf{u}_{2,\infty}(c)
-c\lVert\mathbf{u}_{2,\infty}(c)\rVert_1\bigr)+p_i(\mathbf{q}_T).\label{equ:ib_steps_comparison_score_equality}
\end{align}

By the definition of the support set $\CS(c)$, for every
$i\in\CS(c)$, we have
$\mathbf{x}_i^\top\mathbf{u}_{2,\infty}(c)
-c\lVert\mathbf{u}_{2,\infty}(c)\rVert_1
=\gamma_{2,\infty}(c)$. Since the logistic loss is decreasing in the
margin, we get
\begin{align}
\ln\left(1+e^{-\mathbf{x}_i^\top\mathbf{w}_t+c\lVert\mathbf{w}_t\rVert_1}\right)&\overset{\eqref{equ:ib_score_comparison}}{\ge}\ln\left(1+e^{-\gamma_{2,\infty}(c)v_t-p_i(\mathbf{r}_t)}\right)\nonumber\\
& \ge e^{-\gamma_{2,\infty}(c)v_t-p_i(\mathbf{r}_t)}
-\frac12e^{-2\gamma_{2,\infty}(c)v_t-2p_i(\mathbf{r}_t)},\label{equ:ib_steps_actual_support_loss}
\end{align}
where the last inequality used $\ln(1+x) \ge x - x^2/2$ for $x>0$.
Discarding the adversarial losses contributed by data points not in $\CS(c)$ gives
\begin{align}
\CL_{\mathrm{Adv}}(\mathbf{w}_t)
&\ge\sum_{i\in\CS(c)}
\ln\left(1+e^{-\mathbf{x}_i^\top\mathbf{w}_t+c\lVert\mathbf{w}_t\rVert_1}\right)\nonumber\\
&\overset{\eqref{equ:ib_steps_actual_support_loss}}{\ge}
\sum_{i\in\CS(c)}\left[
 e^{-\gamma_{2,\infty}(c)v_t-p_i(\mathbf{r}_t)}
-\frac12e^{-2\gamma_{2,\infty}(c)v_t-2p_i(\mathbf{r}_t)}
\right]\nonumber\\
& = e^{-\gamma_{2,\infty}(c)v_t}F(\mathbf{r}_t)
- E_{2,t} \label{equ:ib_steps_actual_loss_lower}
\end{align}
For the adversarial loss at the comparison point, use $\ln(1+x) < x$ for $x>0$, we get
\begin{align}
\CL_{\mathrm{Adv}}&(v_t\mathbf{u}_{2,\infty}(c)+\mathbf{q}_T)\overset{\eqref{equ:ib_steps_comparison_score_equality}}{\le}\sum_{i=1}^N
\exp\left[-v_t\bigl(\mathbf{x}_i^\top\mathbf{u}_{2,\infty}(c)
-c\lVert\mathbf{u}_{2,\infty}(c)\rVert_1\bigr)-p_i(\mathbf{q}_T)\right]\nonumber\\
&\quad=e^{-\gamma_{2,\infty}(c)v_t}F(\mathbf{q}_T)
+E_{3,t,T}.\label{equ:ib_steps_comparison_loss_upper}
\end{align}

Combining the two bounds in \eqref{equ:ib_steps_comparison_loss_upper} and \eqref{equ:ib_steps_actual_loss_lower} gives
\begin{align*}
\CL_{\mathrm{Adv}}(v_t\mathbf{u}_{2,\infty}(c)+\mathbf{q}_T)
-\CL_{\mathrm{Adv}}(\mathbf{w}_t)&\le
\bigl[e^{-\gamma_{2,\infty}(c)v_t}F(\mathbf{q}_T)+E_{3,t,T}\bigr]
-\bigl[e^{-\gamma_{2,\infty}(c)v_t}F(\mathbf{r}_t)-E_{2,t}\bigr]\\
&\quad=
 e^{-\gamma_{2,\infty}(c)v_t}
 \bigl[F(\mathbf{q}_T)-F(\mathbf{r}_t)\bigr]
 +E_{2,t}+E_{3,t,T}\\
&\quad=E_{1,t,T}+E_{2,t}+E_{3,t,T}.
\end{align*}
\end{proof}

In the following Lemma~\ref{lem:ib_loss_gap_optimality} to Lemma \ref{lem:ib_loss_gap_nonsupport}, we will bound the $E_{1,t,T}$, $E_{2,t}$ and $E_{3,t,T}$ terms by some absolute constant separately.

\begin{lem}\label{lem:ib_loss_gap_optimality}
Let $T_0$ be as in Lemma~\ref{lem:ib_loss_decomposition}.
There is a constant $C_7\ge0$, independent of $T$, such that, for every
$T\ge T_0$, we have
\[
\sum_{t=t_T}^{T-1}E_{1,t,T}\le C_7.
\]
\end{lem}
\begin{proof}
Since $\mathbf{r}_t\perp\mathbf{u}_{2,\infty}(c)$,
\begin{align}
e^{-\gamma_{2,\infty}(c)v_t}\bigl(F(\mathbf{q}_T)-F(\mathbf{r}_t)\bigr)
&\overset{\eqref{equ:ib_qT}}{\le}\frac{e^{-\gamma_{2,\infty}(c)v_t}}{\ln(T+2)}
\overset{\eqref{equ:ib_v_lower}}{\le}\frac{C_1}{(t+1)\ln(T+2)}.\label{equ:ib_steps_F_pointwise}
\end{align}
Consequently, we have
\begin{align}
\sum_{t=t_T}^{T-1}E_{1,t,T}
=\sum_{t=t_T}^{T-1}e^{-\gamma_{2,\infty}(c)v_t}
\bigl(F(\mathbf{q}_T)-F(\mathbf{r}_t)\bigr)&\overset{\eqref{equ:ib_steps_F_pointwise}}{\le}\frac{C_1}{\ln(T+2)}\sum_{t=t_T}^{T-1}\frac1{t+1}\nonumber\\
&\le\frac{C_1(1+\ln T)}{\ln(T+2)}\nonumber\\
&\le 2C_1=C_7,\label{equ:ib_sum_F}
\end{align}
where the second inequality used $\sum_{t=0}^{T-1}\frac1{t+1} \le 1+\ln T$.
\end{proof}

\begin{lem}\label{lem:ib_loss_gap_logistic}
Let $T_0$ be as in Lemma~\ref{lem:ib_loss_decomposition}.
There is a constant $C_8\ge0$, independent of $T$, such that, for every
$T\ge T_0$, we have
\[
\sum_{t=t_T}^{T-1}E_{2,t}\le C_8.
\]
\end{lem}
\begin{proof}
By Proposition~\ref{lem:ib_preliminary}, we have
\begin{align}
e^{-2p_i(\mathbf{r}_t)}
&\le e^{2|p_i(\mathbf{r}_t)|}
\overset{\eqref{equ:ib_steps_p_norm}}{\le} e^{2M\lVert\mathbf{r}_t\rVert_2}
\overset{\eqref{equ:ib_preliminary}}{\le} e^{2MC_2\sqrt{\ln(t+2)}}.\label{equ:ib_steps_quadratic_score}
\end{align}
The inequality
$(\sqrt{x/2}-\sqrt2MC_2)^2\ge0$ gives
$2MC_2\sqrt{x}\le x/2+2M^2C_2^2$. Therefore
\begin{align}
\frac12 e^{-2\gamma_{2,\infty}(c)v_t}
\sum_{i\in\CS(c)}e^{-2p_i(\mathbf{r}_t)}
&\overset{\eqref{equ:ib_v_lower},\eqref{equ:ib_steps_quadratic_score}}{\le}\frac{NC_1^2}{2(t+1)^2}e^{2MC_2\sqrt{\ln(t+2)}}\nonumber\\
&\le\frac{NC_1^2e^{2M^2C_2^2}(t+2)^{1/2}}{2(t+1)^2}\nonumber\\
&=\frac{C_8}{3(t+1)^{3/2}},\label{equ:ib_steps_quadratic_pointwise}
\end{align}
where $C_8 = NC_1^2e^{2M^2C_2^2}$. Summing the above bound yields
\begin{align}
\sum_{t=t_T}^{T-1}E_{2,t}
=\sum_{t=t_T}^{T-1}\frac12e^{-2\gamma_{2,\infty}(c)v_t}
\sum_{i\in\CS(c)}e^{-2p_i(\mathbf{r}_t)}&\overset{\eqref{equ:ib_steps_quadratic_pointwise}}{\le}\frac{C_8}{3}\sum_{t=t_T}^{T-1}(t+1)^{-3/2}\nonumber\\
&\le \frac{C_8}{3}
\left(1+\int_1^\infty x^{-3/2}\,\mathrm dx\right)
=\frac{C_8}{3}(1+2)=C_8.\label{equ:ib_sum_quadratic}
\end{align}
\end{proof}

\begin{lem}\label{lem:ib_loss_gap_nonsupport}
Let $T_0$ be as in Lemma~\ref{lem:ib_loss_decomposition}.
There is a constant $C_9\ge0$, independent of $T$, such that, for every
$T\ge T_0$,
\[
\sum_{t=t_T}^{T-1}E_{3,t,T}\le C_9.
\]
\end{lem}
\begin{proof}
If $[N]\setminus\CS(c)=\varnothing$, then $E_{3,t,T}=0$ and we take
$C_9=0$. For the remaining case, define
\begin{align}
\delta=\min_{i\notin\CS(c)}
\left\{\frac{\mathbf{x}_i^\top\mathbf{u}_{2,\infty}(c)
-c\lVert\mathbf{u}_{2,\infty}(c)\rVert_1}
{\gamma_{2,\infty}(c)}-1\right\}>0.\label{equ:ib_gap_delta}
\end{align}
For such an index in $[N]\setminus\CS(c)$, we have
\begin{align}
\exp\left[-v_t\bigl(\mathbf{x}_i^\top\mathbf{u}_{2,\infty}(c)
-c\lVert\mathbf{u}_{2,\infty}(c)\rVert_1\bigr)-p_i(\mathbf{q}_T)\right] & \overset{\eqref{equ:ib_gap_delta},\eqref{equ:ib_steps_p_norm}}{\le} e^{-(1+\delta)\gamma_{2,\infty}(c)v_t}
e^{M\lVert\mathbf{q}_T\rVert_2} \nonumber\\
\nonumber\\
&=\bigl(e^{-\gamma_{2,\infty}(c)v_t}\bigr)^{1+\delta}
e^{M\lVert\mathbf{q}_T\rVert_2}
\overset{\eqref{equ:ib_v_lower}} \nonumber\\
\nonumber\\
&\le\frac{C_1^{1+\delta}e^{M\lVert\mathbf{q}_T\rVert_2}}
{(t+1)^{1+\delta}}.\label{equ:ib_steps_nonsupport_pointwise}
\end{align}
Moreover, we have the following bound:
\begin{align}
\sum_{t=t_T}^{\infty}(t+1)^{-1-\delta}
&\le\sum_{t=t_T}^{\infty}\int_t^{t+1}x^{-1-\delta}\,\mathrm dx
=\int_{t_T}^{\infty}x^{-1-\delta}\,\mathrm dx
=\frac{t_T^{-\delta}}{\delta}
&\overset{\eqref{equ:ib_steps_start_ceiling}}{\le}\frac1\delta e^{-\delta(\ln\ln(T+2))^2}.\label{equ:ib_steps_nonsupport_integral}
\end{align}
It follows that
\begin{align}
\sum_{t=t_T}^{T-1}E_{3,t,T}&=\sum_{t=t_T}^{T-1}\sum_{i\notin\CS(c)}
\exp\left[-v_t\bigl(\mathbf{x}_i^\top\mathbf{u}_{2,\infty}(c)
-c\lVert\mathbf{u}_{2,\infty}(c)\rVert_1\bigr)-p_i(\mathbf{q}_T)\right]\nonumber\\
&\overset{\eqref{equ:ib_steps_nonsupport_pointwise}}{\le} NC_1^{1+\delta}e^{M\lVert\mathbf{q}_T\rVert_2}
\sum_{t=t_T}^{\infty}(t+1)^{-1-\delta}\nonumber\\
&\overset{\eqref{equ:ib_steps_nonsupport_integral},\eqref{equ:ib_steps_qT_substitution}}{\le}\frac{NC_1^{1+\delta}}{\delta}
\exp\left[MC_6(1+\ln\ln(T+2))
-\delta(\ln\ln(T+2))^2\right].\label{equ:ib_sum_nonsupport}
\end{align}

Completing the square in the exponent gives
\begin{align}
&MC_6(1+\ln\ln(T+2))-\delta(\ln\ln(T+2))^2\nonumber\\
&\quad=MC_6+\frac{M^2C_6^2}{4\delta}
-\delta\left(\ln\ln(T+2)-\frac{MC_6}{2\delta}\right)^2
\le MC_6+\frac{M^2C_6^2}{4\delta}.\label{equ:ib_steps_complete_square}
\end{align}
Thus we can choose $C_9$ as
\begin{align}
C_9=
\begin{cases}
\displaystyle\frac{NC_1^{1+\delta}}{\delta}
\exp\left(MC_6+\frac{M^2C_6^2}{4\delta}\right),
&[N]\setminus\CS(c)\ne\varnothing,\\
0,&[N]\setminus\CS(c)=\varnothing.
\end{cases}\label{equ:ib_steps_const_9}
\end{align}
\end{proof}

Now we can prove Proposition~\ref{lem:ib_loss_gap}.

\begin{proof}[Proof of Proposition~\ref{lem:ib_loss_gap}]
Take $T_0$ from Lemma~\ref{lem:ib_loss_decomposition}. Then $t_T<T$ and
\eqref{equ:ib_three_loss_terms} holds for every $T\ge T_0$ and
$t_T\le t<T$. Summing this inequality and applying
Lemmas~\ref{lem:ib_loss_gap_optimality},
\ref{lem:ib_loss_gap_logistic}, and~\ref{lem:ib_loss_gap_nonsupport}
gives
\begin{align*}
&\sum_{t=t_T}^{T-1}
\bigl[\CL_{\mathrm{Adv}}(v_t\mathbf{u}_{2,\infty}(c)+\mathbf{q}_T)
-\CL_{\mathrm{Adv}}(\mathbf{w}_t)\bigr]\\
&\qquad\le
\sum_{t=t_T}^{T-1}E_{1,t,T}
+\sum_{t=t_T}^{T-1}E_{2,t}
+\sum_{t=t_T}^{T-1}E_{3,t,T}
\le C_7+C_8+C_9.
\end{align*}
The constants are independent of $T$, as required.
\end{proof}

\subsection{Refined estimate on
\texorpdfstring{$\lVert\mathbf r_t\rVert_2$}{the norm of r_t}}
\label{sec:rertl2}

In the following, we provide a refined estimate of $\lVert\mathbf r_t\rVert_2$ based on the estimates established in Section~\ref{sec:clc}. In particular, we show that the preliminary estimates of $\lVert\mathbf r_t\rVert_2$ with $\mathcal O(\sqrt{\ln(t+2)})$ can be improved to  $\mathcal O(\ln\ln t)$

\begin{prop}\label{lem:ib_residual}
There is a fixed $C_{\mathrm r}>0$ which is independent of $t$, such that
\begin{align}
\lVert\mathbf{r}_t\rVert_2\le C_{\mathrm r}\ln\ln t,
\qquad t\ge3.\label{equ:ib_residual_final}
\end{align}
\end{prop}
\begin{proof}
Fix $T_0,C_7,C_8,C_9$ from Proposition~\ref{lem:ib_loss_gap}, and let
$T\ge T_0$. Let the vector $\mathbf{q}_T$ be the same as \eqref{equ:ib_qT_definition}. Subtracting the decompositions at two consecutive
times gives
\begin{align}
&\mathbf{r}_{t+1}-\mathbf{r}_t
\overset{\eqref{equ:decwur}}{=}\mathbf{w}_{t+1}-\mathbf{w}_t
-(v_{t+1}-v_t)\mathbf{u}_{2,\infty}(c),\nonumber\\
&\langle\mathbf{r}_t-\mathbf{q}_T,\mathbf{u}_{2,\infty}(c)\rangle
=\langle\mathbf{r}_t,\mathbf{u}_{2,\infty}(c)\rangle
-\langle\mathbf{q}_T,\mathbf{u}_{2,\infty}(c)\rangle=0,\label{equ:ib_steps_projected_update}
\end{align}
where the second equality comes from both $\mathbf{r}_t$ and $\mathbf{q}_T$ are orthogonal to $\mathbf u_{2,\infty}(c)$. Thus, we have
\begin{align}
\langle\mathbf{r}_t-\mathbf{q}_T,\mathbf{r}_{t+1}-\mathbf{r}_t\rangle
&=\langle\mathbf{r}_t-\mathbf{q}_T,\mathbf{w}_{t+1}-\mathbf{w}_t\rangle-(v_{t+1}-v_t)
\langle\mathbf{r}_t-\mathbf{q}_T,\mathbf{u}_{2,\infty}(c)\rangle\nonumber\\
&=\langle\mathbf{r}_t-\mathbf{q}_T,\mathbf{w}_{t+1}-\mathbf{w}_t\rangle.\label{equ:ib_steps_projected_cross_identity}
\end{align}
The comparison point $v_t\mathbf{u}_{2,\infty}(c)+\mathbf{q}_T$ differs from the trajectories of $\{\mathbf w_t\}$ generated by \eqref{equ:gd} in that
\begin{align}
v_t\mathbf{u}_{2,\infty}(c)+\mathbf{q}_T-\mathbf{w}_t
\overset{\eqref{equ:decwur}}{=}v_t\mathbf{u}_{2,\infty}(c)+\mathbf{q}_T
-(v_t\mathbf{u}_{2,\infty}(c)+\mathbf{r}_t)
=\mathbf{q}_T-\mathbf{r}_t.\label{equ:ib_steps_comparison_displacement}
\end{align}
Consequently, the subgradient inequality reads
\begin{align}
\CL_{\mathrm{Adv}}(v_t\mathbf{u}_{2,\infty}(c)+\mathbf{q}_T)
-\CL_{\mathrm{Adv}}(\mathbf{w}_t)&\ge\left\langle
\frac{\mathbf{w}_t-\mathbf{w}_{t+1}}{\eta},
v_t\mathbf{u}_{2,\infty}(c)+\mathbf{q}_T-\mathbf{w}_t
\right\rangle\nonumber\\
&=\frac1\eta\langle\mathbf{r}_t-\mathbf{q}_T,
\mathbf{w}_{t+1}-\mathbf{w}_t\rangle,\label{equ:ib_steps_projected_subgradient}
\end{align}
where the first inequality used the fact that $\CL_{\mathrm{Adv}}$ is a convex function. 

Combining the two identities in \eqref{equ:ib_steps_projected_cross_identity} and \eqref{equ:ib_steps_projected_subgradient}, and multiplying by $\eta$ yields
\begin{align}
\langle\mathbf{r}_t-\mathbf{q}_T,\mathbf{r}_{t+1}-\mathbf{r}_t\rangle
\le\eta\bigl[\CL_{\mathrm{Adv}}(v_t\mathbf{u}_{2,\infty}(c)+\mathbf{q}_T)
-\CL_{\mathrm{Adv}}(\mathbf{w}_t)\bigr].\label{equ:ib_orthogonal_cross}
\end{align}
Since $\mathbf{r}_{t+1}-\mathbf{r}_t\perp\mathbf{u}_{2,\infty}(c)$, we get
\begin{align}
\lVert\mathbf{w}_{t+1}-\mathbf{w}_t\rVert_2^2
&\overset{\eqref{equ:ib_steps_projected_update}}{=}\left\lVert(v_{t+1}-v_t)\mathbf{u}_{2,\infty}(c)
+(\mathbf{r}_{t+1}-\mathbf{r}_t)\right\rVert_2^2\nonumber\\
&=(v_{t+1}-v_t)^2+\lVert\mathbf{r}_{t+1}-\mathbf{r}_t\rVert_2^2\nonumber\\
&\ge\lVert\mathbf{r}_{t+1}-\mathbf{r}_t\rVert_2^2.\label{equ:ib_steps_projection_contracts}
\end{align}
Expanding the distance to the fixed $\mathbf{q}_T$ gives
\begin{align}
\lVert\mathbf{r}_{t+1}-\mathbf{q}_T\rVert_2^2
&=\left\lVert(\mathbf{r}_t-\mathbf{q}_T)
+(\mathbf{r}_{t+1}-\mathbf{r}_t)\right\rVert_2^2\nonumber\\
&=\lVert\mathbf{r}_t-\mathbf{q}_T\rVert_2^2
+2\langle\mathbf{r}_t-\mathbf{q}_T,\mathbf{r}_{t+1}-\mathbf{r}_t\rangle
+\lVert\mathbf{r}_{t+1}-\mathbf{r}_t\rVert_2^2\nonumber\\
&\overset{\eqref{equ:ib_orthogonal_cross},\eqref{equ:ib_steps_projection_contracts}}{\le}\lVert\mathbf{r}_t-\mathbf{q}_T\rVert_2^2
+2\eta\bigl[\CL_{\mathrm{Adv}}(v_t\mathbf{u}_{2,\infty}(c)+\mathbf{q}_T)
-\CL_{\mathrm{Adv}}(\mathbf{w}_t)\bigr]+\lVert\mathbf{w}_{t+1}-\mathbf{w}_t\rVert_2^2.\label{equ:ib_final_recursion}
\end{align}

Summing \eqref{equ:ib_final_recursion} from $t=t_T$ to $T-1$ gives:
\begin{align}
\lVert\mathbf{r}_T-\mathbf{q}_T\rVert_2^2
-\lVert\mathbf{r}_{t_T}-\mathbf{q}_T\rVert_2^2&=\sum_{t=t_T}^{T-1}
\bigl[\lVert\mathbf{r}_{t+1}-\mathbf{q}_T\rVert_2^2
-\lVert\mathbf{r}_t-\mathbf{q}_T\rVert_2^2\bigr]\nonumber\\
&\quad\overset{\eqref{equ:ib_final_recursion}}{\le}2\eta\sum_{t=t_T}^{T-1}
\bigl[\CL_{\mathrm{Adv}}(v_t\mathbf{u}_{2,\infty}(c)+\mathbf{q}_T)
-\CL_{\mathrm{Adv}}(\mathbf{w}_t)\bigr]
+\sum_{t=t_T}^{T-1}\lVert\mathbf{w}_{t+1}-\mathbf{w}_t\rVert_2^2\nonumber\\
&\quad\overset{\eqref{equ:ib_cumulative_loss_gap}}{\le}2\eta(C_7+C_8+C_9)
+\sum_{t=0}^{\infty}\lVert\mathbf{w}_{t+1}-\mathbf{w}_t\rVert_2^2 \nonumber\\
&\quad\overset{\eqref{equ:ib_summable_updates}}{\le}2\eta(C_7+C_8+C_9)+C_4:=C_{10}.\label{equ:ib_steps_final_telescope}
\end{align}
Then the triangle inequality gives
\begin{align}
\lVert\mathbf{r}_T\rVert_2
&\le\lVert\mathbf{r}_T-\mathbf{q}_T\rVert_2+\lVert\mathbf{q}_T\rVert_2\nonumber\\
&\overset{\eqref{equ:ib_steps_final_telescope}}{\le}\sqrt{\lVert\mathbf{r}_{t_T}-\mathbf{q}_T\rVert_2^2+C_{10}}
+\lVert\mathbf{q}_T\rVert_2\nonumber\\
&\le\lVert\mathbf{r}_{t_T}-\mathbf{q}_T\rVert_2+\sqrt{C_{10}}
+\lVert\mathbf{q}_T\rVert_2\nonumber\\
&\le\lVert\mathbf{r}_{t_T}\rVert_2+2\lVert\mathbf{q}_T\rVert_2
+\sqrt{C_{10}}\nonumber\\
&\overset{\eqref{equ:ib_start_estimates},\eqref{equ:ib_steps_qT_substitution}}{\le} C_2\sqrt{(\ln\ln(T+2))^2+\ln4}
+2C_6(1+\ln\ln(T+2))+\sqrt{C_{10}}.\label{equ:ib_final_distance}
\end{align}
Since $\ln\ln(T+2)>0$, we have
\begin{align*}
&\sqrt{(\ln\ln(T+2))^2+\ln4}
\le\ln\ln(T+2)+\sqrt{\ln4}\le(1+\sqrt{\ln4})(1+\ln\ln(T+2)),\nonumber\\
\end{align*}
which gives
\begin{align}
\lVert\mathbf{r}_T\rVert_2
&\overset{\eqref{equ:ib_final_distance}}{\le}\bigl[C_2(1+\sqrt{\ln4})+2C_6+\sqrt{C_{10}}\bigr]
(1+\ln\ln(T+2))\nonumber\\
\nonumber\\
& \le \bigl[C_2(1+\sqrt{\ln4})+2C_6+\sqrt{C_{10}}\bigr] \left(1+\frac{1+\ln2}{\ln\ln3}\right) \ln\ln(T+2)\nonumber\\
\nonumber\\
&=C_{11}(1+\ln\ln(T+2))\nonumber\\
\nonumber\\
& \le \tilde{C_{\mathbf r}}\ln\ln (T) \quad \mathrm{for}\quad  T\ge T_0.\label{equ:ib_steps_C11_absorption}
\end{align}

For the remaining integers $3\le t<T_0$, orthogonality and the
initial update bound give
\begin{align*}
\lVert\mathbf{r}_t\rVert_2
&\le\sqrt{v_t^2+\lVert\mathbf{r}_t\rVert_2^2}
=\lVert\mathbf{w}_t\rVert_2
\overset{\eqref{equ:ib_steps_iterate_norm}}{\le}\eta NMt\le\eta NMT_0 \le\frac{\eta NMT_0}{\ln\ln3}\ln\ln t
\le \hat{C_{\mathrm r}}\ln\ln t.
\end{align*}
Then we can choose $C_{\mathbf r} = \max\{\tilde{C_{\mathbf r}},\hat{C_{\mathbf r}}\}$, this proves the bound for every $t\ge3$.
\end{proof}

\subsection{Proof of  Theorem~\ref{thm:main1}}
\begin{proof}[Proof of Theorem~\ref{thm:main1}]
According to Lemma~\ref{lem:ib_preliminary}, we have
$\CL_{\mathrm{Adv}}(\mathbf{w}_t)\to0$ as $t \to \infty$.
We also have
\begin{align}
v_t&\le\lVert\mathbf{w}_t\rVert_2
=\sqrt{v_t^2+\lVert\mathbf{r}_t\rVert_2^2}
\le v_t+\lVert\mathbf{r}_t\rVert_2.\label{equ:ib_steps_direction_norm}
\end{align}
Here $v_t>0$. Subtracting $v_t$ yields
$0\le\lVert\mathbf{w}_t\rVert_2-v_t\le\lVert\mathbf{r}_t\rVert_2$.
The normalized difference is
\begin{align}
\frac{\mathbf{w}_t}{\lVert\mathbf{w}_t\rVert_2}
-\mathbf{u}_{2,\infty}(c)
&\overset{\eqref{equ:decwur}}{=}\left(\frac{v_t}{\lVert\mathbf{w}_t\rVert_2}-1\right)
\mathbf{u}_{2,\infty}(c)
+\frac{\mathbf{r}_t}{\lVert\mathbf{w}_t\rVert_2}.\label{equ:ib_steps_direction_decomposition}
\end{align}
Applying the triangle inequality and then the two lemmas,
\begin{align}
\left\lVert\frac{\mathbf{w}_t}{\lVert\mathbf{w}_t\rVert_2}
-\mathbf{u}_{2,\infty}(c)\right\rVert_2
&\overset{\eqref{equ:ib_steps_direction_decomposition}}{\le}\left|\frac{v_t}{\lVert\mathbf{w}_t\rVert_2}-1\right|
\lVert\mathbf{u}_{2,\infty}(c)\rVert_2
+\frac{\lVert\mathbf{r}_t\rVert_2}{\lVert\mathbf{w}_t\rVert_2}\nonumber\\
&=\frac{\lVert\mathbf{w}_t\rVert_2-v_t+\lVert\mathbf{r}_t\rVert_2}
{\lVert\mathbf{w}_t\rVert_2}\nonumber\\
&\overset{\eqref{equ:ib_steps_direction_norm}}{\le}\frac{2\lVert\mathbf{r}_t\rVert_2}{v_t}\nonumber\\
&\le\frac{2\gamma_{2,\infty}(c)C_{\mathrm r}\ln\ln t}
{\ln(1+\frac{\eta\gamma_{2,\infty}^2(c)}2t)},\label{equ:ib_direction_final}
\end{align}
where the last inequality comes from Proposition~\ref{lem:trainingproj} and Proposition~\ref{lem:ib_residual}.

Moreover, we have
\begin{align}
\ln\left(1+\frac{\eta\gamma_{2,\infty}^2(c)}2t\right)
&\ge\min\left\{1,\frac{\eta\gamma_{2,\infty}^2(c)}2\right\}\ln(1+t)
=\frac{\ln(1+t)}{C_1}
\ge\frac{\ln t}{C_1}.\label{equ:ib_steps_log_denominator_final}
\end{align}

Since the denominator is positive, we have
\begin{align*}
\left\lVert\frac{\mathbf{w}_t}{\lVert\mathbf{w}_t\rVert_2}
-\mathbf{u}_{2,\infty}(c)\right\rVert_2
\overset{\eqref{equ:ib_direction_final},\eqref{equ:ib_steps_log_denominator_final}}{\le}2\gamma_{2,\infty}(c)C_1C_{\mathrm r}
\frac{\ln\ln t}{\ln t}
=\mathcal O \left(\frac{\ln\ln t}{\ln t}\right).
\end{align*}
This finished the proof.
\end{proof}

\subsection{Proof of Proposition~\ref{prop:tight_rate}}

\begin{prop}\label{prop:ib_tight}
Let $d=N=2$, $y_1=y_2=1$, and
\begin{align*}
c=\frac14,\qquad
\mathbf{x}_1=\left(\frac12,\frac14\right),\qquad
\mathbf{x}_2=\left(\frac12,\frac12\right).
\end{align*}
This dataset is robustly linearly separable, with
$\hat{\mathbf{w}}(c)=(4,0)$,
$\mathbf{u}_{2,\infty}(c)=(1,0)$, and
$\gamma_{2,\infty}(c)=1/4$.
For every fixed $\eta>0$, the iteration from $\mathbf{w}_0=0$ satisfies
\begin{align*}
\lim_{t\to\infty}\frac{\ln t}{\ln\ln t}
\left\lVert\frac{\mathbf{w}_t}{\lVert\mathbf{w}_t\rVert_2}
-\mathbf{u}_{2,\infty}(c)\right\rVert_2=1.
\end{align*}
\end{prop}
\begin{proof}
Recall that
$
\gamma_\infty
=\max_{\lVert\mathbf w\rVert_1=1}
\min_{i=1,2}\mathbf x_i^\top\mathbf w.
$
For any $\mathbf w=(w_1,w_2)$ with
$\lVert \mathbf w \rVert_1=1$, we have
\begin{align*}
\min_{i=1,2}\mathbf x_i^\top\mathbf w
&\le \mathbf x_1^\top\mathbf w
=\frac12w_1+\frac14w_2\le \frac12|w_1|+\frac14|w_2|
\le \frac12(|w_1|+|w_2|)
=\frac12.
\end{align*}
Taking the maximum over all such $\mathbf w$ gives
$\gamma_\infty\le1/2$. Conversely, the vector $\mathbf w=(1,0)$ satisfies
$\lVert\mathbf w\rVert_1=1$ and we have
\[
\mathbf x_1^\top\mathbf w
=\mathbf x_2^\top\mathbf w=\frac12,
\]
hence $\gamma_\infty\ge1/2$.
Combining the two bounds yields
$\gamma_\infty=1/2>c=1/4$, thus according to Lemma~\ref{lem:expl}, we have the proposed data set is robustly linearly separable.

According to the definition, every solution of \eqref{equ:rsvm}
satisfies 
\[
1 \le \min_{\lVert \boldsymbol\delta_i \rVert_{\infty} \le c} \langle \mathbf{x}_i+\boldsymbol{\delta}_i,\mathbf{w}\rangle = \langle  \mathbf{x}_i, \mathbf{w}\rangle - c \lVert \mathbf{w} \rVert_1,
\]
take $\mathbf{x}_1$ into above, we get
\begin{align*}
1
&\le\frac12w_1+\frac14w_2-\frac14(|w_1|+|w_2|)=\frac14w_1+\frac14(w_1-|w_1|)+\frac14(w_2-|w_2|)
\le\frac14w_1.
\end{align*}
Hence, for any solution $\mathbf{w}$ of \eqref{equ:rsvm}, we need $w_1\ge4$ and
$\lVert\mathbf{w}\rVert_2^2=w_1^2+w_2^2\ge16$.
On the otherside, for the vector $\mathbf w = (4,0)$, we have:
\begin{align*}
\langle  \mathbf{x}_1, \mathbf{w}\rangle - c \lVert \mathbf{w} \rVert_1 &= \left\langle\left(\frac12,\frac14\right),(4,0)\right\rangle
-\frac14\lVert(4,0)\rVert_1=1,\\
\langle  \mathbf{x}_2, \mathbf{w}\rangle - c \lVert \mathbf{w} \rVert_1 &= \left\langle\left(\frac12,\frac12\right),(4,0)\right\rangle
-\frac14\lVert(4,0)\rVert_1=1.
\end{align*}
Thus for this data set, the solution of \eqref{equ:rsvm} is unique, and we have
\begin{align*}
\hat{\mathbf{w}}(c)=(4,0),\qquad
\mathbf{u}_{2,\infty}(c)=\frac{(4,0)}4=(1,0),\qquad
\gamma_{2,\infty}(c)=\frac14,\qquad \CS(c)=\{1,2\}.
\end{align*}

Let $\mathbf s^* = (s_1^*,s_2^*) \in \partial \lVert \hat{\mathbf w}(c)\lVert_1$. By Proposition~\ref{prop:dualrsvm}, 
there exist $\alpha_1,\alpha_2\ge0$ and
$\mathbf s^*\in\partial\lVert\hat{\mathbf w}(c)\rVert_1$ such that
\[
\hat{\mathbf w}(c)
=\alpha_1(\mathbf x_1-c\mathbf s^*)
+\alpha_2(\mathbf x_2-c\mathbf s^*).
\]
Since $\hat{\mathbf w}(c)=(4,0)$, we have
$s_1^*=1$ and $s_2^*\in[-1,1]$, which gives
\begin{align}
4&=\alpha_1\left(\frac12-\frac14\right)
+\alpha_2\left(\frac12-\frac14\right)
=\frac14(\alpha_1+\alpha_2),\nonumber\\
0&=\alpha_1\left(\frac14-\frac14s_2^{*}\right)
+\alpha_2\left(\frac12-\frac14s_2^{*}\right)=\frac{1-s_2^{*}}4(\alpha_1+\alpha_2)+\frac14\alpha_2
\ge\frac14\alpha_2\ge0.\label{equ:ib_lower_kkt}
\end{align}
It follows that
\begin{align*}
\alpha_1=16,\quad \alpha_2=0,\quad \mathrm{and} \quad s_2^{*}=1.
\end{align*}
Consequently, we have
\begin{align}
\mathbf{z}_1(c) = \mathbf{x}_1 - c \mathbf s^*
&=\left(\frac12,\frac14\right)-\frac14(1,1)
=\left(\frac14,0\right), \quad
\mathbf{z}_2(c) = \mathbf{x}_2 - c \mathbf s^*
=\left(\frac12,\frac12\right)-\frac14(1,1)
=\left(\frac14,\frac14\right).\label{equ:ib_lower_vectors}
\end{align}

At $\mathbf{w}_0=0$, recall from the definition of $\lambda_{i,t}$ in \eqref{equ:deflambda},  we have $\lambda_{1,0}=\lambda_{2,0}=1/2$.
We choose $\mathbf s_0=\mathbf0$, which belongs to
$\partial\lVert\mathbf w_0\rVert_1=[-1,1]^2$.
The first update is then
\begin{align*}
\mathbf{w}_1
&=\frac\eta2(\mathbf{x}_1+\mathbf{x}_2)
=\frac\eta2\left(1,\frac34\right)
=\left(\frac\eta2,\frac{3\eta}8\right).
\end{align*}
Both coordinates are positive. Whenever $\mathbf{w}_t$ has positive
coordinates, its norm subgradient is $(1,1)$, and the update adds
positive multiples of $(1/4,0)$ and $(1/4,1/4)$. Induction shows
that both coordinates remain positive for all $t\ge1$. Therefore
$v_t=w_{t,1}$, $\mathbf{r}_t=(0,w_{t,2})$, and
\begin{align}
\mathbf{x}_1^\top\mathbf{w}_t-c\lVert\mathbf{w}_t\rVert_1
&=\frac12v_t+\frac14w_{t,2}-\frac14(v_t+w_{t,2})=\frac{v_t}4,\nonumber\\
\mathbf{x}_2^\top\mathbf{w}_t-c\lVert\mathbf{w}_t\rVert_1
&=\frac12v_t+\frac12w_{t,2}-\frac14(v_t+w_{t,2})
=\frac{v_t+w_{t,2}}4.\label{equ:ib_steps_lower_scores}
\end{align}
Substitution into the update gives, for $t\ge1$, we get
\begin{align}
\lambda_{1,t}=\frac1{1+e^{v_t/4}},
\quad\lambda_{2,t}=\frac1{1+e^{(v_t+w_{t,2})/4}},\quad
v_{t+1}-v_t=\frac\eta4(\lambda_{1,t}+\lambda_{2,t}),\quad 
w_{t+1,2}-w_{t,2}=\frac\eta4\lambda_{2,t}.\label{equ:ib_lower_recursion}
\end{align}
For $x\ge0$, $0<e^{-x}\le1$ implies
\begin{align}
\frac12e^{-x}\le\frac1{1+e^x}\le e^{-x},\quad \mathrm{and} \quad
\frac1{1+e^x}\le\frac12.\label{equ:ib_steps_lower_logistic_bounds}
\end{align}
Since $w_{t,2}>0$, we have
$0<\lambda_{2,t}\le\lambda_{1,t}\le1/2$. It follows that
\begin{align}
\frac{v_{t+1}-v_t}{4}
&\overset{\eqref{equ:ib_lower_recursion}}{=}\frac\eta{16}(\lambda_{1,t}+\lambda_{2,t})
\ge\frac\eta{16}\lambda_{1,t}
\overset{\eqref{equ:ib_steps_lower_logistic_bounds}}{\ge}\frac\eta{32}e^{-v_t/4},\quad
\frac{v_{t+1}-v_t}{4} \le\frac\eta{16}(2\lambda_{1,t})
\overset{\eqref{equ:ib_steps_lower_logistic_bounds}}{\le}\frac\eta8e^{-v_t/4},\label{equ:ib_steps_lower_v_increment_sides1}
\end{align}
and
\begin{align}
0<\frac{v_{t+1}-v_t}{4}
&\overset{\eqref{equ:ib_steps_lower_logistic_bounds}}{\le}\frac\eta{16}\left(\frac12+\frac12\right)=\frac\eta{16}.\label{equ:ib_steps_lower_v_increment_sides2}
\end{align}

Using the fact that $xe^x > x$ for $x>0$, we have
\begin{align}
&e^{v_{t+1}/4}-e^{v_t/4}
=e^{v_t/4}\left(e^{(v_{t+1}-v_t)/4}-1\right)\ge e^{v_t/4}\left(\frac{v_{t+1}-v_t}{4}\right)
\overset{\eqref{equ:ib_steps_lower_v_increment_sides1}}{\ge} \frac\eta{32},\label{equ:ib_steps_lower_exp_v_lower}\\
&e^{v_{t+1}/4}-e^{v_t/4}
\le e^{v_t/4}\left(\frac{v_{t+1}-v_t}{4}\right)
e^{(v_{t+1}-v_t)/4}\overset{\eqref{equ:ib_steps_lower_v_increment_sides1}}{\le}\frac\eta8e^{\eta/16}.\label{equ:ib_steps_lower_exp_v_upper}
\end{align}
Since $v_1=\eta/2$, we have
\begin{align}
e^{v_t/4}
&=e^{v_1/4}+\sum_{s=1}^{t-1}
\left(e^{v_{s+1}/4}-e^{v_s/4}\right),\quad
e^{\eta/8}+\frac\eta{32}(t-1)
\overset{\eqref{equ:ib_steps_lower_exp_v_lower}}{\le} e^{v_t/4}
\overset{\eqref{equ:ib_steps_lower_exp_v_upper}}{\le} e^{\eta/8}+\frac\eta8e^{\eta/16}(t-1).\label{equ:ib_lower_exp_v}
\end{align}
Let
$C_{12}=\min\{e^{\eta/8},\eta/32\}$ and $C_{13}=\max\{e^{\eta/8},(\eta/8)e^{\eta/16}\}$, then we have
\begin{align}
e^{\eta/8}+\frac\eta{32}(t-1)
\ge C_{12}t,\quad
e^{\eta/8}+\frac\eta8e^{\eta/16}(t-1)
\le C_{13}t.\label{equ:ib_steps_lower_v_constants}
\end{align}
Thus we have
\begin{align}
C_{12}t \le e^{v_t/4} \le C_{13}t,\quad
\ln C_{12}+\ln t \le\frac{v_t}4\le\ln C_{13}+\ln t,\quad
4\ln C_{12}\le v_t-4\ln t\le4\ln C_{13}.\label{equ:ib_lower_v_bounds}
\end{align}
Reciprocation of the first inequality in \eqref{equ:ib_lower_v_bounds} gives
\begin{align}
\frac1{C_{13}t} \le e^{-v_t/4} \le \frac1{C_{12}t}.\label{equ:ib_steps_lower_v_reciprocal}
\end{align}

For the second coordinate, we have
\begin{align}
\frac{w_{t+1,2}-w_{t,2}}4
&\overset{\eqref{equ:ib_lower_recursion}}{=}\frac\eta{16}\lambda_{2,t}
\overset{\eqref{equ:ib_steps_lower_logistic_bounds}}{\ge}\frac\eta{32}e^{-(v_t+w_{t,2})/4},\nonumber\\
\frac{w_{t+1,2}-w_{t,2}}4
&\overset{\eqref{equ:ib_steps_lower_logistic_bounds}}{\le}\frac\eta{16}e^{-(v_t+w_{t,2})/4},\qquad
0<\frac{w_{t+1,2}-w_{t,2}}4\overset{\eqref{equ:ib_steps_lower_logistic_bounds}}{\le}\frac\eta{32}.\label{equ:ib_lower_increment_r}
\end{align}
Applying $x\le e^x-1\le xe^x$, we get
\begin{align}
&e^{w_{t+1,2}/4}-e^{w_{t,2}/4}
=e^{w_{t,2}/4}
\left(e^{(w_{t+1,2}-w_{t,2})/4}-1\right)
\ge e^{w_{t,2}/4}\frac{w_{t+1,2}-w_{t,2}}4
\overset{\eqref{equ:ib_lower_increment_r}}{\ge} \frac\eta{32}e^{-v_t/4},\label{equ:ib_steps_lower_exp_r_lower}\\
&e^{w_{t+1,2}/4}-e^{w_{t,2}/4}
\le e^{w_{t,2}/4}\frac{w_{t+1,2}-w_{t,2}}4
e^{(w_{t+1,2}-w_{t,2})/4}
\overset{\eqref{equ:ib_lower_increment_r}}{\le} \frac\eta{16}e^{\eta/32}e^{-v_t/4}.\label{equ:ib_steps_lower_exp_r_upper}
\end{align}
The reciprocal bounds therefore imply
\begin{align}
\frac{\eta}{32C_{13}t}
&\overset{\eqref{equ:ib_steps_lower_v_reciprocal}}{\le}\frac\eta{32}e^{-v_t/4}
\overset{\eqref{equ:ib_steps_lower_exp_r_lower}}{\le} e^{w_{t+1,2}/4}-e^{w_{t,2}/4}
\overset{\eqref{equ:ib_steps_lower_exp_r_upper}}{\le}\frac\eta{16}e^{\eta/32}e^{-v_t/4}
\overset{\eqref{equ:ib_steps_lower_v_reciprocal}}{\le}\frac{\eta e^{\eta/32}}{16C_{12}t}.\label{equ:ib_lower_exp_r}
\end{align}
Summing from $s=1$ to $t-1$, and using $w_{1,2}=3\eta/8$,
\begin{align}
e^{w_{t,2}/4}-e^{3\eta/32}
=\sum_{s=1}^{t-1}
&\left(e^{w_{s+1,2}/4}-e^{w_{s,2}/4}\right),\quad \mathrm{which \,implies}\nonumber\\
\frac{\eta}{32C_{13}}\sum_{s=1}^{t-1}\frac1s
&\overset{\eqref{equ:ib_lower_exp_r}}{\le} e^{w_{t,2}/4}-e^{3\eta/32}
\overset{\eqref{equ:ib_lower_exp_r}}{\le}\frac{\eta e^{\eta/32}}{16C_{12}}\sum_{s=1}^{t-1}\frac1s.\label{equ:ib_steps_lower_r_telescope}
\end{align}
For $t\ge2$, using the inequality $\ln t \le \sum^{t-1}_{s=1}1/s \le 1+\ln t$, we get 
\begin{align}
\frac{\eta}{32C_{13}}\ln t
\overset{\eqref{equ:ib_steps_lower_r_telescope}}{\le} e^{w_{t,2}/4}-e^{3\eta/32}
\overset{\eqref{equ:ib_steps_lower_r_telescope}}{\le}\frac{\eta e^{\eta/32}}{16C_{12}}(1+\ln t).\label{equ:ib_lower_r_sums}
\end{align}
Define
\begin{align}
C_{14}=\frac{\eta}{32C_{13}},\qquad
C_{15}=\frac{e^{3\eta/32}}{\ln2}
+\frac{\eta e^{\eta/32}}{16C_{12}}
\left(1+\frac1{\ln2}\right).\label{equ:ib_const_14_15}
\end{align}
We have
\begin{align}
C_{14}\ln t \le e^{w_{t,2}/4}
\le\left[\frac{e^{3\eta/32}}{\ln2}
+\frac{\eta e^{\eta/32}}{16C_{12}}
\left(\frac1{\ln2}+1\right)\right]\ln t
\overset{\eqref{equ:ib_const_14_15}}{=}C_{15}\ln t.\label{equ:ib_steps_lower_r_constants}
\end{align}
Taking logarithms of above, we get
\begin{align}
4\ln C_{14}
&\le w_{t,2}-4\ln\ln t\le4\ln C_{15}.\label{equ:ib_lower_r_bounds}
\end{align}
Since these constants depend only on $\eta$, in conclusion, we get
\begin{align}
v_t&\overset{\eqref{equ:ib_lower_v_bounds}}{=}4\ln t+\CO_\eta(1),\qquad
\lVert\mathbf{r}_t\rVert_2=w_{t,2}
\overset{\eqref{equ:ib_lower_r_bounds}}{=}4\ln\ln t+\CO_\eta(1).\label{equ:ib_steps_lower_asymptotics}
\end{align}

Finally, compute the implicit bias error explicitly gives
\begin{align}
\left\lVert\frac{\mathbf{w}_t}{\lVert\mathbf{w}_t\rVert_2}
-(1,0)\right\rVert_2^2
&=\left(\frac{v_t}{\sqrt{v_t^2+w_{t,2}^2}}-1\right)^2
+\left(\frac{w_{t,2}}{\sqrt{v_t^2+w_{t,2}^2}}\right)^2
=2-\frac2{\sqrt{1+(w_{t,2}/v_t)^2}}.\label{equ:ib_lower_direction}
\end{align}
Taking \eqref{equ:ib_lower_direction} into the calculation, we get
\begin{align*}
&\frac{\ln t}{\ln\ln t}
\left\lVert\frac{\mathbf{w}_t}{\lVert\mathbf{w}_t\rVert_2}-(1,0)\right\rVert_2\quad=
\frac{\left\lVert\mathbf{w}_t/\lVert\mathbf{w}_t\rVert_2-(1,0)\right\rVert_2}
{w_{t,2}/v_t}\,
\frac{w_{t,2}/\ln\ln t}{v_t/\ln t}
\rightarrow 1,
\end{align*}
as $t \to \infty$. This finished the proof.
\end{proof}

\newpage
\section{Proofs in Section~\ref{sec:op}}\label{app_s4}

\subsection{Proof of Theorem~\ref{prop:counter_exp_opt}}\label{app_ppceo}
\underline{We first prove the \textbf{\textit{no acceleration by large step sizes}} part of Theorem~\ref{prop:counter_exp_opt}}.
\begin{proof}
We establish the lower bound for gradient
descent with $\mathbf{w}_0=\mathbf{0}$ , which already suffices to rule out a general
acceleration guarantee. For an integer time budget $T$, let $q=4T^2$ and
$d_T=q^2+1=16T^4+1$. Define the one-sample data set
\begin{align*}
\mathcal{D}_T=\{(\mathbf{x},1)\},\qquad
\mathbf{x}=
\left(2,\underbrace{q^{-1},\ldots,q^{-1}}_{q^2\ \text{coordinates}}\right)
\in\BR^{d_T}.
\end{align*}
Note that this data set depends only on $T$, and we have
$\lVert\mathbf{x}\rVert_2=\sqrt{5}$. Fix any adversarial radius $0<c<2$ such that $T>64/c$, and any $\eta>0$.
Since $cq=4cT^2>256T$, we have $q^{-1}<c$.
Lemma~\ref{lem:one-sample-robust-margin} therefore gives
\begin{align*}
\gamma_{2,\infty}(c)=2-c>0\quad \mathrm{and}\quad
\mathbf{u}_{2,\infty}(c)=(1,0,\ldots,0).
\end{align*}
In particular, $\mathcal{D}_T$ is robustly separable under $\ell_{\infty}$ adversarial attack with radius $c$. 

For data set $\mathcal{D}_T$, write the robust margin of $\mathbf x$ with respect to $\mathbf w$ as
\begin{align}
M_t:=\mathbf{x}^\top\mathbf{w}_t-c\lVert\mathbf{w}_t\rVert_1,\label{equ:def_Mt}
\end{align}
which gives
$\CL_{\mathrm{Adv}}(\mathbf{w}_t)=\ln\left(1+\exp(-M_t)\right)$. The \eqref{equ:gd} dynamics can be written as
\begin{align}
\mathbf{w}_{t+1} & = \mathbf{w}_t - \eta \,\partial_{\mathbf{w}}\mathcal{L}_{\mathrm{Adv}}(\mathbf{w}_t) \nonumber\\
& =\mathbf{w}_t+\eta\lambda_{t}
\left(\mathbf{x}-c\,\Sign(\mathbf{w}_t)\right),
\quad \mathrm{where} \quad
\lambda_{t}=\frac{1}{1+\exp(M_t)} < 1. \label{equ:gdlambdaM}
\end{align}
Since the last $q^2$ coordinates of $\mathbf{x}$ and $\mathbf{w}_0$ are the same, we have the last $q^2$ coordinates of $\mathbf{w}_t$ are also the same. Denote
\begin{align}
\mathbf{w}_t=
\left(v_t,\eta c r_t,\ldots,\eta c r_t\right),\label{equ:v1r1}
\end{align}
where $v_t=\langle\mathbf{w}_t,\mathbf{u}_{2,\infty}(c)\rangle$.
According to \eqref{equ:gdlambdaM}, the first update gives 
\begin{align}
v_1 = v_0 + \eta \lambda_0 (2 - c \cdot 0) = \eta, \quad \mathrm{and} \quad 
r_1  = \left(\eta c r_0 + \eta \lambda_0 (q^{-1} - c \cdot 0)\right)/(\eta c) = \beta/2,\label{equ:v_1w_1}
\end{align}
where $\beta=(cq)^{-1}<1$. 

Through explicitly writing down \eqref{equ:gdlambdaM} with the data in $\mathcal{D}_T$, we have 
\begin{align}\label{equ:ceo_recursion}
v_{t+1}&=v_t+\eta(2-c)\lambda_{t},\nonumber \\ 
r_{t+1}&=r_t+\lambda_{t}\left(\beta-\Sign(r_t)\right)
\quad \mathrm{for}\quad t\geq1.
\end{align}
Since $2-c>0$, this implies $v_t>0$ for every $t\geq1$.
Consequently, under above notation, we have
\begin{align}\label{equ:ceo_margin}
M_t=\mathbf{x}^\top\mathbf{w}_t-c\lVert\mathbf{w}_t\rVert_1= (2-c)v_t+\eta c^2q^2\left(\beta r_t-|r_t|\right).
\end{align}
For the first term in \eqref{equ:ceo_margin}, by using the recursion formula in \eqref{equ:ceo_recursion}, we have
\begin{align}\label{equ:ceo_first_coordinate}
(2-c)v_t & = (2-c) v_1 + \eta (2-c)^2 \left(\sum^{t-1}_{s=1}\lambda_t\right)\nonumber\\
\nonumber\\
& \overset{\eqref{equ:gdlambdaM},\eqref{equ:v1r1}}{\le} (2-c) \eta + \eta (2-c)^2 (t-1) \leq 4\eta t,\qquad t\geq1.
\end{align}

Let $A:=\ln(8cqT)$ and $H:=(\eta cq)/16$. The assumptions on $c$ and $T$ give
\begin{align}\label{equ:ceo_parameters}
cq = 4cT^2 >256T,\qquad
\beta T<\frac{1}{256},\qquad
0<A\leq\ln(64T^3)\leq4\ln(3T).
\end{align}
We consider the following two cases:

\textit{(1).}  $H<A$, i.e., small step size $\eta < 16 A/cq$ is used. 
Since $\beta<1$, the second term in \eqref{equ:ceo_margin} is
nonpositive. Hence, we have 
\begin{align*}
M_T \overset{\eqref{equ:ceo_margin},\eqref{equ:ceo_first_coordinate}}{\leq} 4\eta T
<\frac{64AT}{cq}
=\frac{16A}{cT}
 \le \frac{64\ln(3T)}{cT}.
\end{align*}
Since the function \(u \to \ln(3u)/u\) is decreasing for
\(u>e/3\). As \(T>64/c>32\), we have
\[
M_T<\frac{64}{c}\,
       \frac{\ln(3(64/c))}{64/c}
     =\ln\frac{192}{c}.
\]
Consequently, we have
\[
L^{\rm Adv}(w_T)
=\ln(1+e^{-M_T})
>\ln\left(1+\frac{c}{192}\right).
\]

\textit{(2).} $H\geq A$, which is the regime where a rather large step size can be used. To facilitate the analysis, we compare the recursion for $r_t$ in
\eqref{equ:ceo_recursion} with an auxiliary sequence obtained by
replacing $\lambda_t$ in \eqref{equ:ceo_recursion}  with $1$, which gives the following sequence $\{\bar{r}_t\}$:
\begin{align}
\bar r_1=\frac{\beta}{2},\qquad
\bar r_{t+1}=\bar r_t+\beta-\Sign(\bar r_t).\label{equ:def_barrt}
\end{align}
Since $\beta <1/(256\,T)$, direct induction gives
\begin{align*}
\bar r_{2k-1}=\frac{4k-3}{2}\beta>0,\qquad
\bar r_{2k}=-1+\frac{4k-1}{2}\beta<0,
\end{align*}
whenever the corresponding index is at most $T$.
In particular, we have $|\bar r_t|\geq\beta/2$ for $1\leq t\leq T$.

We first show that the error between $r_t$ and $\bar{r}_t$ belongs to $\CO(\beta)$ under the current regime. In fact, we have:
\begin{align}\label{equ:ceo_comparison}
|r_t-\bar r_t|\leq2(t-1)\exp(-H)\leq\frac{\beta}{4},
\qquad 1\leq t\leq T.
\end{align}
The second inequality follows from
\begin{align*}
2T\exp(-H)\leq2T\exp(-A)=\frac{1}{4cq}=\frac{\beta}{4}.
\end{align*}
The first inequality of \eqref{equ:ceo_comparison} can be proved by induction on $t$. It holds at $t=1$ because $r_1=\bar r_1$ from \eqref{equ:v_1w_1} and \eqref{equ:def_barrt}.
Suppose it holds at some $t<T$.
Then $r_t$ and $\bar r_t$ have the same sign, and $|r_t|\geq\beta/4$, which is because $\bar{r}_t \ge \beta/2$.
Since $\beta\leq1/2$, we obtain
\begin{align*}
\eta c^2q^2\left(\beta r_t-|r_t|\right)
&\leq-\eta c^2q^2(1-\beta)|r_t| \quad (\mathrm{Since }\, \beta r_t-|r_t| < (\beta-1) |r_t| )\\
\\
&\leq-\frac{\eta c^2q^2\beta}{8}
=-\frac{\eta cq}{8}=-2H.
\end{align*}
Combining this with \eqref{equ:ceo_margin} and
\eqref{equ:ceo_first_coordinate}, and using $4\eta T\leq H$, yields
\begin{align*}
M_t\leq4\eta t-2H\leq-H,
\qquad
1-\lambda_{t}
=\frac{\exp(M_t)}{1+\exp(M_t)}
\leq\exp(-H).
\end{align*}
Subtracting the two recursions and using
$\Sign(r_t)=\Sign(\bar r_t)$, we conclude that
\begin{align*}
|r_{t+1}-\bar r_{t+1}|
&\leq|r_t-\bar r_t|
+(1-\lambda_{1,t})|\beta-\Sign(r_t)|\\
\\
&\leq2(t-1)\exp(-H)+2\exp(-H)
=2t\exp(-H).
\end{align*}
This proves \eqref{equ:ceo_comparison} by induction.

In particular, $|r_T|\geq\beta/4$, so the same estimate in
\eqref{equ:ceo_margin} gives $M_T\leq-H$.
Therefore,
\begin{align*}
\CL_{\mathrm{Adv}}(\mathbf{w}_T)
=\ln\left(1+\exp(-M_T)\right)
\geq-M_T\geq H\geq A>\ln\left(1+\frac{c}{192}\right).
\end{align*}
 This completes the proof.
\end{proof}

\underline{We then prove the \textbf{\textit{no phase transition time}} part of Theorem~\eqref{prop:counter_exp_opt} in the following}. 

\begin{proof}
We use the same data set $\mathcal{D}_T$ and the same initialization
and subgradient choice as above.
Fix any admissible $T,c$ and any $\eta>0$.
In the following, $T$ and $q=4T^2$ remain fixed.
Recall from \eqref{equ:ceo_parameters} that
\begin{align*}
0<\beta<\frac12,\qquad cq-1>(2-c)^2.
\end{align*}
Moreover,
\eqref{equ:ceo_recursion} gives
\begin{align*}
r_{t+1}
&=
\begin{cases}
r_t-\lambda_t(1-\beta), & r_t>0,\\
\lambda_t\beta, & r_t=0,\\
r_t+\lambda_t(\beta+1), & r_t<0.
\end{cases}
\end{align*}
Combining with $0<\lambda_t<1$ and $0<\beta<1/2$, this implies $|r_{t+1}| \leq \max\{|r_t|,1+\beta\}$ by using the inequality $|a-b| \le \max \{a,b\}$. Since $|r_1|=\beta/2<1+\beta$, $|r_{t+1}| \leq \max\{|r_t|,1+\beta\}$ yields
\begin{align}\label{equ:ceo_bounded_r}
|r_t|\leq1+\beta,\qquad t\geq1.
\end{align}
Moreover, $v_t$ is increasing and must tend to infinity.
Indeed, suppose that $v_t\leq V$ for all $t$. Then
$M_t\leq(2-c)V$ by \eqref{equ:ceo_margin}, and hence
\[
\lambda_t\geq
\frac{1}{1+\exp((2-c)V)}
=:\delta>0.
\]
By \eqref{equ:ceo_recursion}, a positive lower bound of $v_t$ implies
\begin{align}\label{equ:vtinfty}
v_t
\geq v_1+\eta(2-c)\delta(t-1)
\rightarrow\infty,\quad t \to \infty,
\end{align}
which contradicts the boundedness of $v_t$.
Combining $v_t\to\infty$ with \eqref{equ:ceo_margin} and
\eqref{equ:ceo_bounded_r}, we obtain
\begin{align*}
M_t
\geq(2-c)v_t-\eta c^2q^2(1+\beta)^2
\rightarrow\infty,\quad t \to \infty.
\end{align*}
Consequently, as $t\ \to \infty$, we have
\begin{align}\label{equ:ceo_long_time_limits}
\CL_{\mathrm{Adv}}(\mathbf{w}_t)\rightarrow0,\quad
\lambda_t\rightarrow0,\quad \mathrm{and} \quad
\sum_{t=1}^{\infty}\lambda_t=\infty,
\end{align}
where the last assertion follows from the recursion for $v_t$
and $v_t\to\infty$ by \eqref{equ:vtinfty}.

Since the function $f(r) = \beta r-|r|$ is $(1+\beta)$-Lipschitz,
\eqref{equ:ceo_recursion} and \eqref{equ:ceo_margin} imply
\begin{align*}
|M_{t+1}-M_t| & \le  (2-c) \big| v_{t+1} -v_t \big| + \eta^2 cq\big|f(r_{t+1}) - f(r_t) \big| \\
\\
& \le \eta(2-c)^2 \lambda_t + (1+\beta)\big|r_{t+1} - r_t \big| \quad (\mathrm{Since}\, \eqref{equ:ceo_recursion}\, \mathrm{and\, Lipschit\, property\, of }\,  f(r)) \\
\\
& = \eta(2-c)^2 \lambda_t + (1+\beta) \lambda_t |\beta - \Sign(t_t)| \quad (\mathrm{From}\, \eqref{equ:ceo_recursion}) \\
\\
& \leq
\eta\left((2-c)^2+c^2q^2(1+\beta)^2\right)\lambda_t
\rightarrow0.\quad (\mathrm{Since}\, |\beta - \Sign(t_t)| \le 1+\beta)
\end{align*}
Together with $M_t\to\infty$, this gives
\begin{align}\label{equ:ceo_lambda_ratio}
\lambda_{t+1}/\lambda_t
=
\frac{1+\exp(M_t)}{1+\exp(M_{t+1})}
\rightarrow 1,\quad t \to \infty.
\end{align}

We now show that there are infinitely many indices $t$ such that
$r_t>0$ and $r_{t+1}<0$.
The sequence $r_t$ cannot be eventually strictly positive.
Otherwise, \eqref{equ:ceo_recursion} would give
$r_{t+1}=r_t-(1-\beta)\lambda_t$ for all sufficiently large $t$,
which contradicts $\sum_t\lambda_t=\infty$ as shown in \eqref{equ:ceo_long_time_limits}.
The same argument shows that $r_t$ cannot be eventually strictly
negative.
Moreover, if $r_t=0$, then
\begin{align*}
r_{t+1}=\beta\lambda_t>0,\qquad
r_{t+2}
=\beta\lambda_t-(1-\beta)\lambda_{t+1}<0
\end{align*}
for all sufficiently large $t$, by \eqref{equ:ceo_lambda_ratio}
and $\beta<1/2$.
Thus, whether $r_t$ equals zero finitely or infinitely often,
there are infinitely many indices $t$ such that
$r_t>0$ and $r_{t+1}<0$.

For each sufficiently large index $\tau$ such that $r_{\tau}>0$ and $r_{\tau+1} < 0$, we also have $r_{\tau+2}>0$.
This is because,
\begin{align*}
r_{\tau+1}
=r_\tau-(1-\beta)\lambda_\tau
>-(1-\beta)\lambda_\tau,
\end{align*}
and therefore
\begin{align*}
r_{\tau+2}
=r_{\tau+1}+(1+\beta)\lambda_{\tau+1}
>-(1-\beta)\lambda_{\tau}+(1+\beta)\lambda_{\tau+1}>0,
\end{align*}
where the last inequality follows from \eqref{equ:ceo_lambda_ratio}.
Hence, there are infinitely many indices satisfying
\begin{align}\label{equ:ceo_sign_crossings}
r_{\tau}>0,\qquad r_{\tau+1}<0,\qquad r_{\tau+2}>0.
\end{align}

For any index $\tau$ satisfying \eqref{equ:ceo_sign_crossings},
the recursions in \eqref{equ:ceo_recursion} give
\begin{align*}
v_{\tau+2}-v_{\tau}
&=\eta(2-c)(\lambda_{\tau}+\lambda_{\tau+1}),\\
r_{\tau+2}-r_{\tau}
&=-(1-\beta)\lambda_{\tau}+(1+\beta)\lambda_{\tau+1}.
\end{align*}
Since $r_{\tau},r_{\tau+2}>0$, it follows from \eqref{equ:ceo_margin} that
\begin{align*}
M_{\tau+2}-M_{\tau}
&=(2-c)(v_{\tau+2}-v_{\tau})
-\eta c^2q^2(1-\beta)(r_{\tau+2}-r_{\tau}).
\end{align*}
Substituting the two identities above, we obtain
\begin{align*}
\frac{M_{\tau+2}-M_{\tau}}{\eta\lambda_{\tau}}
=
(2-c)^2\left(1+\frac{\lambda_{\tau+1}}{\lambda_{\tau}}\right)+
c^2q^2(1-\beta)
\left((1-\beta)-(1+\beta)\frac{\lambda_{\tau+1}}{\lambda_{\tau}}\right).
\end{align*}
By \eqref{equ:ceo_lambda_ratio} and recall that we have $\beta = 1/cq$, the right-hand side converges to
\begin{align*}
2(2-c)^2-2c^2q^2\beta(1-\beta)
=
2\left((2-c)^2-(cq-1)\right)<0.
\end{align*}
Since $\eta\lambda_t>0$, we have $M_{\tau+2}<M_{\tau}$ for infinitely
many $\tau$. Therefore,
\begin{align*}
\CL_{\mathrm{Adv}}(\mathbf{w}_{\tau+2})
>
\CL_{\mathrm{Adv}}(\mathbf{w}_{\tau})
\end{align*}
for infinitely many $\tau$.
For each such $\tau$, at least one of the two updates from $\tau$ to
$\tau+2$ strictly increases the loss. Consequently,
\begin{align*}
\forall s\in\mathbb{N},\quad
\exists \tau\geq s:\quad
\CL_{\mathrm{Adv}}(\mathbf{w}_{\tau+1})
>
\CL_{\mathrm{Adv}}(\mathbf{w}_{\tau}).
\end{align*}
Thus, no finite time exists after which the loss remains
non-increasing. This completes the proof.
\end{proof}
\newpage
\section{Appendix of Section \ref{sec:er}}\label{app_3}
\subsection{Proof of Proposition \ref{prop:stand_adv}}

\paragraph{High-Level Overview of the Proof.} We will analysis the dynamics given by Lemma \eqref{lem:gd_linquad}, which is the explicit form of the adversarial training dynamics for $(x_t,y_t)$. To study this dynamics, we first introduce the following time horizon, which is motivated by Lemma 6 in \cite{ahn2023learning}:
\begin{align}\label{equ:time_horizon}
s:= \inf \{ t \ge 0 \mid x_t <0\,\, \mathrm{or}\,\, q_ty_t \le 1 \}    
\end{align}
The definition of $q_t$ is presented in \eqref{equ:q_t}. In other worlds, for all $t \le s$, the trajectories $(x_t,y_t)$ lies in the linear region, and $x_t \ge 0$. Moreover, for small $\eta$, the  initialization  given by \eqref{equ:eos_reg} must lies in the linear region, thus $s > 0$ and is well-defined.  It turns out that in the region $t \le s$, the following value
\begin{align}\label{equ:Dt}
D_t : = y^2_t - q^2_t    
\end{align}
is nearly conversed. Since the  initialization  given by \eqref{equ:eos_reg} makes $D_0 = \Omega(1/\eta)$, thus for $t < s$, $y_t$ keeps a magnitude of $\CO(1/\sqrt{\eta})$, which result in $x_t$ significantly decreases in each iteration according to \eqref{equ:lin_reg}. 
In Lemma \ref{lem:seta}, we will prove $s = \CO(1/\eta)$. This results in the training trajectories enter a small tube with $|x_s| = \CO(\eta)$ as shown in Lemma \ref{lem:qsent}. Moreover, in Lemma \ref{lem:inv_qsent}, we prove that once the trajectory enters this tube, it will remain in this tube for the rest of the training time. After the trajectory enter the tube, we will show the $y$-coordinate of the trajectory gradually be compressed to zero and the trajectory future enter the quadratic region, then we use Lemma \ref{lem:sharpness_quad} to prove the upper bound on the sharpness when $t$ tends to infinity.


We first note that as in \eqref{equ:advloss}, the inner max problem of \eqref{equ:advhuber} has an explicit solution:
\begin{align}
\max_{|\delta| \le c} \ell\left((x+\delta)y\right) = \ell \left((x+c\,\Sign(x))\, y\right).
\end{align}
Thus \eqref{equ:advloss} has an explicit formula:
\begin{align*}
\max_{|\delta| \le c} \ell\bigl((x+\delta)y\bigr)
=
\begin{cases}
\dfrac{\bigl((x+c\,\Sign(x))\,y\bigr)^2}{2},
& \bigl|(x+c\,\Sign(x))y\bigr| \le 1,\\[6pt]
\bigl|(x+c\,\Sign(x))y\bigr| - \dfrac{1}{2},
& \bigl|(x+c\,\Sign(x))\, y\bigr| > 1.
\end{cases}
\end{align*}

In the following, we define the region where $\bigl|(x+c\,\Sign(x))y\bigr| \le 1$ holds as the \textit{quadratic region}, and the region where $\bigl|(x+c\,\Sign(x))y\bigr| > 1$ holds as the \textit{linear region}. 

To simplify the notation, we will also denote
\begin{align}\label{equ:q_t}
q_t:=|x+c\,\Sign(x)| =  |x_t| + c.
\end{align}

\begin{lem} \label{lem:gd_linquad}
If $y_t>0$, then in the linear region, the gradient descent dynamic can be written as:
\begin{align}\label{equ:lin_reg}
& x_{t+1} = x_t - \eta\,\Sign(x_t)\, y_t, \nonumber\\
& y_{t+1} = y_t - \eta q_t.
\end{align}
While in the quadratic region, the gradient descent dynamic can be written as:
\begin{align}\label{equ:quad_reg}
& x_{t+1} = (1-\eta\,y_t^2)x_t - \eta\,\Sign(x_t)\,c\,y_t^2 , \nonumber\\
& y_{t+1} =(1 -\eta\,q_t^2)y_t.
\end{align}
\end{lem}

\begin{proof}
Since we assume $y_t>0$, we have
\begin{align*}
\bigl|(x_t+c\,\Sign(x_t))y_t\bigr| = \bigl|x_t+c\,\Sign(x_t)\bigr| y_t = q_t y_t.
\end{align*}

We first consider the linear region $\bigl|(x_t+c\,\Sign(x_t))y_t\bigr| > 1$.  
In this case we have
\begin{align*}
\max_{|\delta| \le c} \ell\left((x+\delta)y\right) =     \bigl|(x+c\,\Sign(x))y\bigr| - \dfrac{1}{2} = q_ty_t - \frac{1}{2},
\end{align*}
and 
\begin{align*}
y_{t+1} = y_t - \eta\, \partial_y \max_{|\delta| \le c} \ell\left((x_t+\delta)y_t\right) = y_t -\eta q_t,    
\end{align*}
this finished the $y$-part in \eqref{equ:lin_reg}. For the $x$-part, we have
\begin{align*}
x_{t+1}  = x_t -  \eta\,  \partial_x \max_{|\delta| \le c} \ell\left((x_t+\delta)y_t\right) & = x_t - \eta\, (\partial_xq_t) \, y_t\\
& =  x_t - \eta\, \Sign(x_t) \, y_t,
\end{align*}
and this finished the $x$-part of the \eqref{equ:lin_reg}.

Next we consider the quadratic region $\bigl|(x_t+c\,\Sign(x))y_t\bigr| \le 1$. In this case, we have
\begin{align*}
\max_{|\delta| \le c} \ell\left((x_t+\delta)y_t\right) =    \dfrac{\bigl((x_t+c\,\Sign(x_t))\,y_t\bigr)^2}{2} = \frac{q^2_t y_t^2}{2},
\end{align*}
and 
\begin{align*}
y_{t+1} = y_t - \eta\, \partial_y\,  \max_{|\delta| \le c} \ell\left((x_t+\delta)y_t\right) = \left( 1- \eta\,q_t^2\right)y_t, 
\end{align*}
this finished the $y$-part in \eqref{equ:quad_reg}. For the $x$-part, we have
\begin{align*}
x_{t+1} & = x_t - \eta\, \partial_x\,  \max_{|\delta| \le c} \ell\left((x_t+\delta)y_t\right)  = x_t - \eta\, (\partial_x q_t^2)\,y_t^2/2.
\end{align*}
Note that
\begin{align*}
\partial_x q_t^2/2 = q_t \partial_x q_t   = \Sign(x_t)\,q_t = \Sign(x_t)\, \left( |x_t| + c \right) = x_t + \Sign(x_t)\, c,
\end{align*}
thus 
\begin{align*}
x_{t+1}  =  x_t - \eta\, (\partial_x q_t^2)\,y_t/2 & = x_t - \eta\, \left(x_t + \Sign(x_t)\, c\right)\,y_t^2\\
& = \left( 1- \eta\,y_t^2\right)x_t - \eta\,\Sign(x_t)\,c\,y^2_t,
\end{align*}
this finished the the $x$-part of the \eqref{equ:quad_reg}.
\end{proof}

\begin{lem}\label{lem:sharpness_quad}
If $(x_y,y_t)$ lies in the quadratic region, we have
\begin{align*}
\lambda^{c}_t \le \left( |y_t|+q_t \right)^2.    
\end{align*}
\end{lem}

\begin{proof}
We first calculate the Hessian matrix $\nabla^2 \CL_{c}(x_t,y_t) = \nabla^2\left(\max_{|\delta| \le c} \ell\left((x_t+\delta)y_t\right)\right)$ in the quadratic region.  We have
\begin{align*}
\nabla^2\left(\max_{|\delta| \le c} \ell\left((x_t+\delta)y_t\right)\right) & = \nabla^2\left(\dfrac{\bigl((x_t+c\,\Sign(x_t))\,y_t\bigr)^2}{2}\right)  = \nabla^2 \left(\dfrac{q_t^2\,y^2_t}{2}\right) 
\end{align*}
Note that we have
\begin{itemize}
\item $\nabla_{xx}\left(\dfrac{q_t^2\,y^2_t}{2}\right) = \nabla_{xx}(q^2_t/2) \cdot y^2_t =  y^2_t$ 
\item $\nabla_{yy}\left(\dfrac{q_t^2\,y^2_t}{2}\right) = q^2_t/2 \cdot \nabla_{yy}(y^2_t) =  q^2_t$ 
\item $\nabla_{xy}\left(\dfrac{q_t^2\,y^2_t}{2}\right) = \partial_x \left(q^2_ty_t \right) = 2\, \Sign(x_t)\,q_t\,y_t$ 
\end{itemize}
Combining above, we get
\begin{align*}
H_t := \nabla^2\left(\max_{|\delta| \le c} \ell\left((x_t+\delta)y_t\right)\right) = \begin{bmatrix}
y^2_t & 2\, \Sign(x_t)\,q_t\,y_t \\
\\
2\, \Sign(x_t)\,q_t\,y_t & q^2_t
\end{bmatrix}.
\end{align*}
Moreover, since
\begin{align*}
\lambda^{c}_t \le \lVert H_t \rVert_{\infty}:= \max \{ y^2_t+2\,q_t\,y_t,\ q^2_t+ 2\,q_t\,y_t\} \le (|y_t|+q_t)^2,   
\end{align*}
this finished the proof.
\end{proof}

\begin{assu}\label{ass:bmbm} We assume the following assumptions hold in the training process:
\begin{itemize}
\item(Boundary-avoidance) The training trajectories do not hit the non-smooth set
\begin{align*}
x_t = 0,\ \left( |x_t|+c \right)y_t =1 
\end{align*}
\item(Magnitudes of Adversarial Radius) We assume $c$ is fixed and  $c \ne 0$.
\end{itemize}
\end{assu}

Note that the boundary-avoidance part of Assumption \ref{ass:bmbm} is to avoid the cases where the solution of inner-max problem of \eqref{equ:advhuber} is not unique when $x=0$, and the Hessian on the boundary points $\left( |x_t|+c \right)y_t =1 $ is not well-defined. Similar assumptions are also needed in \cite{ahn2023learning}.

\begin{lem}\label{lem:seta}
Under Assumption \ref{ass:bmbm}, let 
\begin{align}\label{equ:enttime}
s:= \inf \{ t \ge 0 \mid x_t <0\,\, \mathrm{or}\,\, q_ty_t \le 1 \},   \tag{Entrance Time}  
\end{align}
then for sufficient small step size $\eta$, we have $s \in \CO(1/\eta)$.
\end{lem}

\begin{proof}
For any $k \le s-1$, according to the definition of \eqref{equ:enttime}, we have
\begin{align*}
    x_k > 0,\quad q_k = x_k + c,\quad q_k\,y_k >1,
\end{align*}
thus the trajectory lies in the linear region, and the dynamics as defined in \eqref{equ:lin_reg} can be written as
\begin{align}\label{equ:lin2}
q_{k+1} = q_k - \eta\,y_k,\quad
y_{k+1} = y_k - \eta\, q_k.
\end{align}
Define 
\begin{align*}
D_k := y^2_k -q^2_k.    
\end{align*}

We first compute $D_0$, where $(x_0,y_0)$ is initialize
according to \eqref{equ:eos_reg}. Since in this case, $q_0 = x_0 + c$, we have
\begin{align*}
D_0 = y_0^2 - q_0^2 = (y_0^2 - x^2_0) -2c\,x_0 - c^2  = (2+\delta)/\eta - 2c\,x_0 - c^2.
\end{align*}
Since $c>0$ is fixed and $x_0=\mathcal{O}(\eta^{-1/2})$,
we have $2cx_0+c^2=\mathcal{O}(\eta^{-1/2})$.
Thus, for sufficiently small $\eta$ which depends on $c$,
$D_0\ge(2+\delta)/(2\eta)$, and hence 
\begin{align}\label{equ:D0}
D_0 = \Omega(1/\eta).    
\end{align}
Moreover, in time region $k \le s$, the trajectory lies in the linear region, we have
\begin{align}\label{equ:dkk}
D_k = y^2_k - q^2_k  = \left(y_{k-1} - \eta\, q_{k-1} \right)^2 - \left( q_{k-1} - \eta\,y_{k-1}\right)^2\nonumber & = (1-\eta^2)\,(y^2_{k-1} - q^2_{k-1})\nonumber \\
& = (1-\eta^2) D_{k-1}.
\end{align}

Using above recurrence relation, we get
\begin{align}\label{equ:y_kge}
y_k \ge \sqrt{D_k} = (1-\eta^2)^{k/2}\sqrt{D_0}.
\end{align}

Moreover, we have
\begin{align*}
0 < x_{s-1} = x_0 - \eta\,\sum^{s-2}_{k=1}y_k  \le  x_0 - \eta\,\sqrt{D_0}\sum^{s-2}_{k=0}\left(1-\eta^2\right)^{k/2},
\end{align*}
where the inequality in the right hand side used \eqref{equ:y_kge}. Thus we have
\begin{align*}
\sum^{s-2}_{k=0}\left(1-\eta^2\right)^{k/2} \le x_0/\left(\eta\,\sqrt{D_0}\right)    
\end{align*}
According to \eqref{equ:eos_reg}, we have 
\begin{align*}
x_0 = \CO(\eta^{-\frac{1}{2}}),\ \mathrm{and}\ \sqrt{D_0} = \Omega(\eta^{-\frac{1}{2}})\quad \mathrm{from\ \eqref{equ:D0}},
\end{align*}
these gives
\begin{align*}
\sum^{s-2}_{k=0}\left(1-\eta^2\right)^{k/2} = \CO(1/\eta).    
\end{align*}
The left hand side is a geometric progression, with an explicit solution given by
\begin{align}\label{equ:geopro}
\sum^{s-2}_{k=0}\left(1-\eta^2\right)^{k/2}  = \frac{1-\left( 1-\eta^2\right)^{\frac{s-1}{2}}}{1-\sqrt{1-\eta^2}}.
\end{align}
For the denominator term $1-\sqrt{1-\eta^2} = \frac{\eta^2}{1+\sqrt{1-\eta^2}}$, since $\eta<1$, we have
\begin{align*}
1 \le 1+\sqrt{1-\eta^2}  \le 2,
\end{align*}
thus
\begin{align*}
\eta^2/2  <  1-\sqrt{1-\eta^2} < \eta^2,
\end{align*}
which gives
\begin{align*}
\frac{1-\left( 1-\eta^2\right)^{\frac{s-1}{2}}}{\eta^2} < \frac{1-\left( 1-\eta^2\right)^{\frac{s-1}{2}}}{1-\sqrt{1-\eta^2}}   = \sum^{s-2}_{k=0}\left(1-\eta^2\right)^{k/2} = \CO(1/\eta).
\end{align*}
This gives
\begin{align*}
1-\left( 1-\eta^2\right)^{\frac{s-1}{2}}    = \CO(\eta), \quad \mathrm{i.e.,}\quad \exists\,M>0,\ \left( 1-\eta^2\right)^{\frac{s-1}{2}} \ge 1-M\eta.
\end{align*}
Now we take $\log$ to the both side, which gives
\begin{align*}
\frac{s-1}{2} < \frac{\log(1-M\eta)}{\log(1-\eta^2)} = \frac{M}{\eta} \cdot \frac{1+\CO(\eta)}{1+\CO(\eta^2)}   = \CO(1/\eta),
\end{align*}
this finished the proof.
\end{proof}


\begin{lem}\label{lem:qsent}
Under Assumption \ref{ass:bmbm}, at entrance time $s$ as defined in \eqref{equ:enttime}, we have $|x_s| \in \CO(\sqrt{\eta})$. Moreover, we have $y_s = \CO(\eta^{-\frac{1}{2}})$ and $y_s >0$.
\end{lem}

\begin{proof}
$y_s = \CO(\eta^{-\frac{1}{2}})$ is easy to prove since in the linear region $k \le s-1$, we have
\begin{align}\label{equ:yk+1<}
y_{k+1} = y_{k} - \eta\,q_{k},\ q_{k}>0 \Rightarrow y_s \le y_0 = \CO(\eta^{-\frac{1}{2}}).    
\end{align}

According to the definition of $s$, we have the following possibilities: $\mathbf{Case\, 1}: q_s\,y_s < 1$ with  $x_s >0$, and $\mathbf{Case\, 2}: x_s < 0$. In the following, we will prove the statement holds in both cases. 

\underline{We first consider Case 1.} Recall that case 1 implies $\forall t < s$, the trajectory lies in the linear region, where the following recurrence holds according to \eqref{equ:dkk}:
\begin{align*}
D_s = D_0\,\left(1-\eta^2 \right)^s.
\end{align*}

Since $s = \CO(1/\eta)$ according to Lemma \ref{lem:seta}, we have $(1-\eta^2)^s = \Omega(1)$. Moreover, we have $D_0 = \Omega(1/\eta)$ according to \eqref{equ:D0}, these together gives
\begin{align*}
D_s = \Omega(1/\eta).
\end{align*}
Since $D_s = y^2_s - q^2_s$, above gives
\begin{align*}
y_s \ge \sqrt{y^2_s- q^2_s}  = \sqrt{D_s}  = \Omega(1/\sqrt{\eta}). 
\end{align*}
Since $q_s\ge c>0$ and $c$ is fixed, the preceding lower bound
on $y_s$ implies $q_sy_s\ge cy_s>1$ for sufficiently small $\eta$.
This contradicts $q_sy_s<1$. Therefore, $x_s<0$ and Case 1 can not happen.

\underline{Now we consider Case 2}. In this case, we have $x_s < 0$ and $x_{k-1} >0$ for all $k \le s-1$. This gives
\begin{align}\label{equ:case2le}
 x_s = x_{s-1} - \eta\, y_{s-1} < 0 \Rightarrow |x_s| = \eta\,y_{s-1} - x_{s-1} < \eta\,y_{s-1}.
\end{align}

Since $y_{s-1} = \CO(\eta^{-\frac{1}{2}})$ according to \eqref{equ:yk+1<}, this gives $|x_s| = \CO(\eta^{\frac{1}{2}})$. 

Finally, since $D_{s-1}>0$, we have $y_{s-1} > q_{s-1}$, and $y_s = y_{s-1} - \eta\, q_{s-1} \ge (1-\eta)y_{s-1} >0$. This finished the $y$-part of the proof.
\end{proof}

From Lemma \ref{lem:qsent}, there exist constant $Q,\,Y$ which are independent of $\eta$, such that
\begin{align}\label{equ:boundqsys}
|x_s| \le Q\,\sqrt{\eta},\ 0<y_s<Y\eta^{-\frac{1}{2}}.   
\end{align}
 Now we fix some constant $M$ such that
\begin{align}
M \ge \max\{Q,\,Y\},
\end{align}
and our aim is to prove the following lemma about the invariance of the tube.

\begin{lem}\label{lem:inv_qsent}
Under Assumption \ref{ass:bmbm} and assume $\eta$ satisfies $\eta \left( c+M\sqrt{\eta}\right)^2<1$, we have for all $t \ge s$, the following holds:
\begin{align}\label{equ:MYub}
|x_t| \le M\, \sqrt{\eta},\ 0< y_t \le Y\, \eta^{-\frac{1}{2}}.
\end{align}
Moreover, $y_t$ is a non-increasing function after $t>s$.
\end{lem}

\begin{proof}
We prove by using induction. The base case $t =s$ is true from \eqref{equ:boundqsys} and the definition of $M$.

Assume that for a time point $t \geq s$, the inequalities in \eqref{equ:MYub} hold. Then we need to prove that at time $t + 1$, \eqref{equ:MYub} still holds.

We first note that the assumption $(c+M\sqrt{\eta})^2\eta < 1$ implies 
\begin{align}\label{equ:eaqt2}
\eta\, q_t^2 \le \eta \left( c+M\sqrt{\eta} \right)^2 <1.
\end{align}    

We dived the discussion into two cases. $\mathbf{Case \,1}: y_t\,q_t>1$ and $\mathbf{Case \,2}: y_t\,q_t<1$,
where we do not include the case $y_t\,q_t=1$ for Assumption \ref{ass:bmbm}.

\underline{We first consider Case 1.} In this region, according to \eqref{equ:lin_reg}, we have
\begin{align}
y_{t+1} = y_t -\eta\,q_t,\ y_t>1/q_t.
\end{align}
 It is easy to see $y_t$ is a non-increasing function. Moreover, since $\eta\,q^2_t < 1$, we have
\begin{align*}
\eta\,q_t < 1/q_t < y_t,    
\end{align*}
therefore $0 < y_{t+1} < y_t \le Y\,\eta^{-\frac{1}{2}}$, and this finished $y_{t+1}$ part of the induction. For the $x_{t+1}$ part, we have
\begin{align*}
|x_{t+1}| = |x_t - \eta\,\Sign(x_t)\,y_t|  = |\ |x_t| - \eta\,y_t\ | \le \max\{|x_t|,\ \eta\, y_t\} \le \max\{M, Y \}\sqrt{\eta},
\end{align*}
and this finished the proof of Case 1. 

\underline{Next we consider Case 2.} If $y_t\,q_t<1$, then according to \eqref{equ:quad_reg}, we have
\begin{align}
y_{t+1} = (1-\eta\,q_t^2)y_t.    
\end{align}
Moreover, since $0<1-\eta\,q_t^2 < 1$ according to \eqref{equ:eaqt2}, we have
\begin{align}
y_{t+1}  =    (1-\eta\,q_t^2)y_t \le y_t,
\end{align}
and this finished $y_{t+1}$ part of the induction.  For the $x$-update, (149) gives
\[
\begin{aligned}
|x_{t+1}|
=\bigl||x_t|-\eta q_ty_t^2\bigr| & \le\max\{|x_t|,\eta q_ty_t^2\}\\
&\le\max\{|x_t|,\eta y_t\}\\
&\le\max\{M,Y\}\sqrt\eta
=M\sqrt\eta,
\end{aligned}
\]
where the second inequality uses $q_ty_t<1$.
This completes the induction.


\end{proof}

In the following, we combine Lemma \ref{lem:seta}-\ref{lem:inv_qsent} to provide a proof of Proposition \ref{prop:stand_adv}.

\begin{proof}[Proof of Proposition \ref{prop:stand_adv}]
    From Lemma \ref{lem:inv_qsent}, we know $y_t > 0$ and it is a non-increasing function of $t$, thus $\lim_{t \to \infty}y_t$ exists, we denote
\begin{align*}
y_{\infty}:=  \lim_{t \to \infty}y_t \ge 0 .
\end{align*}
We will show in fact $y_{\infty} = 0$. If $y_{\infty} > 0$, note that in the linear region $q_t\,y_t>1$, after each iteration $y_t$
will decrease $\eta\,q_t > \eta\,c$. Thus there must exists some $T>0$, such that for $t>T$, the trajectory $(x_t,y_t)$ always lies in the quadratic region where $q_t\,y_t < 1$, and we have
\begin{align*}
 y_{t+1} = (1-\eta\,q_t^2)\,y_t.   
\end{align*}
Since $q_t > c$, we have $y_{t+1} \le (1-\eta\,c^2)y_t$, and this enforces $\lim_{t \to \infty}y_t = 0$.

Finally, according to Lemma \ref{lem:sharpness_quad} and Lemma \ref{lem:inv_qsent}, we have
\begin{align*}
\lambda^{c}_t \le \left( y_t+q_t \right)^2,\ q_t \le c + M\sqrt{\eta}.  
\end{align*}
Through taking the limit in above as $t \to \infty$, we get
\begin{align*}
\lim_{t \to \infty}\lambda^{c}_t \le   \limsup_{t \to \infty} q_t^2 = \CO(\eta+ c^2),  
\end{align*}
and this finished the proof.

\end{proof}
\newpage
\subsection{Experimental Details of Section \ref{sec:er}}\label{app:aes1}

In this section, we summarize the experimental details in Figure~\ref{fig:sharpness_1} and Figure~\ref{fig:sharpness_2} in Section~\ref{sec:er}. Our experimental setting follows the standard setting of the edge-of-stability literature \citep{cohen2021gradient,cohen2025understanding} with a modification to the adversarial training setting. All experiments were conducted using two NVIDIA RTX 5090 GPUs.

\paragraph{Neural network architectures \& Data Set.} We use four architectures: MLP, CNN, ResNet, and ViT, all applied to a 4 class, $1000$ example subset of CIFAR-10 (250 examples per class). The architectures are the same as those used by standard  EoS literature \citep{cohen2021gradient,cohen2025understanding}.

\emph{MLP.} A fully connected network with two hidden layers of width 256 and GELU activations, followed by a linear output layer. This gives 854K
parameters.

\emph{CNN.} A two-layer convolutional network: two $3\times3$ convolutions (base width 32, doubling to 64 in the second layer) with GELU activations and $2\times2$ average pooling after each, followed by a fully connected hidden layer (width 128) and a linear output layer. This gives 544K parameters.

\emph{ResNet.} A pre-activation residual network with 3 stages of 3 basic blocks each (9 residual blocks total), base width 16 doubling at each stage transition, GroupNorm, GELU activations, and residual branches initialized to output zero at initialization (so the network starts as the identity map). This gives 293K parameters.

\emph{ViT.} A Vision Transformer with patch size 4 (64 patches per image), embedding dimension 64, depth 4, 8 attention heads, and MLP dimension 256, using learnable LayerNorm and global average pooling in place of a class token. This gives 661K parameters.

\paragraph{Adversarial attacks.}
In Figure~\ref{fig:sharpness_1}, adversarial perturbations are constrained to an $\ell_\infty$ ball with radii $2/255$, $3/255$, and $4/255$, which bound the maximum absolute perturbation to each pixel channel. In Figure~\ref{fig:sharpness_2}, perturbations are constrained to an $\ell_2$ ball with radii $28/255$, $70/255$, and $112/255$, which bound the Euclidean norm of the perturbation over the entire image. The adversarial example is generated by the \ref{equ:fgsm} method. These radii are smaller than those commonly used in adversarial robustness benchmarks on CIFAR-10, such as $8/255$ for $\ell_\infty$ attacks and $127/255$ for $\ell_2$ attacks \citep{Pang2021}. This choice reflects our experimental objective: \textbf{to characterize differences in sharpness evolution between standard and adversarial training, rather than to maximize robust accuracy}. Our experiments demonstrate that these differences arise even with relatively small adversarial perturbations. 

Sharpness is estimated from the parameter Hessian with the generated adversarial inputs held fixed. We use warm-started LOBPCG to refresh the leading eigenvector every 100 training steps, with the solver tolerance parameter set to ($10^{-10}$). Between refreshes, we estimate sharpness using the Rayleigh quotient along the most recently computed direction.

\subsection{Additional Experimental for Section \ref{sec:er}}\label{app:aes2}

\begin{figure}[t]
    \centering
    \includegraphics[width=\textwidth]{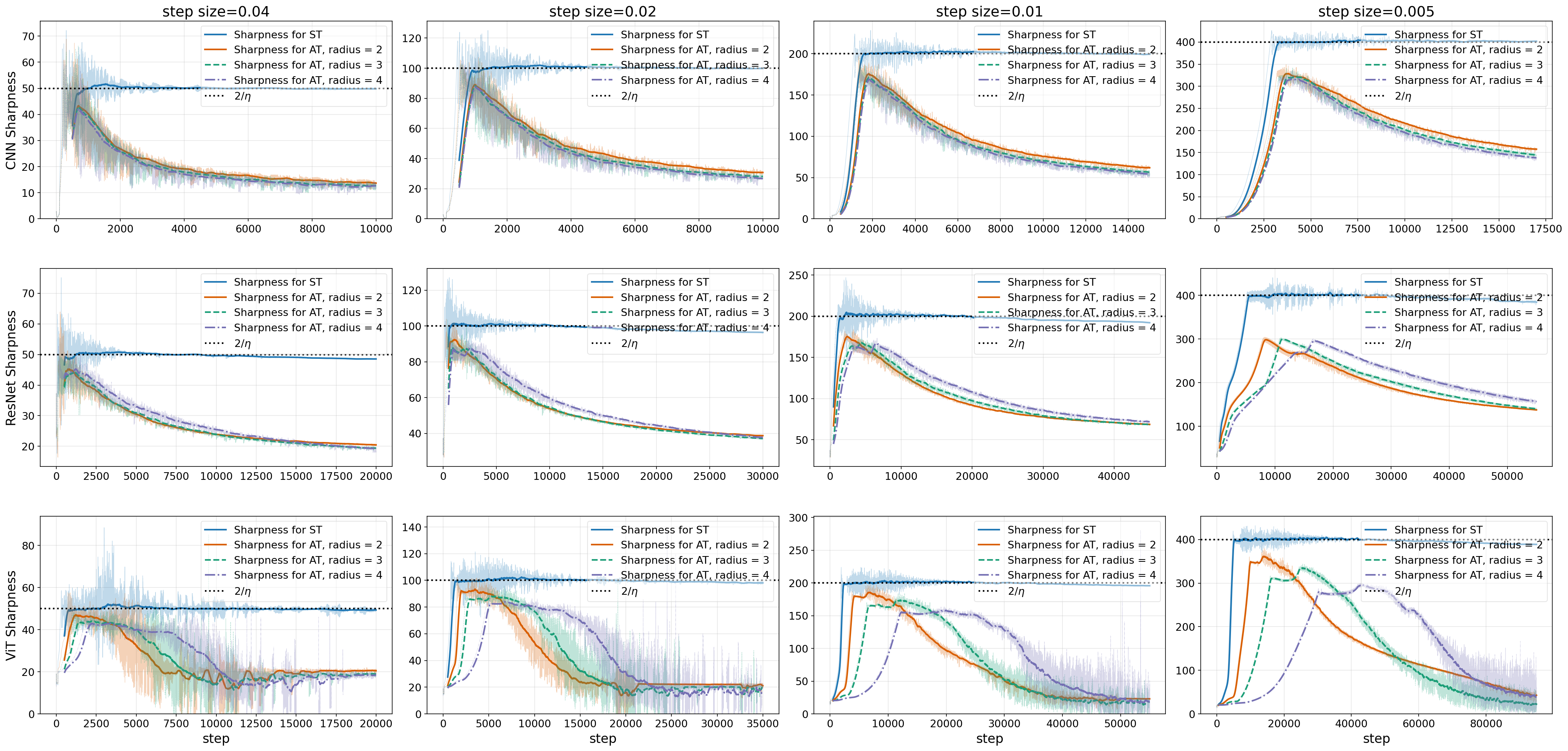}
    \caption{Sharpness evolution under standard training and $\ell_\infty$-adversarial training on CIFAR-10. Adversarial examples are generated using \eqref{equ:pgd}.
}
\label{fig:sharpness_3}
\end{figure}

\begin{figure}[t]
    \centering
    \includegraphics[width=\textwidth]{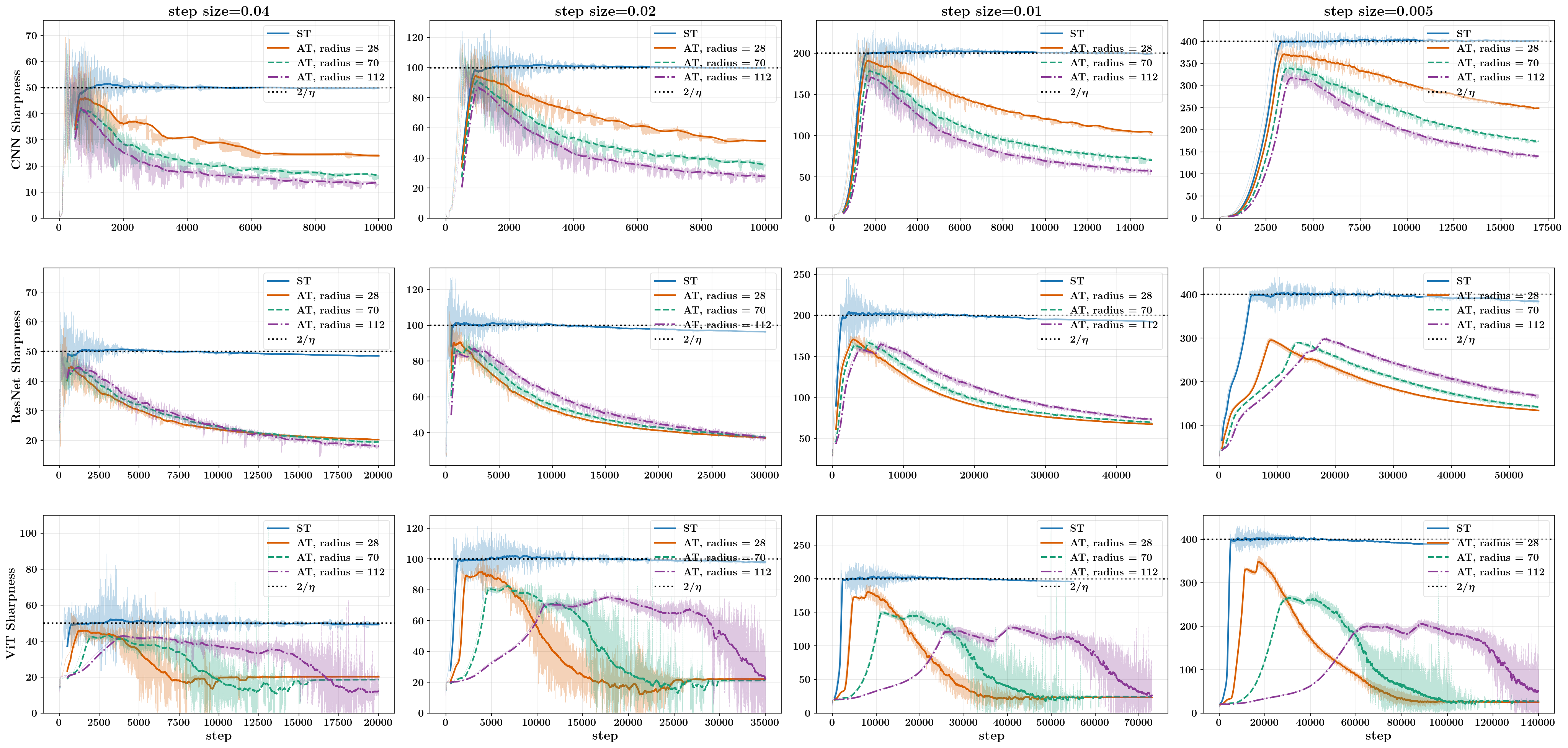}
    \caption{Sharpness evolution under
standard training and $\ell_2$-adversarial training on CIFAR-10. Adversarial examples are generated using \eqref{equ:pgd}.}
\label{fig:sharpness_4}
\end{figure}

\paragraph{Different attack methods.} In Figure~\ref{fig:sharpness_3} and Figure~\ref{fig:sharpness_4}, we present numerical results that use \ref{equ:pgd} to generate the adversarial example. Besides this, all the experimental setting are the same as those in Figure~\ref{fig:sharpness_1} and Figure~\ref{fig:sharpness_2}. Claims in Section~\ref{sec:er} are also supported by this group of experiments.

\paragraph{Evolution of adversarial losses.}
In the edge-of-stability phenomenon, when sharpness begins to oscillate around the $2/\eta$ threshold, the standard loss also begins to oscillate while decreasing over longer timescales \citep{cohen2021gradient}. We observe similar loss behavior in adversarial training: the adversarial loss initially decreases almost monotonically while adversarial sharpness increases. Once adversarial sharpness begins to oscillate with an overall decreasing trend, the adversarial loss also begins to oscillate while continuing to decrease over longer timescales. We present the comparison between 
the sharpness and loss curves by using the smallest step sizes in above experiments ($\eta = 0.005$), as in these experiments the sharpness have the longest increasing period.

\begin{figure}[t]
    \centering
    \includegraphics[width=\textwidth]{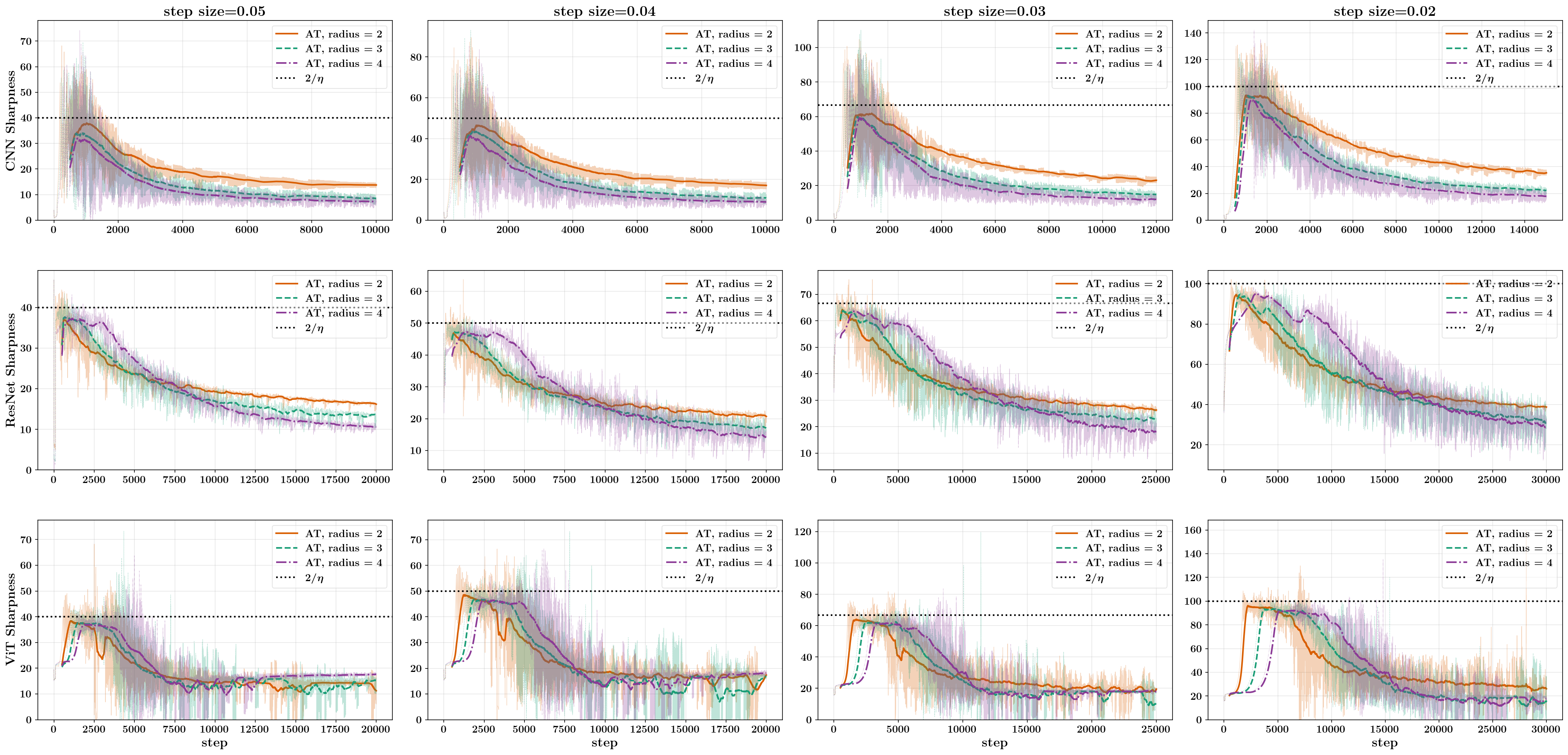}
    \caption{Sharpness evolution of CNN, ResNet, and ViT on the SVHN dataset. The adversarial examples are generated by the \ref{equ:fgsm} method with $\ell_{\infty}$ geometry. It can be observed that the sharpness also exhibits the two-stage behavior described in Section~\ref{sec:eos}.}
\label{fig:SVHN_sharpness}
\end{figure}

\paragraph{Different Dataset.} In Figure~\ref{fig:SVHN_sharpness}, we present experimental results on the SVHN dataset. SVHN consists of $32\times32$ RGB crops of house-number digits. It can be observed that the sharpness also exhibits the two-stage behavior.

\begin{figure}[h]
    \centering
    \includegraphics[width=\textwidth]{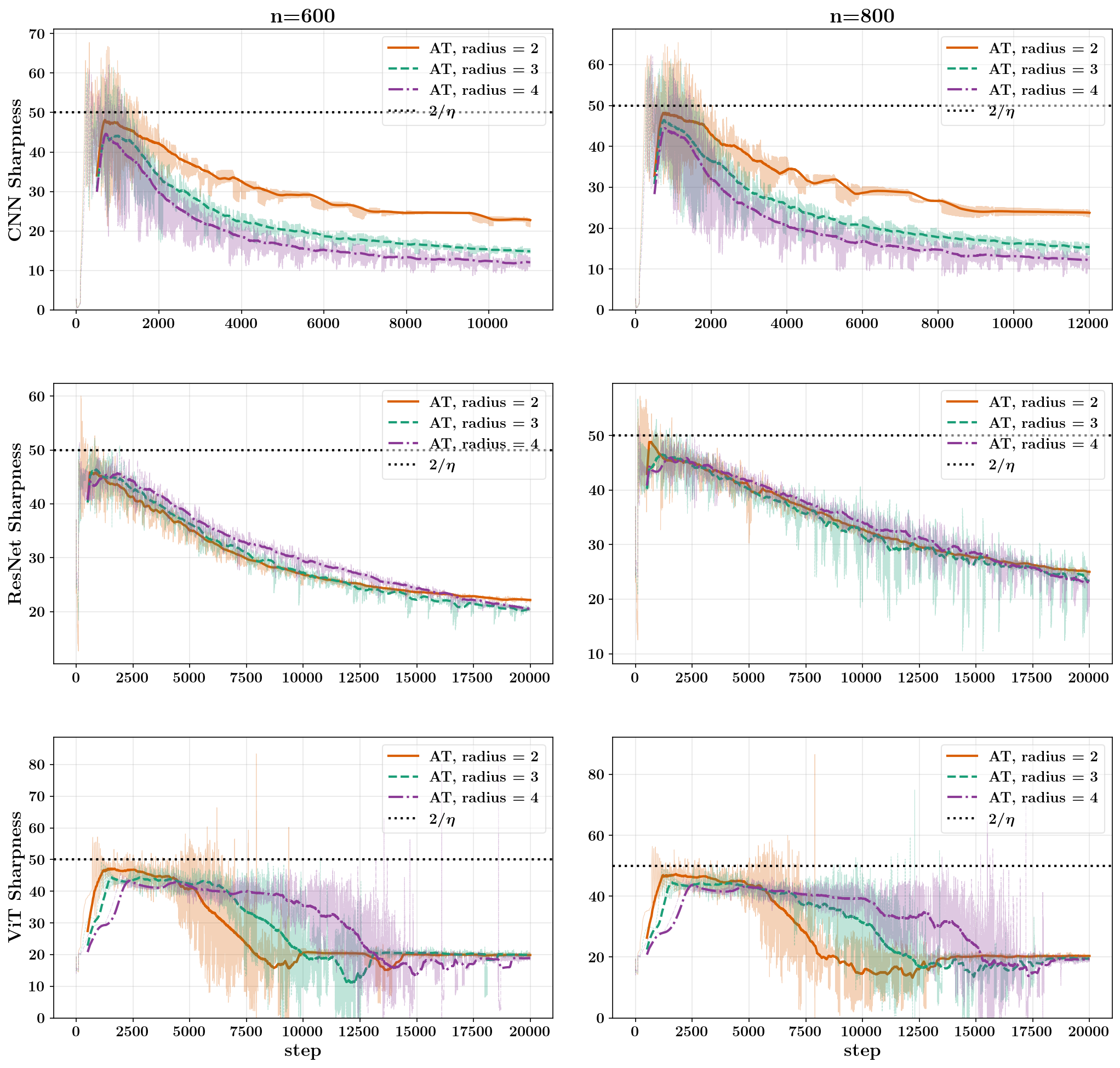}
    \caption{Sharpness evolution of CNN, ResNet, and ViT with different sample sizes.}
\label{fig:sample_sizes}
\end{figure}

\paragraph{Different Sample Sizes.} In Figure~\ref{fig:sample_sizes}, we present numerical results with sample sizes $n=600$ and $800$ fo the deterministic gradient descent. These experiments show that two-stage behavior described in Section~\ref{sec:eos} holds for differnt sample sizes.

\begin{figure}[t]
    \centering
    \includegraphics[width=\textwidth]{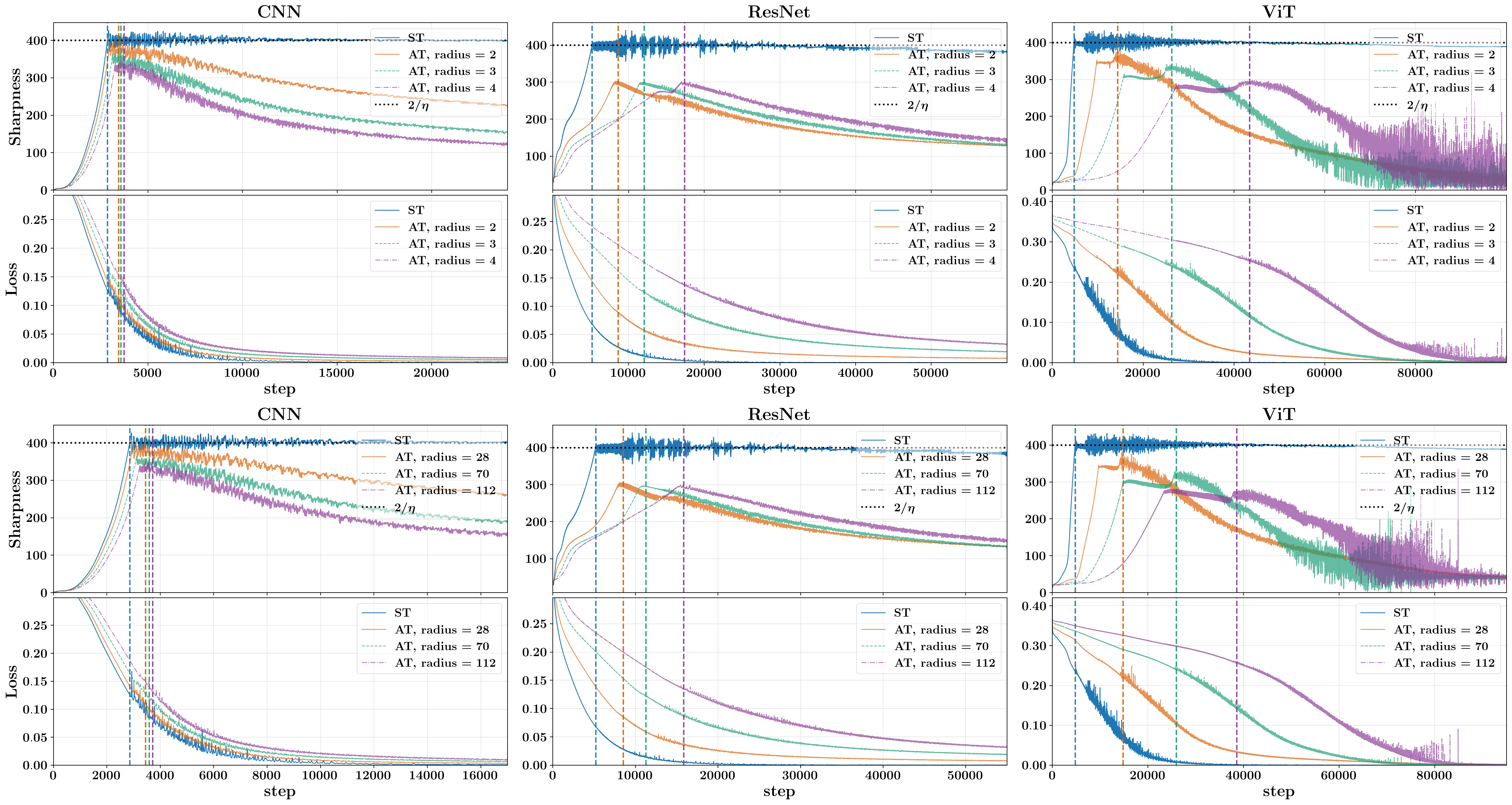}
    \caption{Comparison of the oscillation behaviors between sharpness and losses. It can be seen that when the sharpness starts to oscillate with a decreasing trend, the losses also start to oscillate. The experimental setting is step size $\eta = 0.005$ with different architectures on CIFAR-10, and the adversarial attacks are generated by \ref{equ:fgsm} with $\ell_{\infty}$ geometry (upper) or $\ell_2$ geometry (lower).
}
\label{fig:sharp_loss1}
\end{figure}

\begin{figure}[t]
    \centering
    \includegraphics[width=\textwidth]{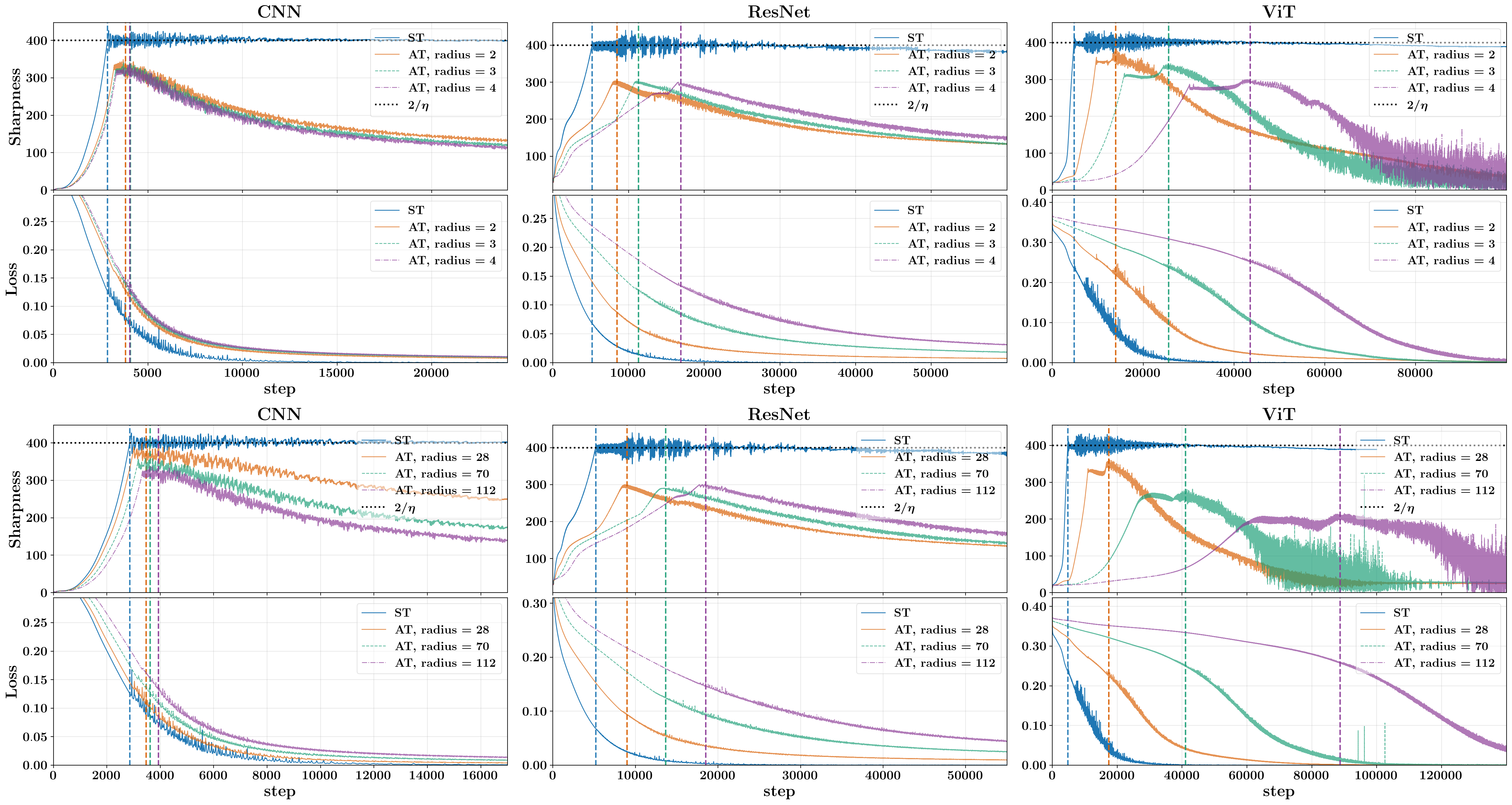}
    \caption{The experimental setting is same as Figure \ref{fig:sharp_loss1}, besides the adversarial attacks are generated by \ref{equ:pgd}.}
\label{fig:sharp_loss2}
\end{figure}


\end{document}